\documentclass[manuscript,screen]{acmart} 
\AtBeginDocument{%
  \providecommand\BibTeX{{%
    Bib\TeX}}}

\setcopyright{acmlicensed}
\copyrightyear{2025}
\acmYear{2025}
\acmDOI{XXXXXXX.XXXXXXX}
\acmJournal{CSUR}
\acmISBN{978-1-4503-XXXX-X/2018/06}

\usepackage{longtable}
\usepackage{multirow}
\usepackage{tikz}
\usetikzlibrary{shapes.geometric, arrows}
\usetikzlibrary{positioning}
\usetikzlibrary{backgrounds,mindmap}
\usepackage{xcolor}
\usetikzlibrary{calc,intersections}
\usepackage{pgfplots}
\usepackage{listings}
\usepackage{forest}
\usepackage{wrapfig}
\usetikzlibrary{arrows.meta,shapes,shadows,trees}

\usepackage{threeparttable}
\usepackage{mathrsfs}
\usepackage[scr=boondox]  
           {mathalpha}

\usepackage[many]{tcolorbox}
\usepackage{enumitem}

\begin{document}

\title{Understanding Knowledge Transferability for Transfer Learning: A Survey}



\author{Haohua Wang}
\authornote{Both authors contributed equally to this research.}
\email{wanghh24@mails.tsinghua.edu.cn}
\orcid{0009-0008-5448-5927}
\author{Jingge Wang}
\authornotemark[1]
\email{wjg22@mails.tsinghua.edu.cn}
\orcid{0000-0002-9231-882X}
\affiliation{%
  \institution{Shenzhen International Graduate School, Tsinghua University}
  \department{Shenzhen Key Laboratory of Ubiquitous Data Enabling}
  \country{China}
}

\author{Zijie Zhao}
\email{zhaozj24@mails.tsinghua.edu.cn}
\affiliation{%
  \institution{Shenzhen International Graduate School, Tsinghua University}
  \department{Shenzhen Key Laboratory of Ubiquitous Data Enabling}
  \country{China}}

\author{Yang Tan}
\email{tanyang1231@163.com}
\affiliation{%
  \institution{Shenzhen International Graduate School, Tsinghua University}
  \department{Shenzhen Key Laboratory of Ubiquitous Data Enabling}
  \country{China}
}

\author{Yanru Wu}
\email{wu-yr21@mails.tsinghua.edu.cn}
\affiliation{%
  \institution{Shenzhen International Graduate School, Tsinghua University}
  \department{Shenzhen Key Laboratory of Ubiquitous Data Enabling}
  \country{China}
}

\author{Hanbing Liu}
\email{liuhb24@mails.tsinghua.edu.cn}
\affiliation{%
  \institution{Shenzhen International Graduate School, Tsinghua University}
  \department{Shenzhen Key Laboratory of Ubiquitous Data Enabling}
  \country{China}}

\author{Jingyun Yang}
\email{yangjy20@mails.tsinghua.edu.cn}
\affiliation{%
  \institution{Shenzhen International Graduate School, Tsinghua University}
  \department{Shenzhen Key Laboratory of Ubiquitous Data Enabling}
  \country{China}}
  
\author{Enming Zhang}
\email{zem24@mails.tsinghua.edu.cn}
\affiliation{%
  \institution{Shenzhen International Graduate School, Tsinghua University}
  \department{Shenzhen Key Laboratory of Ubiquitous Data Enabling}
  \country{China}}

\author{Xiangyu Chen}
\email{davechenhello@gmail.com}
\affiliation{%
  \institution{Shenzhen International Graduate School, Tsinghua University}
  \department{Shenzhen Key Laboratory of Ubiquitous Data Enabling}
  \country{China}}

\author{Zhengze Rong}
\email{rzz22@mails.tsinghua.edu.cn}
\affiliation{%
  \institution{Shenzhen International Graduate School, Tsinghua University}
  \department{Shenzhen Key Laboratory of Ubiquitous Data Enabling}
  \country{China}}

\author{Shanxin Guo}
\affiliation{%
  \institution{Shenzhen Institute of Advanced Technology Chinese Academy of Sciences}
  \department{Center for Geo-information}
  \country{China}}
\email{sx.guo@siat.ac.cn}

\author{Yang Li}
\authornote{Corresponding author.}
\affiliation{%
  \institution{Shenzhen International Graduate School, Tsinghua University}
  \department{Shenzhen Key Laboratory of Ubiquitous Data Enabling}
  \country{China}}
\email{yangli@sz.tsinghua.edu.cn}

\renewcommand{\shortauthors}{Wang et al.}

\begin{abstract}
Transfer learning has become an essential paradigm in artificial intelligence, enabling the transfer of knowledge from a source task to improve performance on a target task. This approach, particularly through techniques such as pretraining and fine-tuning, has seen significant success in fields like computer vision and natural language processing. However, despite its widespread use, how to reliably assess the transferability of knowledge remains a challenge. Understanding the theoretical underpinnings of each transferability metric is critical for ensuring the success of transfer learning. In this survey, we provide a unified taxonomy of transferability metrics, categorizing them based on transferable knowledge types and measurement granularity. This work examines the various metrics developed to evaluate the potential of source knowledge for transfer learning and their applicability across different learning paradigms emphasizing the need for careful selection of these metrics. By offering insights into how different metrics work under varying conditions, this survey aims to guide researchers and practitioners in selecting the most appropriate metric for specific applications, contributing to more efficient, reliable, and trustworthy AI systems. Finally, we discuss some open challenges in this field and propose future research directions to further advance the application of transferability metrics in trustworthy transfer learning.
\end{abstract}



\received{23 June 2025}
\received[revised]{XX XXXX 202X}
\received[accepted]{XX XXXX 202X}

\maketitle

\section{Introduction}
Transfer learning has emerged as a powerful paradigm for leveraging knowledge from a source task to improve learning on a target task, especially when the target task has limited training data \cite{Pan2010Survey}.
One of the most prevailing practices is to pre-train a model on a source dataset and then fine-tune it on the target dataset. 
This approach has achieved widespread success in fields such as computer vision and natural language processing, etc. For instance, the general linguistic knowledge in general language models, such as GPT \cite{achiam2023gpt}, BERT \cite{2019BERT}, can be transfered to  aid better text understanding in specific target tasks. 
In computer vision, models pretrained on general image datasets like ImageNet can be fine-tuned for specific tasks, such as medical image diagnosis (e.g., detecting tumors). 
However, 
different target tasks have been shown to benefit from different source model architectures and pre-training data, meaning that the source knowledge transfers well to one target may be suboptimal for another. 
Also, transfering from dissimilar source task with very incompatible knowledge \cite{Pan2010Survey} may yield worse transfer performance, a phenomenon known as  negative transfer. 
Thus, understanding when and what source knowledge to transfer is crucial for the reliable application of transfer learning.
This leads to the introduction of   {\bf transferability metrics}, which aim to quantify the extent of candidate source knowledge, either training samples or model parameters, could help the learning of the target task under distribution shift. 
A number of such metrics have been proposed in recent years. 
These metrics generally work by applying the source knowledge to the target task, and then computing a statistic to quantify the compatibility between the source model’s learned knowledge and the new target task.
Early examples include measures based on information theory and feature dispersion (e.g., H-score \cite{bao2022informationtheoreticapproachtransferabilitytask}), probabilistic task likelihood (e.g. Log Expected Empirical Prediction or LEEP \cite{pmlr-v119-nguyen20b}), and Bayesian evidence frameworks (e.g., LogME \cite{you2021logmepracticalassessmentpretrained}), etc.
Empirical studies have reported encouraging results for many of these metrics \cite{NIPS2014_375c7134,DBLP:journals/corr/abs-1804-08328}, suggesting that effective transferability estimation is feasible and enables practitioners in different scenarios to quickly identify promising source knowledge without expensive trial-and-error.




Despite the progress, there remains little consensus on which transferability metric is most reliable, as different studies in various applications or learning paradigms have yielded conflicting conclusions. Notably, a recent comprehensive study \cite{agostinelli2022stable} systematically evaluated a broad range of transferability metrics across hundreds of thousands of transfer scenarios, and found that the evaluation of these metrics is highly dependent on the experimental setup – including the choice of source models, source datasets, target tasks, and even the evaluation protocol.
In other words, each metric has certain scenarios in which it excels and others in which it struggles. 
Also, even minor changes in the conditions can change the relative performance of metrics \cite{agostinelli2022stable}, making it difficult to draw general conclusions about which metric works the best in all scenarios. 

Since no single metric dominates across all scenarios, it is crucial for researchers and practitioners to understand the core principles and assumptions behind each transferability metric to apply them appropriately. By recognizing what each metric is actually measuring and the conditions under which it is theoretically justified, one can better match the choice of transferability metric to the specifics of a given application or transfer learning scenario, such as the nature of the applications, the similarity between source and target domains, the availability of unlabeled vs. labeled data, etc.
Take the transfer timing as an example, transferability measured at different stages  has distinct practical constraints.
When measured {\em prior to training},  for model or dataset selection,  they should ideally be computationally efficient and training-free to facilitate quick decision-making. In contrast, when
  transferability metrics are invoked {\em during} or {\em post-training}, such as when they are used for online or   retrospective analysis  of a model,  
  they thus can leverage additional performance signals such as gradients and logits of the adapted model.  

The concept of transferability has been explored in several previous surveys, all offering different perspectives on transferability metrics.
Jiang et al. \cite{jiang2022transferabilitydeeplearningsurvey} presented a survey connecting pre-training and adaptation learning to transferability, highlighting the challenges in defining and measuring transferability.  
This work emphasizes strategies to enhance transferability during model training, rather than providing specific estimation methods. 
Xue et al. \cite{10639517} systematically categorize transferability estimation methods into four types: task relatedness, source-target comparison, representation analysis, and optimal transport,
and offer guidance for selecting the most appropriate metrics for different tasks via six evaluation criteria. 
Though each division is grounded in a certain principle or application scope, this early taxonomy could not encompass more recent methods  such as gradient-based and information-theoretic methods.   
Ding et al. \cite{ding2024modeltransfersurveytransferability} provided a taxonomy and review of algorithms for  transferability restricted to the model transfer scenario, and it is the first to distinguish source-free and source-dependent estimation methods.  
More recently, 
knowledge transferability is discussed as one component of trustworthy transfer learning \cite{wu2024trustworthy}. In this work, authors reviewed methods for quantifying knowledge transferability from data, task and model perspective, and delve into methods tailored to different modalities such as graphs, text and time-series.  

Despite the volumetric works attempting to explain transferability,  previous works tend to focus on summarizing  different types of transferability metrics and often seek to identify a one-size-fits-all solution in a given context. 
In contrast, our survey aims to provide insights into how the underlying principles and assumptions of each metric influence its effectiveness, thus helping practitioners make informed decisions about selecting the right transferability metric for specific scenarios. 



This survey offers a comprehensive review of transferability metrics across various transfer learning scenarios, providing a unified definition and taxonomy of these metrics. 
We bring together the key approaches from the literature, analyzing their assumptions, strengths, and limitations, while synthesizing empirical findings to reveal how transferability works in different scenarios.
We introduce an {\bf application-friendly taxonomy} that classifies transferability metrics based on (1) the types of transferable knowledge, and (2) the granularity at which transferability is assessed.
By organizing the metrics into a coherent framework and discussing representative methods within each category, we aim to offer a clear understanding of which metrics are most appropriate for specific situations. 
In addition, we show how transferability can be {\bf applied in optimizing knowlegde transfer in eight learning scenarios}. 
Our goal is that this survey will serve not only as a guide for selecting the right transferability metric for different applications but also deepen the understanding of transfer learning itself by exploring the factors that contribute to successful or unsuccessful knowledge transfer.

The remainder of the paper is organized as follows: Section \ref{sec:sec2} presents the unified definition and taxonomy of transferability metrics, while Section \ref{sec:sec3} offers theoretical insights. Sections \ref{sec:sec4} and \ref{sec:sec5} outline various transferability estimation methods, categorized by the types of transferable knowledge and transfer learning paradigms, respectively. In Section \ref{sec:sec6}, we discuss {\bf open challenges and promising future research directions}, particularly regarding the improvement and application of transferability metrics in this important and rapidly evolving area.

 \section{How to Understand Transferability}
 \label{sec:sec2}

\subsection{A Unified Definition for Transferability }
Here, we give several definitions related to transferability, and the summary of these notations and their descriptions used in this survey can be found in Table \ref{tab:notations}.
Given a dataset $\mathcal{D}_S=\{(x_1, y_1), (x_2, y_2), \ldots, (x_n, y_n)\}$ sampled i.i.d. from a source distribution $D_S$, a learning {task} is to learn the decision function $f$ to map the data on the input space $\mathcal{X}$ to the label space $\mathcal{Y}$. 
The goal is to find a hypothesis function from the set of all possible functions that can best estimate the function $f$. 
In practical supervised learning, the decision function $f$ is often implemented through a combination of a feature extractor $g(x):\mathcal{X}\rightarrow\mathcal{Z}$ and a task head $h(g(x)):\mathcal{Z}\rightarrow\mathcal{Y}$.

\begin{table}[]
\small
\begin{tabular}{p{2cm} p{4.5cm} | p{2cm} p{4.5cm}} 
\toprule
\textbf{Symbol} & \textbf{Description}                  & \textbf{Symbol}       & \textbf{Description}                                    \\ \midrule
$f$             & Decision function                     & $H(A)$                & Entropy of event A                                      \\
$\mathscr{h}$             & Hypothesis function                   & $\star$               & Placeholder for S and T                                 \\
$\theta$        & Model parameters                      & $\lozenge$            & Placeholder for source and target  \\
$h$        & Task head                      & ~            & corresponding $\star$ \\
$\mathcal{H}$   & Hypothesis space                      & $D_\star$             & Distribution of $\lozenge$ data                         \\
$k$             & Kernel function                       & $\mathcal{D}_\star$   & The $\lozenge$ dataset                                  \\
$\mathbb{H}$    & Hilbert Space                         & $\mathcal{T}_\star$   & The $\lozenge$ task                                     \\
$d$             & A certain distance                    & $\mathcal{X}_\star$   & The $\lozenge$ instance space                           \\
$\mathcal{L}$   & Loss function                         & $\mathcal{Y}_\star$   & The $\lozenge$ label space                              \\
$g$ or $\phi$          & Feature extractor                     & $N_\star$             & Number of $\lozenge$ data                               \\
$p$             & Probability density function          & $X_\star$             & The $\lozenge$ instance set                             \\
$\mathbb{E}[X]$ & Expectation of random variable \(X\)  & $Y_\star$             & Label set corresponding $X_\star$                       \\
$\text{Var}(X)$ & Variance of \(X\)                     & $x_\star$             & an instance from $\lozenge$ dataset                     \\
$\pi(x,y)$      & Joint distribution of \(x\) and \(y\) & $y_\star$             & Label corresponding $x_\star$                           \\
$P(A)$          & Probability of event \(A\)            & $Trf(S\rightarrow T)$ & Transferability from S to T                             \\ \bottomrule
\end{tabular}
\caption{A summary of notations used in this survey.}
\label{tab:notations}
\end{table}

\begin{figure}[htbp]
    \centering
    \includegraphics[width=0.85\linewidth]{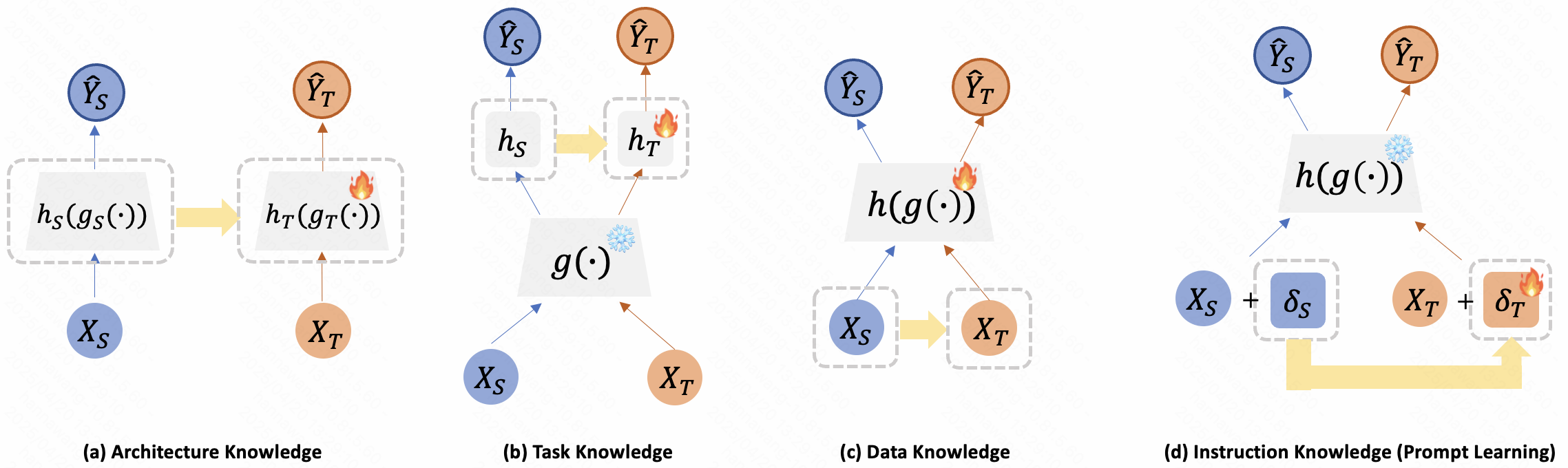}
    \caption{Different types of knowledge to be transferred. (a) Both model backbone and task head are tunable. (b) (c) The backbone of the model remains frozen, only task head is tunable. (d) An overview of prompt-based transfer learning. The backbone of the model remains frozen, while the task-specific head and prompt modules are tunable.
}
    \label{fig:prompt-learn}
\end{figure}

Transferring important source knowledge to target tasks is a common topic in various learning paradigms, while different learning problems transfer different types of knowledge. 
In a classical supervised learning setting as shown in Fig.~\ref{fig:prompt-learn} (a), a model pre-trained on a source dataset can be fine-tuned on a target task, thus reusing the entire model architecture and its learned parameters.
In other cases, only task knowledge is carried over as shown in Fig.~\ref{fig:prompt-learn} (b), for example, retaining a pre-trained feature extractor and learning a new task head, or conversely, retaining the task head while adapting it to a new feature representation; or reusing the labeled source data as important prior knowledge to assist the training for the target task shown in Fig.~\ref{fig:prompt-learn} (c).
Some other lately popular learning paradigms reveal novel knowledge types when examined through the lens of “what knowledge can be transferred.” 
For instance, in prompt learning shown in Fig.~\ref{fig:prompt-learn} (d), 
instead of updating the model weights, the pre-trained model is kept frozen while being instructed towards target tasks via learning the lightweight embeddings, named prompt.
Similarly, in reinforcement learning, the previously learned policy or transition pattern can be reused from one environment to another, thus transferring the underlying behavioral knowledge across tasks. 


Transfer learning occurs whenever knowledge acquired in one task can be reused to assist the learning of a target task.
The heterogeneity of transferable knowledge across multiple learning paradigms necessitates a unified framework. 
In this survey, we refer to these varied transferable elements—such as model parameters, model architectures, task-specific modules, data distributions, or learned strategies—as different {\bf knowledge modality}.
Furthermore, we define {\bf transferability} as a quantitative measure of how effectively the knowledge in a given source task can be reused in a new task. 
We consider transferability at two levels of granularity: {\bf task-level} transferability, which evaluates the impact of the transferred knowledge on the overall target task performance, whereas {\bf instance-level} transferability assesses its influence on individual data instances. 
Through this unified perspective of transferable knowledge type and measurement granularity, this survey offers a novel understanding of how transferability fits in diverse learning paradigms. 

\begin{definition}[Task]
    A {\em task} $\mathcal{T}$ consists of a label space $\mathcal{Y}$ and a decision function $f$, i.e., $\mathcal{T} = \{\mathcal{Y},f\}$. The decision function $f$ is an implicit one which is expected to be learned from the sample data.
\end{definition}

\begin{definition}[Transferability]
   The {\em transferability} between the source learning task $\mathcal{T}_s$ and target learning task $\mathcal{T}_t$ is defined as the effectiveness of transferring the source knowledge ${K}_s$ to the target, denoted as $Trf( {K}_s, {K}_t, E)$, where $K_s$ is the source knowledge modality that is being transferred, $K_t$ is the target knowledge modality used in computing transferability and $E\subseteq  D_t$ is the evaluation target.  
\end{definition}


\begin{figure}[htbp]
    \centering
    \includegraphics[width=0.65\linewidth]{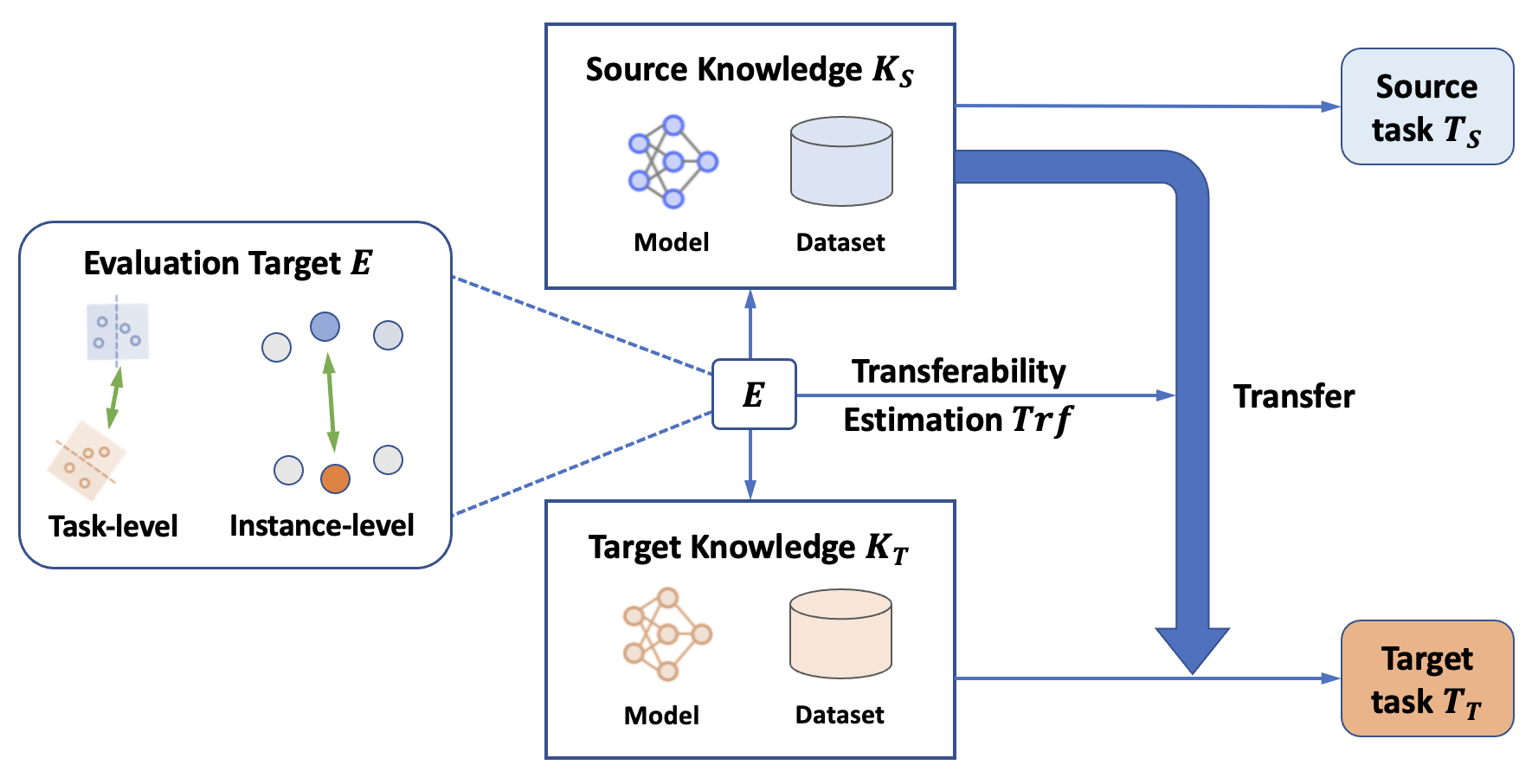}
    \caption{Transferability estimation between source and target model modality, where transferability is computed over different granularities (e.g. tasks or instances).}
    \label{fig:TE-taxonomy}
\end{figure}

\subsection{A Taxonomy of Transferability}
Using the unified transferability definition, we can group transferability problems based on the knowledge modalities they are measuring, the granularity of evaluation, and the stage where transferability is computed as shown in Fig.~\ref{fig:TE-taxonomy} and Table~\ref{tab:taxonomy-TE}. In this section, we will discuss each of these categorizations, focusing on how they are related to different transferability estimation mechanisms and applications. 
\subsubsection{Measuring Transferability for Different Knowledge Modalities}


\begin{table}[htbp]
\small
\begin{tabular}{l p{5.5cm} p{5.5cm}}
\toprule
                 & Dataset transferability             & Model transferability                                                                            \\ \midrule 

Task-level   & $Trf(\mathcal{D}_s, \mathcal{D}_t,\mathcal{D}_t)$ \quad\quad\quad\quad\quad\quad\quad\quad\quad\quad\quad\quad Application: Task embedding, dataset similarity, domain selection    & $Trf( ( f_s [,\mathcal{D}_s]), \mathcal{D}_t, \mathcal{D}_t)$ \quad\quad\quad\quad\quad\quad\quad\quad  Application: Pretrained model selection      \\
Instance-level & $Trf(\mathcal{D}_s,[\mathcal{D}_t], (x_t,y_t)) $ \quad\quad\quad\quad\quad\quad\quad\quad  Application: OOD detection, active sample selection in domain adaptation   & $Trf((f_s [,\mathcal{D}_s]), \mathcal{D}_t , (x_t,y_t))$ \quad\quad\quad\quad\quad\quad\quad\quad Application: Instance-adaptive transfer learning  \\ \bottomrule
\end{tabular}
\caption{A taxonomy of transferability based on the knowledge modality being transferred (columns) and granularity of transferability evaluation (rows).}
\label{tab:taxonomy-TE}
\end{table}

Having categorized “what to transfer” into the above three categories, we can discuss transferability measurement for each of these categories. We will consider the following question: what do we mean by transferability of a model, data and etc, and how to go about measuring them from data. 

\begin{itemize}


\item {\bf Dataset transferability $Trf(\mathcal{D}_s, \mathcal{D}_t,\cdot)$. }  From a transferability measurement perspective, thanks to the generalization analysis of domain adaptation, we can approximte transferability by statistical divergence between source and target distributions. Therefore dataset transfer essentially estimates the relationship between two tasks irrespective to the source model choice.

\item {\bf Model transferability $Trf( ( f_s [,\mathcal{D}_s]), \mathcal{D}_t, \cdot)$. }  Measuring the transferability in model transfer depends on estimating the optimal performance of the target model given source model information. 
Whether the source data is used in addition to source model would lead to {\bf source-dependent} and {\bf source-free} transferability metrics.

\noindent{\bf Prompt transferability.}
As a special case of model transferability, prompt transferability measures the effectiveness of transferring a prompt $\delta_s$ generated for a source model $f_s$ or dataset $\mathcal{D}_s$, to a target model $f_t$ or task $\mathcal{D}_t$. Prompt transferability can be formulated as:
\[
Trf((\delta_s [,\mathcal{D}_s]), (f_t, \mathcal{D}_t), \cdot) := \min_{\delta_s} \mathbb{E}_{(x,y) \sim D_t} \mathcal{L}(f_t(\delta_s(x)), y),
\]
where the objective is to minimize the loss $\ell(\cdot)$, ensuring that the prompt successfully transfers to the target model or task.


{\bf Relationship with dataset transferability. } It can be argued if the source model (such as a deep network) has sufficient capacity and is well-trained, model architecture makes no significant difference in its transfer performance as long as the source and target datasets are fixed. Therefore, one could potentially define dataset transferability via a model transferability with an optimal source model: 
\[Trf(\mathcal{D}_s,\mathcal{D}_t,\cdot):=\max_{f_s} Trf((f_s,\mathcal{D}_s),\mathcal{D}_t,\cdot)\]

\end{itemize}

\subsubsection{Granularity of Transferability Evaluation}
\begin{itemize}
\item {\bf Task level $Trf(\cdot, \cdot,\mathcal{D}_t)$.} Also known as domain level or population level. Computes transferability over multiple target 
samples. This is the most common form of transferability estimation. 

{\bf Region level $Trf(\cdot, \cdot,  R)$.} As a special case of task level transferability, which uses a subset of target samples for evaluation, in tasks with high dimensional output, such as 2D and 3D semantic segmentation, the transferability of source knowledge modality  can vary by region or semantic class. Let  $R \subset \mathcal{D}_t$  represent the target samples belonging to the same region or semantic class. If we treat the model  prediction over $R$  as an individual task,   the transfer performance on   subset  $R$   can be seen as a special kind of task-level transferability.

\item {\bf Instance level $Trf(\cdot, \cdot,\{x_t,y_t\})$.} Computes transferability with respect to a single sample. Methods based on population statistics (e.g. covariance) of the target data, are no longer applicable. On the other hand, models based on sample hypothesis testing. 
\end{itemize}

\subsubsection{When is Transferability Computed}

\begin{figure}[htbp]
    \centering
    \includegraphics[width=0.75\linewidth]{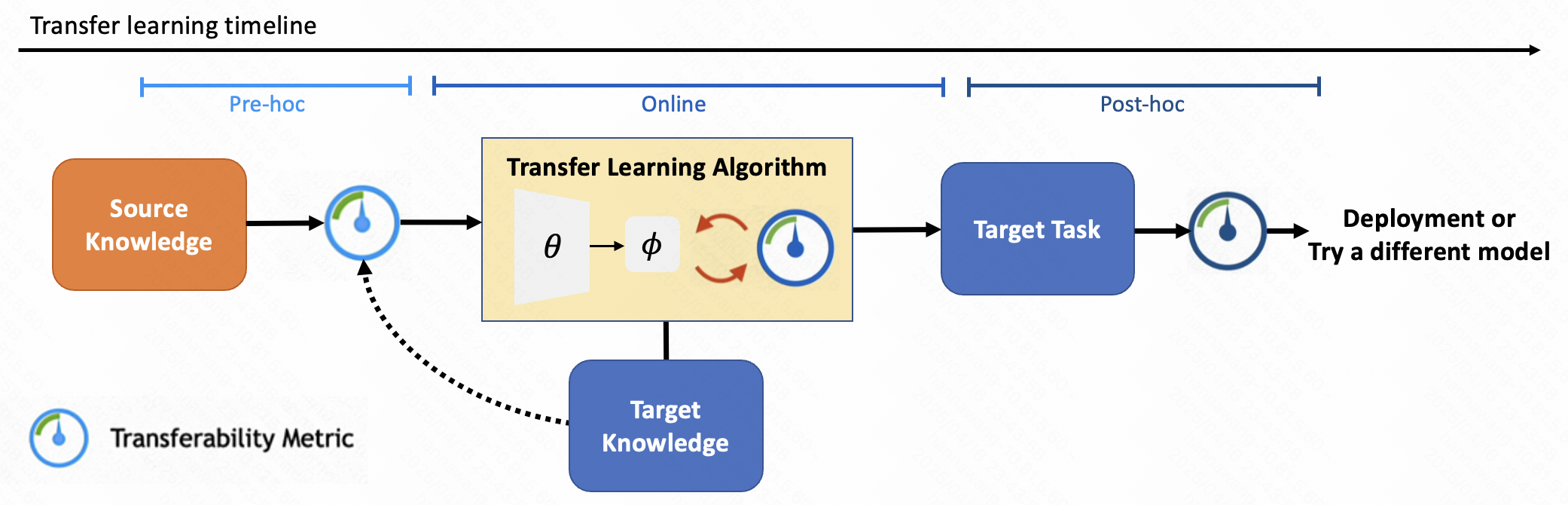}
    \caption{Three stages (pre-hoc, online, post-hoc) of when transferability is measured.}
    \label{fig:three-stages}
\end{figure}

Transferability metrics are often designed to be applied at different stages of target model training as shown in Fig.~\ref{fig:three-stages}. With different design goals, the available information and the computation requirement vary, each bringing unique advantage and shortcomings. 

{\bf Before training (pre-hoc). }  Often used as a filter to select transferable task or model prior to training, it needs to be efficient and easy to compute. Therefore statistical estimation of transferability or data-driven approach to predict transferability from available task information are preferred.

{\bf During training (online). }  Allows the utilization of gradient information or weight information during training to compute transferability. Due to the need for training information, such methods are not practicle for offline model selection. However, it is useful in dynamic learning algorithms, offering adaptive control of the transfer process.   

{\bf After training (post-hoc). } Utilizes the final training loss or accuracy to measure transferability empirically. Its obvious drawback is the computation efficiency.  As a post-hoc method, it is mostly used for analyzing and storing transferability of certain types of tasks or models.  Nevertheless, one could still use it with a small validation dataset to perform approximated model selection in a timely manner.

 \subsubsection{Beyond Pairwise Transferability}  
 Traditionally transferability is defined between a single source task and a single target task. Overtime, researchers have extended these pairwise transferability to involve multiple tasks. 
 In {multi-source transferability}, the goal is to evaluate how an ensemble of source tasks perform on the target task. 
 Besides computing pairwise and multi-source transferability, researchers have approached transferability estimation from a {\bf metric learning} perspective, i.e.
 using pre-computed pairwise transferability among a large zoo of tasks, each containing  model and dataset attributes, thus find a task embedding such that the embedding distance approximates the pairwise task transferability. Zhang et al. \cite{zhang2021quantifying} provides a method to compute the transferability among multiple domains by considering the largest domain gap.

\begin{tcolorbox}[breakable,enhanced,title = {\em Summary},left=2mm,right=2mm,fonttitle=\small,fontupper=\small, fontlower=\small]
    
\begin{itemize}[leftmargin=*]
    \item Transferability refers to how effectively knowledge from a source task can be reused in a target task.
    \item Classical examples of transferable knowledge include model parameters, neural network architectures, task-specific modules, data distributions, and learned strategies, etc.
    \item Knowledge modalities are types of knowledge that can be transferred. There are two main modalities: (1) 
    {\bf model transferability} involves transferring a source model’s knowledge to a target task, 
    and (2) {\bf dataset transferability} focuses on transferring knowledge from one dataset to another, often based on statistical divergence measurements.
    \item Granularity of transferability evaluation includes {\bf task-level transferability} which measures overall task performance across target samples, and {\bf instance-level transferability} which evaluates transferability with respect to individual samples. 
    \item Transferability can be computed {\bf pre-hoc} (before training) for task selection, {\bf online} (during training) for adaptive control, or {\bf post-hoc} (after training) for empirical evaluation.
\end{itemize}
\end{tcolorbox}

\section{Theoretical Motivation}
\label{sec:sec3}
Before we introduce the methods to measuring transferability, we provide the theoretical motivation of how accurate we could calculate the generalization ability of a model trained from source data to adapt to target data. 



\subsection{Generalization Performance based on Distribution Relationship}


In this section, we will analyze generalization bounds in a simplified learning setting, focusing on how to calculate bounds using only observed data distributions. Given the source and target data distribution $D_S$ and $D_T$, the transferability estimation $Trf(S \to T)$ can be measured using the domain divergence $d (D_S,D_T)$. We discuss common divergence metrics, such as \(H\Delta H\)-divergence, Wasserstein Distance and Maximum Mean Discrepancy (MMD), in relation to their corresponding generalization bounds.

\begin{definition}[Domain]
    A {\em domain} $\langle \mathcal{D}, f\rangle$ is composed of two parts, a dataset $\mathcal{D}$ and a decision function $f$. The dataset $\mathcal{D}$ is composed of an instant set on feature space $X = \{x| x_i \in \mathcal{X}, i = 1, ..., n\}$ on feature space $\mathcal X$ following an edge probability distribution $P(X)$. The decision function $f: \mathcal{X} \to \mathcal{Y}$ is the ground truth labeling function that outputs the conditional probability distribution $f(x_i) = \{P(y_j|x_i)|y_j \in \mathcal{Y}, j = 1, ..., |\mathcal{Y}|\}$ mapping the feature set $\mathcal{X}$ to the label space $\mathcal{Y}$. 
\end{definition}

\noindent {\bf Simplified learning problem.}
We denote the source domain and target domain as $\langle \mathcal{D}_S, f_S\rangle$ and $\langle \mathcal{D}_T, f_T\rangle$. Our problem is to find a single hypothesis function $\mathscr{h}: \mathcal{X} \to \mathcal{Y}$ from the hypothesis space $\mathcal{H}$ of all possible functions that can best estimate the decision function $f$. We define the probability according to the dataset $\mathcal{D}_S$ that a hypothesis $\mathscr{h}$ disagrees with a decision function $f$ as
\[
    \epsilon_S(\mathscr{h}, f) = \mathbf{E}_{\mathbf{x}\sim D_S}[\mathcal{L}(\mathscr{h}(\mathbf{x}),f(\mathbf{x}))].
\]

We aim to select the best hypothesis function $\mathscr{h}$ based on the minimization of its generalization error on the target domain (target error) $\epsilon_T(\mathscr{h}) = \epsilon_T(\mathscr{h}, f_T)$, which is impossible to calculate. Thus, we use bounds in terms of the source error $\epsilon_S(\mathscr{h})$ which we can calculate to measure the target error. In following discussion about bounds, we assume that the loss function \(\mathcal{L}\) is positive and symmetric and obeys the triangle inequality.

\subsubsection{Generalization Bound Measure based on $H\Delta H$-divergance}


For computational convenience, we formalize the problem as binary classification problem, which the label space can be rewritten as $\mathcal{Y} = \{0, 1\}$.

\noindent {\bf Error bounds.} Let \( \mathscr{h}^* = \arg\min_{\mathscr{h} \in \mathcal{H}} \{ \epsilon_S(\mathscr{h}) + \epsilon_T(\mathscr{h}) \} \) be the ideal joint hypothesis that minimizes the combined error on both the source and target domains, and let \( \lambda = \epsilon_S(\mathscr{h}^*) + \epsilon_T(\mathscr{h}^*) \) represent the corresponding combined error.
For a hypothesis space \( \mathcal{H} \), its corresponding \( \mathcal{H} \Delta \mathcal{H} =\{\mathscr{h}|\mathscr{h}=h_1\otimes h_2,h_1,h_2\in \mathcal{H}\}\).
Then, we can define the \( \mathcal{H}\Delta \mathcal{H} \)-divergence as \cite{ben2010theory}:
\begin{equation}
\begin{aligned}
&d_{\mathcal{H}\Delta \mathcal{H}}(D_S, D_T) 
= 2 \mathop{\text{sup}}_{\mathscr{h}, \mathscr{h}^{'}\in \mathcal{H}} |\epsilon_S(\mathscr{h}, \mathscr{h}^{'}) - \epsilon_T(\mathscr{h}, \mathscr{h}^{'})|
\end{aligned}
\end{equation}

With these definitions, we can give a bound in the following theorem: 
\begin{equation}
\label{eq:6}
    \epsilon_T(\mathscr{h}) \leq \epsilon_S(\mathscr{h}) + \frac{1}{2} d_{\mathcal{H}\Delta \mathcal{H}}(D_S, D_T) + \lambda
\end{equation}
with the empirical version as:
\begin{equation}
\label{eq:7}
\begin{aligned}
    \epsilon_T(\mathscr{h}) \leq \epsilon_S(\mathscr{h}) + \frac{1}{2} d_{\mathcal{H}\Delta \mathcal{H}}(D_S, D_T) + 4\sqrt{\frac{2d\log{(2m^{'})} + \log{(\frac{2}{\delta})}}{m^{'}}}
    + \lambda
\end{aligned}
\end{equation}

The bound on the target error \( \epsilon_T(\mathscr{h}) \) indicates that
if the \( \mathcal{H}\Delta \mathcal{H} \)-divergence between the source and target domains is small, then the target error will be close to the source error;
even if the domains are quite similar, there may still be an irreducible error \( \lambda \) due to differences between the two domains that cannot be eliminated by any hypothesis in the hypothesis space.

In practical terms, this bound suggests that if we can minimize the domain divergence (i.e., the \( \mathcal{H}\Delta \mathcal{H} \)-divergence), we can improve the generalization from the source to the target domain. 

\vspace{1em}
\noindent {\bf Learning bounds on data.}
When we arrive at domain adaptation learning problem, provided a train dataset $X = (X_S, X_T)$, $X_S$ sample $(1-\beta) m$ instances from $\mathcal{D}_S$, $X_T$ sample $\beta m$ instances from $\mathcal{D}_T$ which $\beta$ is very small. The problem setting changes to minimize the sum of source and target error 
${\epsilon}_\alpha(\mathscr{h})=\alpha{\epsilon}_T(\mathscr{h})+(1-\alpha){\epsilon}_S(\mathscr{h})$.


\subsubsection{Generalization Bound Measure based on Wasserstein Distance}
Optimal Transport (OT) provides a powerful framework for measuring transferability between source and target domains. The fundamental idea is to quantify how well data from a source domain can be mapped to a target domain, effectively bridging the gap between the two distributions \cite{shen2018wasserstein}. 

\vspace{1em}
\noindent {\bf Original OT problem.}
The Monge-Kantorovich \cite{kantorovich2006translocation} problem underlies the OT framework, aiming to find a joint probability measure \(\pi\) over two distributions \(D_S\) (source) and \(D_T\) (target) that minimizes the transport cost. Formally, this is represented as:
\begin{equation}
\pi^* = \underset{\pi \in \Pi(D_S, D_T)}{\arg \min} \int_{\Omega \times \Omega} c(\mathbf{x}, \mathbf{y})^p d\pi(\mathbf{x}, \mathbf{y}),
\end{equation}
\noindent where \(c(\mathbf{x}, \mathbf{y})\) is a cost function defining the cost of transporting a unit mass from \(\mathbf{x}\) to \(\mathbf{y}\), and \(\Pi(D_S, D_T)\) is the set of all joint distributions with marginals \(D_S\) and \(D_T\). The Wasserstein distance of order \(p\) is then defined as:
\[W_p^p(D_S, D_T) = \inf_{\pi \in \Pi(D_S, D_T)} \int_{\Omega \times \Omega} c(\mathbf{x}, \mathbf{y})^p d\pi(\mathbf{x}, \mathbf{y}).\]





By evaluating how well the source data can be aligned with the target data through the computed coupling matrix \(\pi^*\), the transferability can be quantified by the 1st order Wasserstein distance \(W_1(\hat{\mu}_S, \hat{\mu}_T)\):
\begin{equation}
W_1(\hat{D}_S, \hat{D}_T) = \min_{\pi \in \Pi(\hat{D}_S, \hat{D}_T)} \langle C, \pi \rangle_F,
\end{equation}
\noindent where \(C\) is the cost matrix defined by the dissimilarity between data points in the source and target domains.


Once the optimal coupling \(\pi^*\) is determined, it can be utilized to transform the source samples into a form that is aligned with the target domain as $\hat{\mathbf{X}}_S = \text{diag}((\pi^* \mathbf{1})^{-1}) \pi^* \mathbf{X}_T$, which enables the alignment of the source data with the target distribution.


\vspace{1em}
\noindent {\bf Generalization bound.}
Consider an unsupervised transfer learning scenario where labels of data in the target domain is unavailable. The aim is to derive a bound that connects the target error \(\epsilon_T(\mathscr{h})\) of a hypothesis \(\mathscr{h}\) to the source error \(\epsilon_S(\mathscr{h})\) and the Wasserstein distance \(W_1(D_S, D_T)\) between the source distribution \(D_S\) and the target distribution \(D_T\).


Let \(D_S\) and \(D_T\) be two probability measures on \(\mathbb{R}^d\). If the cost function is defined as \(c(\mathbf{x}, \mathbf{y}) = \|g(\mathbf{x}) - g(\mathbf{y})\|_{\mathcal{H}_{k_\mathcal{L}}}\), where \(\mathbb{H}_{k_\mathcal{L}}\) is a Reproducing Kernel Hilbert Space (RKHS) induced by a kernel \(k_l\), and if the loss function \(\mathcal{L}(\mathscr{h}(x), f(x))\) is convex, symmetric, bounded, and satisfies the triangle inequality, then the following relationship holds:
\begin{equation}
\epsilon_T(\mathscr{h}, \mathscr{h}') \leq \epsilon_S(\mathscr{h}, \mathscr{h}') + W_1(D_S, D_T)
\end{equation}
for any hypotheses \(\mathscr{h}\) and \(\mathscr{h}'\).





Let \(N_S\) and \(N_T\) be the sample sizes from the source and target domains, respectively. Then, with high probability, the target error can be bounded as follows:
\begin{equation}
\epsilon_T(\mathscr{h}) \leq \epsilon_S(\mathscr{h}) + W_1(\hat{D}_S, \hat{D}_T) + \lambda + \sqrt{2\log\left(\frac{1}{\delta}\right) / \varsigma'}\left(\sqrt{\frac{1}{N_S}} + \sqrt{\frac{1}{N_T}}\right)
\end{equation}
where \(\lambda\) represents the combined error of the optimal hypothesis minimizing the source and target errors, and \(\varsigma'\) is a constant determined by the distribution.


This bound emphasizes that minimizing the Wasserstein distance between the source and target empirical measures can enhance the target error's generalization, especially when labeled data in the target domain is sparse. 


\subsubsection{Generalization Bound Measure based on MMD}


To derive the generalization bound for Maximum Mean Discrepancy (MMD), let's start by connecting it with the domain adaptation framework and kernel embedding methods mentioned \cite{long2015learning}. The objective here is to estimate the risk on the target domain \( \epsilon_T(\mathscr{h}) \) by using the available information from the source domain \( \epsilon_S(\mathscr{h}) \), as MMD can be used to measure the discrepancy between distributions \( D_S \) (source) and \( D_T \) (target).

The general bound on the target risk \( \epsilon_T(\theta) \) for a hypothesis \( \theta \in \mathcal{H} \) can be derived as:
\begin{equation}
\epsilon_T(\mathscr{h}) \leq \epsilon_S(\mathscr{h}) + d_H(D_S, D_T) + C_0,
\end{equation}
where \( d_\mathcal{H}(D_S, D_T) \) is the \( \mathcal{H} \)-divergence between distributions \( D_S \) and \( D_T \), and \( C_0 \) is a constant that depends on the complexity of the hypothesis space and the best possible risk for both domains \cite{ben2010theory}.
A key step is approximating this divergence in a tractable way.
MMD provides a way to bound the \( \mathcal{H} \)-divergence by using kernel embeddings. If we use a kernel function \( k \) to represent distributions, MMD is defined as:
\begin{equation}
\text{MMD}^2(D_S, D_T; \mathbb{H}_k) = \left\| \mathbb{E}_{x \sim D_S}[g(x)] - \mathbb{E}_{x \sim D_T}[g(x)] \right\|_{\mathbb{H}_k}^2,
\end{equation}
where \( g: X \rightarrow \mathbb{H}_k \) is the feature map induced by the kernel \( k \), and \( \mathbb{H}_k \) is the Reproducing Kernel Hilbert Space (RKHS) associated with \( k \).
Using results from \cite{fukumizu2009kernel}, we can relate \( d_\mathcal{H}(D_S, D_T) \) with MMD. Specifically, if the RKHS \( \mathbb{H}_k \) is rich enough, \( d_\mathcal{H}(D_S, D_T) \) can be bounded by MMD and we get:
\begin{equation}
\epsilon_T(\mathscr{h}) \leq \epsilon_S(\mathscr{h}) + 2 \cdot \text{MMD}(D_S, D_T; \mathbb{H}_k) + C,
\end{equation}
where \( C  \) is a constant depending on both the hypothesis space complexity and the approximation error.

This bound shows that the target risk \( \epsilon_T(\mathscr{h}) \) can be controlled by the source risk \( \epsilon_S(\mathscr{h}) \) and the MMD between source and target distributions. By minimizing MMD (i.e., reducing the distributional difference between source and target domains), we can reduce the upper bound on the target risk, thus improving generalization on the target domain.
This is especially useful in domain adaptation, where the aim is to learn a hypothesis that generalizes well to a new, unseen domain. MMD acts as a measure of discrepancy that we can minimize to ensure better adaptation and generalization.



\subsubsection{Generalization Bound Measure based on Information Theory}

Information theory offers a principled framework to analyze generalization in transfer learning by quantifying distributional discrepancies through entropy and divergence measures. Recently, information-theoretic techniques have emerged as powerful tools to derive generalization bounds, particularly in settings involving domain adaptation and transfer learning \cite{wu2024generalization}.

Let \( \mathscr{h} \) denote the hypothesis induced by a learning algorithm trained on a dataset \( X_S=(x_S^1...x_S^{(1-\beta) n}) \) drawn from the source distribution \( D_S \) and \( X_T=(x_T^1...x_T^{\beta n}) \) drawn from the target distribution \( D_T \), and the cumulant generating function of the random variable \( \mathcal{L}(\mathscr{h}, X)-\mathbb{E}[\mathcal{L}(\mathscr{h}, X)]  \) is upper bounded by \( \psi(\lambda) \). Then we have:
\begin{equation}
\begin{aligned}
\mathbb{E}_{\mathscr{h}X_S}[ \epsilon_T(\mathscr{h})-\hat{\epsilon}_S(\mathscr{h})] \leq \frac{1}{n} \sum_{i=1}^{n} \psi^{*-1}_{-} (I(\mathscr{h}; x_S^i) + D(D_S || D_T)).
\\
-\mathbb{E}_{\mathscr{h}X_S}[ \epsilon_T(\mathscr{h})-\hat{\epsilon}_S(\mathscr{h})] \leq \frac{1}{n} \sum_{i=1}^{n} \psi^{*-1}_{+} (I(\mathscr{h}; x_S^i) + D(D_S || D_T)).
\end{aligned}
\end{equation}

Here, \( \psi^{*-1}_{\pm}(x) \) are defined as:
\[
\psi^{*-1}_{-}(x) = \inf_{\lambda \in [0, -b_{-})} \frac{x + \psi(-\lambda)}{\lambda}, \quad \psi^{*-1}_{+}(x) = \inf_{\lambda \in [0, b_{+})} \frac{x + \psi(\lambda)}{\lambda}.
\]

Let \( P_h \) be the marginal distribution induced by \( D_S, D_T \) and \( P_{\mathscr{h}|D_SD_T} \) for the ERM algorithm, and assume the loss function \( \mathcal{L}(\mathscr{h},X) \) is \( r \)-subgaussian under the distribution \( P_h \otimes D_T \). Then the following inequality holds:

\[
\mathbb{E}_\mathscr{h}[\epsilon_T(\hat {\mathscr{h}}_S)-\epsilon_T(\mathscr{h}^*_T))] \leq \frac{\sqrt{2r_2}}{n} \sum_{i=1}^{n} \sqrt{I(\hat{\mathscr{h}}; x_S^i) + D(D_S || D_T)}+d_{\mathcal{H}}(D_S, D_T) 
\]

where
\begin{equation}
    d_{\mathcal{H}}(D_S, D_T) 
= 2 \mathop{\text{sup}}_{\mathscr{h}\in \mathcal{H}} |\epsilon_S(\mathscr{h}) - \epsilon_T(\mathscr{h})|
\end{equation}

The result provides several key insights into transfer learning:
\vspace{-0.7em}
\begin{itemize}
    \item \textbf{Dependence on data distribution:} The bound incorporates the KL divergence \( D(D_S || D_T) \), which quantifies the domain shift. This indicates that larger distribution shifts generally result in weaker generalization.
    \item \textbf{Algorithmic stability:} Good learning algorithms should ensure that \( I(\mathscr{h}; x_S^i) \) vanishes as \( n \to \infty \), reinforcing the notion that stable algorithms generalize better. In other words, if the algorithm is not so good, the \( I(\mathscr{h}; x_S^i) \) can be infinity.
\end{itemize}

\subsection{Definition of Model Transferability}

Building upon the generalization bounds discussed in the previous section, we now shift our focus to the formal definition of model transferability. While divergence-based measures capture the distributional gap between domains, a more comprehensive understanding requires a clear definition of how we measure “how transferable” a model is.

Given the target training data $\{X_t, Y_t\}$ and the source model $\theta_S = \text{argmin}_\theta \sum \log P(Y_s|X_s; \theta)$ trained on source data, the transferability estimation can be defined using the empirical loss $Trf(S \to T) = \mathbf{E}_T[ P(Y_t|X_t; [\theta_S, \theta_T])]$, where $\theta_T$ is the target model. The model $\theta$ stands for the parameters of the decision function $f$ including the feature extractor $g$ and task head $h$.

We estimate the transferability between target and source tasks using their data and corresponding labels. Consider a sequence of samples with their corresponding labels from the source task \(\{X_s, Y_s\}\) and target task \(\{X_t, Y_s\}\). 
To train a model \(\theta_s\) for the source task \(\mathcal{T}_s\), we minimize the loss on \(Y_s\):
\begin{equation}
\theta_s = \arg \min_{\theta \in \Theta}\mathcal{L}_{Y_s}(\theta), 
\end{equation}
where \(\Theta\) represents the respective spaces for source model. 


To transfer the model to the target task \(\mathcal{T}_t\), we retain the transformation function \(w_Z\) and retrain only the classifier using the labels from \(\mathcal{T}_Y\). Let \(\theta_t\) denote the new classifier, chosen from a space \(\Theta\) of target classifiers. We optimize \(\theta_t\) by minimizing the loss on \(Y_t\):
\begin{equation}
\theta_t = \arg \min_{\theta \in \Theta} \mathcal{L}_{Y_t}(\theta),
\end{equation}
where \(\mathcal{L}_{Y_t}\) is analogous to the loss function for the source task but applied to the labels \(Y_t\).

The transferability of task \(\mathcal{T}_Z\) to task \(\mathcal{T}_Y\) is quantified by the expected accuracy of the model \(\theta_t\) on a test example \((x, y)\) from the target task:
\begin{equation}
\text{Trf}(\mathcal{T}_Z \rightarrow \mathcal{T}_Y) = \mathbb{E}[\text{acc}(y| x; \theta_t)],
\end{equation}
which evaluates the performance of the representation trained on \(\mathcal{T}_s\), when applied to \(\mathcal{T}_t\).

In practice, under the assumption of non-overfitting, the log-likelihood on the training set, \(\ell_{Y_t}(\theta_t)\), serves as a reliable proxy for the transferability measure. This assumption holds even for large networks trained and evaluated on datasets from the same distribution. Consequently, in subsequent sections, we employ the log-likelihood as an alternative indicator of task transferability:
\begin{equation}
\text{Trf}(\mathcal{T}_s \rightarrow \mathcal{T}_t) = \mathbb{E}[\text{log} P(Y_t| X_t; \theta_t)],
\end{equation}

\begin{tcolorbox}[breakable,enhanced,title = {\em Summary},left=2mm,right=2mm,fonttitle=\small,fontupper=\small, fontlower=\small]
\begin{itemize}[leftmargin=*]
\item A domain is defined as a pair $\langle \mathcal{D}, f\rangle$ where 1) $\mathcal{D}$ is a dataset from distribution $P(X)$ over feature space $\mathcal{X}$, 2) $f: \mathcal{X} \rightarrow \mathcal{Y}$ is the ground-truth labeling function.
\item Generalization Bound via $\mathcal{H}\Delta\mathcal{H}$-Divergence: $\mathcal{H}\Delta\mathcal{H}$-divergence measures the difference in predictions between source and target domains. 
\item Learning Bounds with Mixed Source/Target Samples: In practice, small portion $\beta$ of labeled or unlabeled target data may be available, so we use weighted error instead of target error. 
\item Generalization Bound via Wasserstein Distance (Optimal Transport): Optimal Transport (OT) theory provides an alternative way to measure how much effort it takes to align two distributions, and use that as a proxy for domain shift. 
\item Generalization Bound via Maximum Mean Discrepancy (MMD): Used to measure the difference between distributions via their mean embeddings in a Reproducing Kernel Hilbert Space (RKHS).
\item Model transferability measures how well a source-trained model performs when adapted to a target task. 

\end{itemize}

\end{tcolorbox}

\section{Tranferability Quantification Methods}
 \label{sec:sec4}

\begin{figure}[htbp]
\begin{forest} for tree={
    grow=east,
    growth parent anchor=east,
    parent anchor=east,
    child anchor=west,
    edge path={\noexpand\path[\forestoption{edge},-, >={latex}] 
         (!u.parent anchor) -- +(5pt,0pt) |- (.child anchor)
         \forestoption{edge label};}
}
[{\textbf{Transferability Metric}}, root
    [Model Transferbility, onode
        [Fine-tuning Evaluation, tnode 
            [Taskonomy \cite{DBLP:journals/corr/abs-1804-08328}{,} Attribution Maps \cite{DeepAttributionMaps}, wnode]
        ]
        [Gradient-based, tnode  
            [PGE \cite{qi2023transferabilityestimationbasedprincipal}, wnode
            ]
        ]
        [Linear Loss Approximation, tnode
            [H-score \cite{bao2022informationtheoreticapproachtransferabilitytask}{,} LogME \cite{you2021logmepracticalassessmentpretrained}{,} Linear MSE \cite{nguyen2023simpletransferabilityestimationregression}{,} LEEP \cite{pmlr-v119-nguyen20b}, wnode]
        ]
        [Uncertainty Approximation, tnode 
            [NCE \cite{tran2019transferabilityhardnesssupervisedclassification}{,} OTCE \cite{tan2024transferability}{,} TMI \cite{xu2023fastaccuratetransferabilitymeasurement}{,} CLUE \cite{DBLP:journals/corr/abs-2010-08666}, wnode]
        ] 
        [Representation Analysis, tnode 
            [Task2Vec \cite{achille2019task2vec}{,} DDS \cite{dwivedi2020dualitydiagramsimilaritygeneric}{,} RSA \cite{dwivedi2019representationsimilarityanalysisefficient}{,} SI \cite{kalhor2020rankingrejectingpretraineddeep}, wnode
            ]
        ] ]
    [Dataset Transferability, onode
        [Classification Driven, tnode 
            [MSP \cite{hendrycks2022baseline}{,} ODIN~\cite{liang2017enhancing}, wnode]
        ] 
        [Similarity Driven, tnode
            [OTDD \cite{DBLP:journals/corr/abs-2002-02923}{,} PAD \cite{NIPS2006_b1b0432c}{,} MMD \cite{long2015learning}{,} KDE \cite{rosenblat1956remarks}{,} CC-FV \cite{yang2023pickbestpretrainedmodel}{,} KL Divergence \cite{zhuang2015supervised}, wnode]
        ] 
    ] 
]
\end{forest}
\caption{Taxonomy of transferability quantification methods.}
\label{fig:taxonomy}
\end{figure}
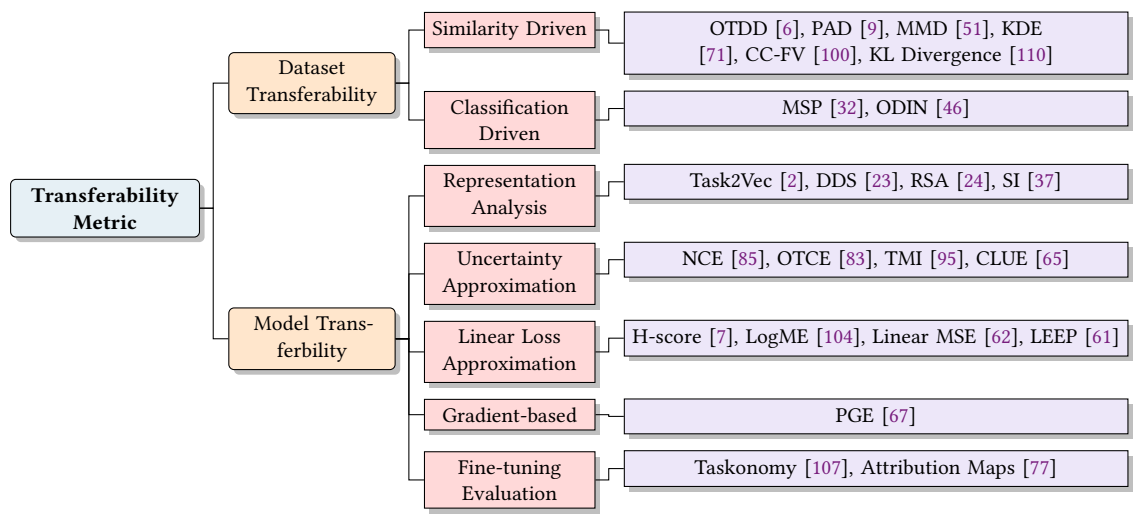

In the following section, we devide transferability quantification methods by two different knowledge modalities: dataset and model. Fig.~\ref{fig:taxonomy} shows a complete taxonomy of how we catagorize transferability metrics based on knowledge modality. Table \ref{tab:all-metrics} gives a quick insight into how we use these metrics (e.g. which stage of model training are we in to use transferability and whether we have to obtain access to data or model).

\tikzstyle{startstop} = [rectangle, rounded corners, minimum width=3cm, minimum height=1cm,text centered, draw=black, fill=red!30]
\tikzstyle{process} = [rectangle, minimum width=3cm, minimum height=1cm, text centered, draw=black, fill=blue!30]
\tikzstyle{arrow} = [thick,->,>=stealth]


\tikzset{
    basic/.style  = {draw, text width=2.3cm, drop shadow, font=\sffamily, rectangle},
    root/.style   = {basic, rounded corners=2pt, thin, align=center,fill=green!40!blue!10, font=\small},
    onode/.style = {basic, thin, rounded corners=2pt, align=center, fill=orange!20,text width=2.0cm, font=\small},
    tnode/.style = {basic, thin, align=center, fill=pink!60, text width=6.5em, font=\small},
    xnode/.style = {basic, thin, rounded corners=2pt, align=center, fill=blue!20,text width=5cm,},
    wnode/.style = {basic, thin, align=center, fill=pink!10!blue!80!red!10, text width=20.5em, font=\small},
    edge from parent/.style={draw=black, edge from parent fork right}
}

\begin{table}[htbp]
\small
\begin{tabular}{@{}cccccccc@{}}
\toprule
\multicolumn{1}{l}{Modality} & Method                    & Granularity    & $D_S$                     & $D_T$                     & $\theta_S$                & $\theta_T$                & Stage              \\ \midrule
\multirow{9}{*}{Dataset}     & \textbf{OTDD} \cite{DBLP:journals/corr/abs-2002-02923}            & task-level     & \checkmark & \checkmark &                           &                           & pre-hoc            \\
                             & \textbf{PAD} \cite{NIPS2006_b1b0432c}             & task-level     & \checkmark & \checkmark &                           &                           & pre-hoc            \\
                             & \textbf{MMD} \cite{long2015learning}  & task-level     & \checkmark & \checkmark &                           &                           & pre-hoc            \\
                             & \textbf{KDE} \cite{rosenblat1956remarks} & task-level     & \checkmark & \checkmark &                           &                           & pre-hoc            \\
                             & \textbf{CC-FV} \cite{yang2023pickbestpretrainedmodel}  & task-level     & \checkmark & \checkmark &                           &                           & pre-hoc            \\
                             & \textbf{KL Divergence}  \cite{zhuang2015supervised}  & task-level     & \checkmark & \checkmark &                           &                           & pre-hoc            \\
                             & \textbf{MSP} \cite{hendrycks2022baseline}  & instance-level & \checkmark & \checkmark &                           &                           & pre-hoc            \\
                             & \textbf{ODIN} \cite{liang2017enhancing}  & instance-level & \checkmark & \checkmark &                           &                           & pre-hoc            \\ \midrule
\multirow{14}{*}{Model}      & \textbf{Task2Vec}  \cite{achille2019task2vec}  & task-level     & \checkmark & \checkmark &                           &                           & pre-hoc            \\
                             & \textbf{DDS}  \cite{dwivedi2020dualitydiagramsimilaritygeneric}   & task-level     & \checkmark & \checkmark & \checkmark & \checkmark & pre-hoc            \\
                             & \textbf{RSA} \cite{dwivedi2019representationsimilarityanalysisefficient}  & task-level     & \checkmark & \checkmark & \checkmark & \checkmark & pre-hoc            \\
                             & \textbf{SI} \cite{kalhor2020rankingrejectingpretraineddeep} & instance-level & \checkmark & \checkmark & \checkmark & \checkmark & pre-hoc            \\
                             & \textbf{NCE} \cite{tran2019transferabilityhardnesssupervisedclassification}      & task-level     &                           & \checkmark & \checkmark &                           & pre-hoc            \\
                             & \textbf{OTCE} \cite{tan2024transferability} & task-level     &                           & \checkmark & \checkmark &                           & pre-hoc            \\
                             & \textbf{TMI}  \cite{xu2023fastaccuratetransferabilitymeasurement}   & task-level     &                           & \checkmark & \checkmark &                           & pre-hoc            \\
                             & \textbf{CLUE} \cite{DBLP:journals/corr/abs-2010-08666}   & instance-level & \checkmark & \checkmark &    \checkmark      &                           & pre-hoc            \\
                             & \textbf{H-score}  \cite{bao2022informationtheoreticapproachtransferabilitytask}   & task-level     &                           & \checkmark & \checkmark & ${ }^*$ & pre-hoc / post-hoc \\
                             & \textbf{LogME} \cite{you2021logmepracticalassessmentpretrained}    & task-level     &                           & \checkmark & \checkmark &                           & pre-hoc            \\
                             & \textbf{Linear MSE} \cite{nguyen2023simpletransferabilityestimationregression} & task-level     &                           & \checkmark & \checkmark &                           & pre-hoc            \\
                             & \textbf{LEEP}  \cite{pmlr-v119-nguyen20b} & task-level     & \checkmark & \checkmark & \checkmark &                           & pre-hoc            \\
                             & \textbf{PGE}   \cite{qi2023transferabilityestimationbasedprincipal}    & task-level     & \checkmark & \checkmark & \checkmark &                           & online             \\
                             & \textbf{Taskonomy}  \cite{DBLP:journals/corr/abs-1804-08328}      & task-level     &                           &                           & \checkmark & \checkmark & post-hoc           \\
                             & \textbf{Attribution Maps} \cite{DeepAttributionMaps} & task-level     &                           &                           & \checkmark & \checkmark & post-hoc           \\ \bottomrule
\end{tabular}
\caption{Transferability metrics categorized by knowledge type, transfer granularity, source and target datasets, data or model accessibility, and transfer stage. A \checkmark at $D_S$, $D_T$, $\theta_S$, $\theta_T$ indicates source data, target data, source model, target model required respectively.}


 \begin{minipage}{\linewidth}
 \footnotesize
    \raggedright
    ${ }^*$ Transferability evaluation based on H-score doesn't require target model parameters. If researchers need the normalized H-score transferability metric, then the target model parameters are needed.
    \end{minipage}

\label{tab:all-metrics}
\end{table}

\subsection{Dataset Tranferability Metrics}



Since dataset transferability does not involve the source model, they can be computed at any training stage, as long as source and target data  are available.

\subsubsection{Similarity Driven}

Directly comparing the similarity between source and target datasets  using a certain similarity  metric, which is often a distance between distributions.	

{\bf OTDD} (Optimal Transport Dataset Distance) \cite{DBLP:journals/corr/abs-2002-02923} 
measures the similarity between two datasets based on optimal transport theory, captures the geometric differences between data distributions in probability space to quantify their similarity. Specifically, OTDD treats each dataset as a probability distribution and then uses optimal transport theory to compute the minimal cost of transferring one distribution to another. The core idea of optimal transport theory is to find a mapping between two probability distributions that minimizes the total “transport cost" of moving from one distribution to the other. The transport cost of transport distance is defined as:
\begin{equation} 
d_{OT}(D_S, D_T) = \min_{\pi \in \prod(\alpha, \beta)}\int_{Z \times Z}d_Z(z, z^{'})d\pi(z, z^{'}),
\end{equation} 
\noindent which $d_Z$ is defined as $d_Z((x_S, y_S), (x_T, y_T)) = (d_X(x_S, x_T)^p + W_p^p(\alpha_{y_S}, \alpha_{y_T}))$. The term $d_X^p$ is the optimal transport cost, and $W_p^p$ is the p-Wasserstein distance between labels.

{\bf PAD} (Proxy A-distance) \cite{NIPS2006_b1b0432c} is
a measure of similarity between datasets from different domains, uses an adversarial validation approach. It is defined as:
\begin{equation}
    d_A = 2(1-2\epsilon),
\end{equation}
\noindent whereas $\epsilon$ is the error output by the proxy model which is trained to differentiate between the two domains. In this method, samples from the source domain are labeled as 0, and samples from the target domain are labeled as 1. When the distributions between the source domain and the target domain are very different, the classifier can easily distinguish the source of the samples, resulting in a low error rate \( \epsilon \). In this case, \( d_A \) is close to 2, indicating a significant distribution difference. When the distributions between the source domain and the target domain are very similar, the classifier finds it difficult to distinguish the source of the samples, leading to a high error rate \( \epsilon \). In this case, \( d_A \) is close to 0, indicating a small distribution difference. 

{\bf MMD} (Maximum Mean Discrepancy) \cite{long2015learning} 
is a kernel based statistical test used to determine whether given two distribution are the same, is a distance (difference) between feature means. 
It is formulated as: 
\begin{equation}
\text{MMD}(D_S, D_T)^2 = ||\mathbb{E}_{x\sim D_S}[g(x)] - \mathbb{E}_{x\sim D_T}[g(x)]||^2_{\mathbb{H}}
\end{equation}
\noindent where \( g(\cdot) \) is a kernel function (e.g.  Gaussian kernel) mapping to a certain feature space and \( \mathbb{H} \) is the reproducing kernel Hilbert space (RKHS) corresponding to the kernel function. MMD represents the distance between the mean embeddings of two distributions in the reproducing kernel Hilbert space. A larger MMD implies a greater distribution difference between the source domain and the target domain.

\begin{figure}[htbp]
    \centering
    \includegraphics[width=0.5\linewidth]{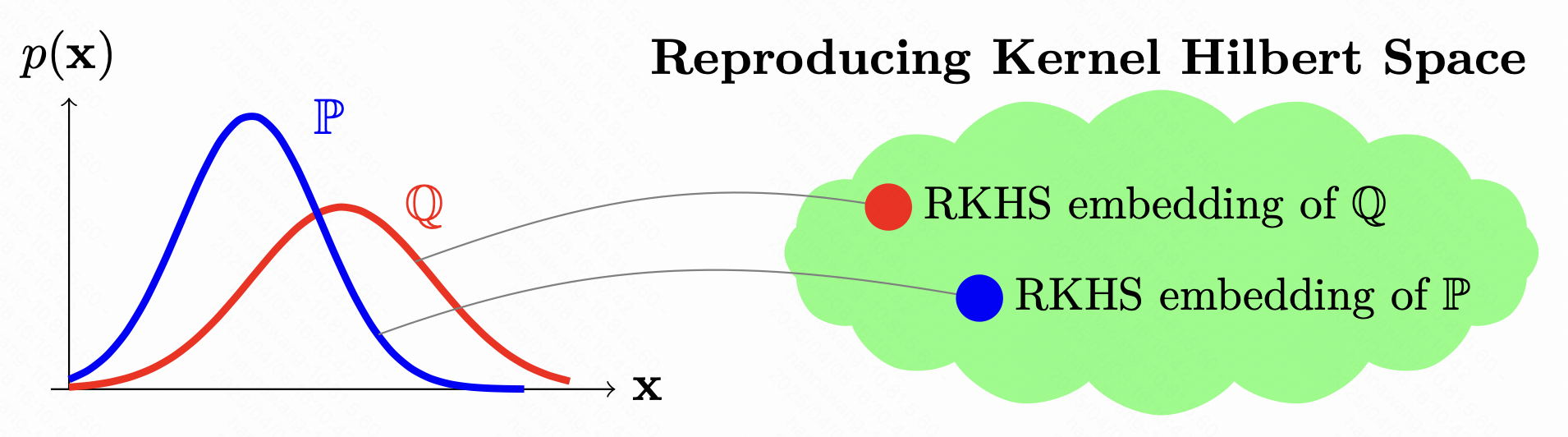}
    \caption{Embedding of marginal distributions: each distribution is mapped into a reproducing kernel Hilbert space via an expectation operation. MMD represents the distance between the mean
embeddings \cite{Muandet_2017}.}
    \label{fig:enter-label}
\end{figure}

{\bf KDE} (Kernel Density Estimation) \cite{rosenblat1956remarks} 
is an unlabelled approach to measuring the similarity between datasets, application of kernel smoothing for probability density estimation, i.e., a non-parametric method to estimate the probability density function of a random variable based on kernels as weights. 
For a dataset of samples $\{x_1, \dots, x_m\} \subset D$ and an unknown density function $p$, KDE defines an estimator for the density as:
\begin{equation}
\hat{p}(x) = \frac{1}{m}\sum^m_{i=1}k_h(x - x_i)
\end{equation}
\noindent where $k_h$ is some kernel function and $h$ is smoothing parameter called the bandwidth. The estimated density functions $\hat{p}_S$ and $\hat{p}_T$ and a distance metric $d : D_S \times D_T \to \mathbb{R}$ can then be used to estimate a distance between $D_S$ and $D_T$ using integration:
\begin{equation}
\int d(\hat{p}_S, \hat{p}_T)dx
\end{equation}

{\bf CC-FV} (Class Consistency and Feature Variety) 
\cite{yang2023pickbestpretrainedmodel} uses distribution of features extracted from foreground voxels of the same class in each sample to calculate the distance of pre-trained models and downstream datasets. Class consistency between data pair is measured by Wasserstein distance, which measures the compactness of the features within each class by evaluating how well the features from the same class cluster together, and reflects how similar the features of a particular class are. The definition of class consistency is:
\begin{equation}
    C_c = \frac{1}{N(N-1)}\sum^C{k=1}\sum_{i \ne j}W_2(F^k_i, F^k_j),
\end{equation} 
\noindent where $W_2$ is the Wasserstein distance between data pair. Feature variety employ hyperspherical potential energy which measures the expressiveness of the features themselves and the uniformity of their probability distribution, which measures the diversity and global distribution of the features, ensuring that the model captures enough variety in the features to generalize well across different datasets or tasks. It is formulated as: 
\begin{equation}
    F_v = \frac{1}{N}\sum^N_{i=1}E_s^{-1}(v),
\end{equation}
\noindent where $E_s(v)$ is hyperspherical potential energy of sampled feature of each image, which is defined as:
\begin{equation}
E_s(v) = \sum^L_{i=1}\sum^L_{j=1, j\neq i}e_s(||v_i - v_j||) = 
\left\{
\begin{array}{l}
\sum_{i\neq j} ||v_i - v_j|| ^ {-s}, s > 0\\
\sum_{i \neq j} \log (||v_i - v_j||^{-1}), s = 0
\end{array}
\right.
\end{equation}

{\bf KL Divergence} (Kullback-Leibler divergence metric / relative entropy) \cite{zhuang2015supervised} 
is a non-symmetric metric that measures the relative entropy or difference in information represented by two distributions. 
The KL divergence between $D_S$ and $D_T$ is defined as:
\begin{equation}
    d_{KL}(D_S||D_T) = \int p_S(x) \log (\frac{p_S(x)}{p_T(x)}) dx, 
\end{equation}
\noindent where $p_S$ and $p_T$ are the associated density functions. A method for approximating the KL divergence for continuous functions over multiple dimensions was presented in \cite{}. The estimated KL divergence from $D_S$ to $D_T$ is defined in \cite{} as:
\begin{equation}
    \hat{d}_{KL}(D_S||D_T) = \frac{n}{N_S}\sum^{N_S}_{i=1} \log \frac{d^k_Sk(x_i)}{d^k_T(x_i)} + \log \frac{N_T}{N_S - 1},
\end{equation}
where there are $n$ dimensions in the shared feature space, $D_S$ has $N_S$ records, $D_T$ has $N_T$ records, and $d^k_S(x_i)$ and $d^k_T(x_i)$ are the Euclidean distances to the $k^{th}$ nearest-neighbour of $x_i$ in $D_S$ and $D_T$.



{\bf MDS} (Mahalanobis distance score) \cite{NEURIPS2018_abdeb6f5} is a popular parametric approach used to measure the distance between a sample and class centroids, accounting for both the correlation between different dimensions and eliminating the influence of different scales or units. This distance is computed as:
\begin{equation}
M(x) = \max_c - \left(g(x) - \widehat{\mu}_c\right)^{\top} \hat{\boldsymbol{\Sigma}}^{-1} \left(g(x) - \widehat{\mu}_c\right),
\end{equation}
where \( x \) is the input sample, \( g(x) \) represents the feature vector of the sample \( x \), \( \widehat{\mu}_c \) is the mean (centroid) of class \( c \), \( \hat{\boldsymbol{\Sigma}} \) is the estimated covariance matrix of the feature space, which models the varianced correlations between the different features, and \( \hat{\boldsymbol{\Sigma}}^{-1} \) is the inverse of the covariance matrix, which standardizes the distance measurement by removing the effect of different scales and considering the correlation between features.

\subsubsection{Classification Driven}

Instance transferability can be evaluated by pre-training a domain discriminator that classifies a sample by its domain label and evaluate the discriminator loss on the target sample. Early research in out-of-distribution (OOD) detection uses such discriminating loss or maximum softmax probability as the indicator score, which to some extent reveals the transferability among in-distribution data and the out-of-domain sample~\cite{hendrycks2022baseline, yang2024generalized}. 

The earliest method that proposed a baseline for OOD detection task is {\bf MSP} (Maximum Softmax Probability) \cite{hendrycks2022baseline}. It directly uses the softmax output of a model. Based on the observation that the maximum softmax score of in-distribution (ID) samples is generally higher than that of OOD samples, this method sets a threshold on the maximum softmax score to distinguish between ID and OOD samples.

From the training-free perspective, it involves using the model's output (e.g., softmax scores or energy) to detect OOD samples after the model has been trained. For instance, samples with low probabilities or high energy values are considered OOD. One notable early work is ODIN~\cite{liang2017enhancing}, which uses temperature scaling and input perturbation to amplify the separability between in-distribution (ID) and OOD samples. By adjusting the temperature parameter and slightly altering the input to enhance the distinctiveness of the model's output, thereby improving the accuracy of OOD detection. The following works~\cite{liu2020energy, lin2021mood} further propose to use energy score for OOD detection, which exhibits a better theoretical interpretation from a likelihood perspective~\cite{morteza2022provable}.  

{\bf ODIN} (Out-of-Distribution Detector for Neural Networks) \cite{liang2017enhancing} improves OOD detection by addressing two key issues: overconfidence in softmax outputs and the challenge of separating ID and OOD samples. To reduce overconfidence, ODIN uses temperature scaling, which smooths the softmax distribution with the formula below: 
\begin{equation}
S_i(\boldsymbol{x}; T) = \frac{\exp(f_i(\boldsymbol{x}) / T)}{\sum_{j=1}^N \exp(f_j(\boldsymbol{x}) / T)},
\end{equation}
where \( T \) is a temperature parameter that flattens the softmax distribution. To increase the gap between ID and OOD scores, ODIN applies input preprocessing by perturbing the input image based on the gradient of the softmax with respect to the predicted class: 
\begin{equation}
\tilde{\boldsymbol{x}} = \boldsymbol{x} - \varepsilon \operatorname{sign}\left(-\nabla_{\boldsymbol{x}} \log S_{\hat{y}}(\boldsymbol{x}; T)\right), 
\end{equation}
where \( \varepsilon \) is a small scalar controlling the perturbation size. These two techniques help ODIN to more effectively distinguish between in-distribution and out-of-distribution samples.

\vspace{-6pt}

\begin{tcolorbox}[breakable,enhanced,title = {\em Summary},left=2mm,right=2mm,fonttitle=\small,fontupper=\small, fontlower=\small] 
Dataset transferability quantification methods vary significantly in their approach, but they share a common goal of evaluating how well a source dataset can be transferred to a target domain. These methods can be broadly categorized based on the type of comparison they perform. 


\begin{itemize}[leftmargin=*]
    \item {\bf Similarity-driven metrics} focus on comparing distributions between source and target datasets. 
    \item {\bf Uncertainty-driven metrics} utilize entropy which measures the distribution information as uncertainty in order to calculate transferability. 
    \item {\bf Distance-driven metrics} 
    quantify transferability by measuring statistical distances between feature distributions or class centroids, offering a granular analysis of how well a model can generalize across datasets.
\end{itemize}

\end{tcolorbox}

\subsection{Model Transferability Metrics}
\label{sec:4.2}

This section focuses on transferability of models, which consists of both pre-hoc (analyzing features without backpropogation, including Representation Analysis, Uncertainty Approximation and Linear Loss Approximation) and online or post-hoc (relying on the training process, including Gradient-based and Fine-tuning) approaches.

\subsubsection{Representation Analysis}

Evaluates the transferability of a series of source tasks by analyzing their embeddings, or assess the transferability of different pre-trained models on target tasks by analyzing (certain properties of) their representations. 

{\bf Task2Vec} 
\cite{achille2019task2vec}
process input (images) through a probe network given a dataset with ground-truth labels and a loss function defined over those labels, and compute an embedding based on estimates of the Fisher information matrix associated with the probe network parameters. Task2Vec embedding is a technique to represent tasks as elements of a vector space based on the Fisher Information Matrix, which the norm of the embedding correlates with the complexity of the task, while the distance between embeddings captures semantic similarities between tasks. 
The Task2Vec distance is defined as: 
\begin{equation}
    d(t_S \to t_T) = d(F_S, F_T) - \alpha d(t_S \to t_0), 
\end{equation}
\noindent where $d(F_S, F_T) = d_{\cos}(\frac{F_S}{F_S + F_T}, \frac{F_T}{F_S + F_T})$ is the cosine distance calculated with the Fisher Information Martix (FIM) task embeddings $F$. The FIM can be written using the Kronecker product as:
\begin{equation}
F = \mathbb{E}_{x, y\sim \hat p (x) p_w(y|x} [(y-p)^2 \cdot S \otimes xx^T],
\end{equation}
\noindent where $p=p_w(y=1 | x)$ and $S = ww^T \circ zz^T \circ (1-z)(1-z)^T$.

{\bf DDS} (Duality Diagram Similarity) \cite{dwivedi2020dualitydiagramsimilaritygeneric} 
uses duality diagram to express the data taking into account the contribution of individual observations and individual feature dimensions. DDS compare features of a set of initialization options (encoders) with features of a new task to get model initialization rankings to select the encoder initialization for learning a new task. The task feature is obtained by doing a feedforward pass through a model trained on that task. 

\begin{figure}[htbp]
    \centering
    \includegraphics[width=0.6\linewidth]{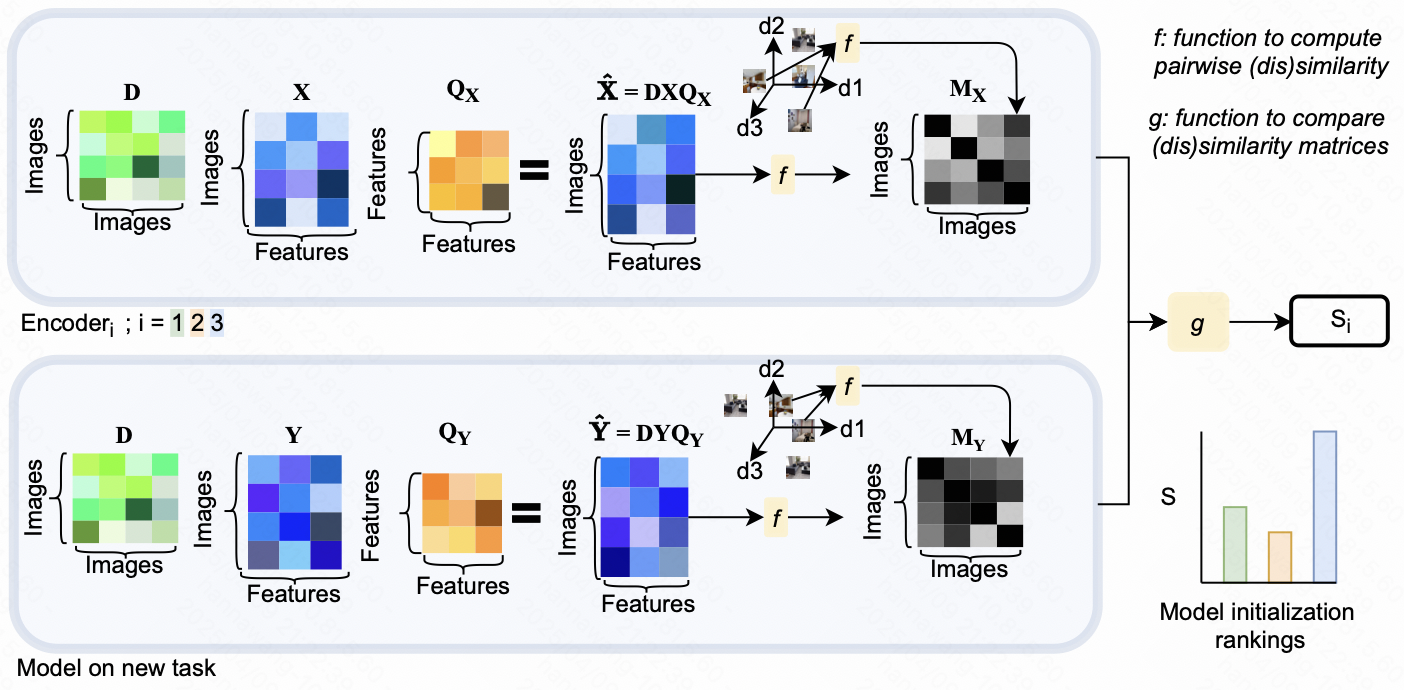}
    \caption{DDS compares model features to obtain model initialization rankings. 
    Matrix $D$ weights the images, $X$ ($Y$) stores the features from the encoder for all images, $Q_X$ ($Q_Y$) stores relations between features dimensions, $M_X$ ($M_Y$) contains pairwise (dis)similarity distances between images \cite{dwivedi2020dualitydiagramsimilaritygeneric}. 
}
    \label{fig:enter-label}
\end{figure}

{\bf RSA} (Representational Similarity Analysis) \cite{dwivedi2019representationsimilarityanalysisefficient} is used to compare DNN representations, and uses model parameters to evaluate transfer learning performance often for model initialization.

{\bf SI} (Seperation Index) \cite{kalhor2020rankingrejectingpretraineddeep} 
is a distance-based complexity metric (neighborhood measure), which compares labels to the nearest neighbor for each target sample and is calculated by the summation of these scores for each sample in the target dataset. The separation index is computed as:
\begin{equation}
SI(\{x_K^q\}^Q_{q=1}, \{y^q\}^Q_{q=1}) = \frac{1}{Q}\sum^Q_{k=1}I(y^q-y^{argmin||x^q_L = x^h_L||_h})
\end{equation}
where \(I\) is an indication function.

\vspace{-3pt}
\begin{tcolorbox}[breakable,enhanced,title={\em Remarks},left=2mm,right=2mm,colback=white,colbacktitle=white!80!gray,coltitle=black,fonttitle=\small,fontupper=\small, fontlower=\small]
\begin{itemize}[leftmargin=*]
    \item Instance-level model transferability (e.g. SI) refers to the ability of a model to adapt or transfer knowledge from one domain or task to another at the level of individual samples. This concept is widely used in tasks such as Out-of-Distribution (OOD) sample detection, anomaly detection, novelty detection, and in-context learning. In instance-level transferability, the focus is on evaluating how well a model can generalize from a source domain to a specific target sample, rather than across entire datasets.
    \item In terms of methods, multisource pretrained models are often used in conjunction with sample-wise model selection, where different models are selected or fine-tuned for individual samples based on their characteristics. 
    \item In contrast to dataset transferability, the key difference is that while dataset transferability focuses on the ability to transfer knowledge across entire datasets or domains, instance-level model transferability focuses on how well a model adapts to specific individual samples, often using more fine-grained metrics such as uncertainty or anomaly detection for decision-making.
\end{itemize}
\end{tcolorbox}

\subsubsection{Uncertainty  Approximation}

Measures transferability by approximating the target loss fucntion or risk via a simpler surrogate problem which has an efficient analytical or numerical solution. 


{\bf NCE} (negative conditional entropy) 
\cite{tran2019transferabilityhardnesssupervisedclassification} treats training labels as random variables and introduce NCE between labels of target samples extracted by a pre-trained model and the corresponding features, this value is related to the loss of the transferred model. It is defined as:
\begin{equation}
    - H(Y | Z) = \frac{1}{n}\sum^n_{i=1}\log \hat{P}(y_i | z_i), 
\end{equation}
\noindent where $\hat{P}$ is the empirical marginal distribution.

{\bf OTCE} (Optimal Transport based Conditional Entropy) \cite{tan2024transferability} 
is a linear combination of domain difference and task difference, which domain difference adopts optimal tranport (OT) definition with entropic regularization and represents in form of 1-Wasserstein distance, task difference computes empirical joint probability distribution of source label set and uses conditional entropy to represent task difference. 
The OTCE score can be formulated as:
\begin{equation}
    OTCE = \lambda_1 \hat{W}_D + \lambda_2 \hat{W}_T + b,
\end{equation}
\noindent where the domain difference is defined as:
\begin{equation}
    W_D = \sum^{m, n}_{i, j = 1} || g(x^i_S) - g(x^j_T)||^2_2 \pi^{*}_{ij},
\end{equation}
\noindent and the task difference is defined as:
\begin{equation}
    W_T = H(Y_T | Y_S) = -\sum_{y_T \in Y_T} \sum_{y_S \in Y_S} \hat{P}(y_S, y_T)\log \frac{\hat{P}(y_S, y_T)}{\hat{P}(y_S)}.
\end{equation}

{\bf TMI} (Transferability Measurement with Intra-class Feature Variance) \cite{xu2023fastaccuratetransferabilitymeasurement} 
measures the transferability by the intra-class feature variance. Specifically, the intra-class variance of target representations is measured using conditional entropy. It is formulated as:
\begin{equation}
Trf(S\to T) = H(\omega_S(X)|Y) = \sum^C_{c=1}\frac{n_c}{n}H(\omega_S(X_c)). 
\end{equation} 

{\bf CLUE} (Cluster-aware Uncertainty Estimation) \cite{DBLP:journals/corr/abs-2010-08666} measures instance-level transferability by leveraging uncertainty and diversity. To capture informativeness, CLUE utilizes predictive entropy \( H(Y|x) \), which quantifies uncertainty in the model’s predictions for a given instance \( x \). For a \( C \)-way classification, entropy is computed as:
\[
H(Y|x) = -\sum_{c=1}^C p_\Theta(Y=c|x) \log p_\Theta(Y=c|x)
\]

Under domain shift conditions, entropy reflects both uncertainty and the degree of domain discrepancy. Instead of explicitly training a domain discriminator, CLUE uses implicit domain classification based on entropy thresholding. Instances are classified as belonging to the target or source domain depending on whether their entropy exceeds a threshold \( \gamma \). The probability of an instance belonging to the target domain is proportional to its normalized entropy, scaled by the maximum possible entropy of a \( C \)-class distribution.

Diversity is assessed in the feature space of the model. Feature embeddings \( g(x) \) are grouped into \( K \) clusters using a partition function \( S \). Each cluster \( X_k \) is represented by a centroid \( \mu_k \), and diversity is evaluated by minimizing the variance \( \sigma^2(X_k) \) within each cluster:
\[
\sigma^2(X_k) = \frac{1}{2|X_k|^2} \sum_{x_i, x_j \in X_k} ||g(x_i) - g(x_j)||^2
\]

To simultaneously capture uncertainty and diversity, CLUE weights the variance of each cluster by its normalized uncertainty (entropy). The objective is to minimize the weighted population variance:
\[
\arg\min_{S, \mu} \sum_{k=1}^K \frac{1}{Z_k} \sum_{x \in X_k} H(Y|x) ||g(x) - \mu_k||^2
\]
where \( Z_k \) is the normalization factor for uncertainty in cluster \( X_k \). This formulation ensures the selection of diverse and informative instances that reflect both the feature-space coverage and the uncertainty arising from the domain shift.

\subsubsection{Linear Loss Approximation}

Linear frameworks approximate the fine-tuning layer into optimization problem. 

\begin{figure}[htbp]
    \centering
    \includegraphics[width=0.22\linewidth]{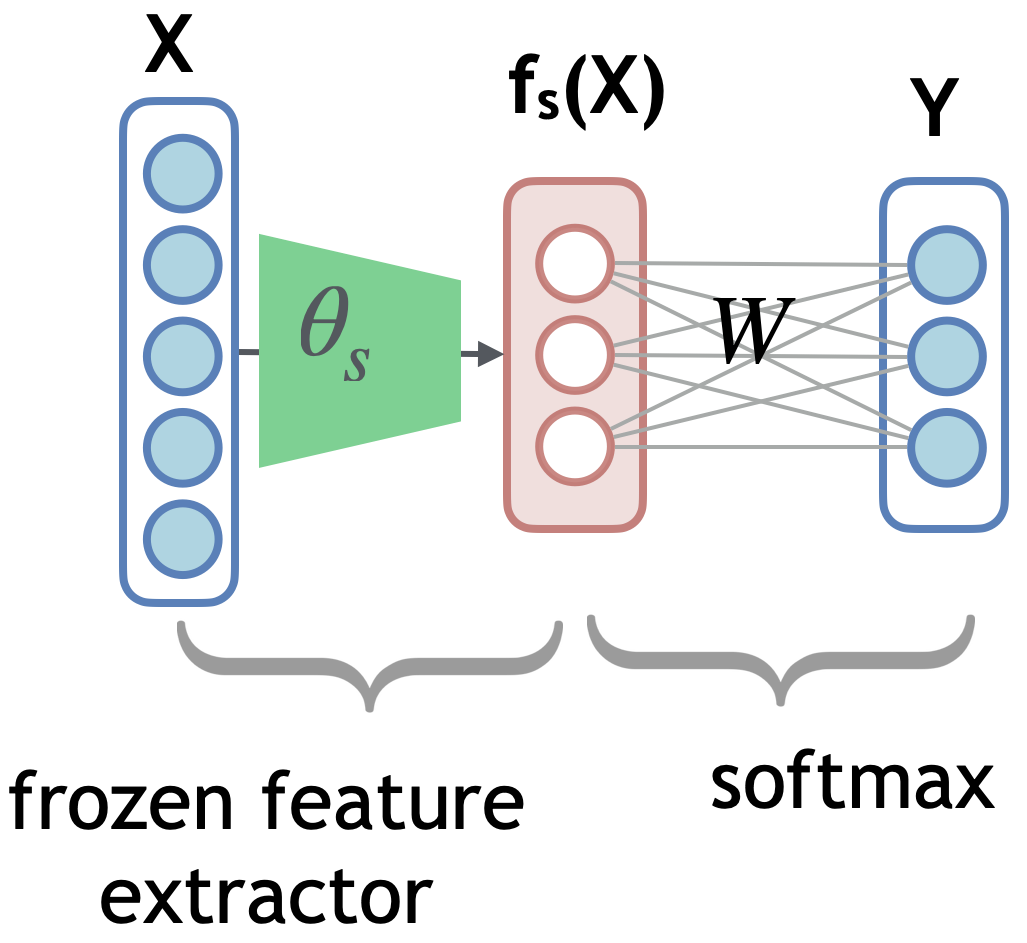}
    \caption{Adds before the target model additional free layers (consist of only linear transformations), whose parameters are optimized with respect to the target label using Alternating Conditional Expectation algorithm or neural network approach for Generalized Maximal HGR Correlation \cite{bao2022informationtheoreticapproachtransferabilitytask}.}
    \label{fig:enter-label}
\end{figure}

{\bf H-score} \cite{bao2022informationtheoreticapproachtransferabilitytask} calculates the matrix product between the pseudo reverse of the feature matrix and the covariance of the class means matrix. A higher H-score indicates a larger inter-class variance and lower feature redundancy. Given a data matrix \( X \in \mathbb{R}^{m \times d} \) with \( m \) samples and \( d \) features, and the corresponding labels \( Y \), let \( g(X) \) denote a feature transformation function with \( k \) dimensions and zero mean. The H-score of \( g \) with respect to a task characterized by a joint probability distribution \( P_{Y,X} \) is defined as:
\[
H(g) = \text{tr}\left( \text{cov}(g(X))^{-1} \, \text{cov}\left( E_{P_{X|Y}}[g(X) | Y] \right) \right),
\]
where \( \text{cov}(g(X)) \) is the covariance matrix of the feature representation \( g(X) \), and \( E_{P_{X|Y}}[g(X) | Y] \) denotes the conditional expectation of \( g(X) \) given the labels \( Y \).

The H-score can be intuitively interpreted from the perspective of a nearest neighbor classifier. A high H-score implies that the inter-class variance, represented by \( \text{cov}\left( E_{P_{X|Y}}[g(X) | Y] \right) \), is large, indicating that the feature function \( g(X) \) does a good job of separating different classes. On the other hand, a small trace of the feature covariance matrix, \( \text{tr}(\text{cov}(g(X))) \), indicates low redundancy in the feature representation.

Compared to approaches that rely on gradient descent optimization (e.g., minimizing log-loss), the H-score has the advantage of being computationally efficient, as it can be computed analytically. It only requires estimating the conditional expectation \( E[g(X) | Y] \) from sample data. Additionally, the H-score provides an operational meaning by characterizing the asymptotic error probability of using \( g(X) \) to estimate \( Y \) in a hypothesis testing context.

The transferability of features from a source task \( T_S \) to a target task \( T_T \) is captured by the transferability score \( T(S, T) \), which is defined as the ratio of the H-score of the source feature \( g_S \) applied to the target task to the H-score of the optimal target task feature \( g_T^{\text{opt}} \):
\[
T(S, T) = \frac{H_T(g_S)}{H_T(g_T^{\text{opt}})},
\]
where \( g_T^{\text{opt}} \) is the minimum error probability feature for the target task \( T_T \), and \( H_T(g_S) \) is the H-score of the source feature \( g_S \) evaluated on the target task. It follows from this definition that \( 0 \leq T(S, T) \leq 1 \), with \( T(S, T) = 1 \) indicating perfect transferability, and \( T(S, T) = 0 \) indicating no transferability.

{\bf LogME} \cite{you2021logmepracticalassessmentpretrained} 
introduces a linear model upon target features and suggests estimating the maximum average log evidence of labels given the target features, and takes bayesian approach and use maximum log evidence of target data as transferability estimator. 
This method first treats the pre-trained model \( g \) as a feature extractor and perform a forward pass of the pre-trained model on the given dataset to obtain the features \( \{\phi_i = g(x_i)\}_{i=1}^n \) and the labels \( \{y_i\}_{i=1}^n \), then uses a general statistical method that measures the relationship between features and labels using the probability density \( p(y | f) \). 
Then it applies marginalized likelihood to measure the relationship between features and labels, which uses the distribution \( p(w) \) to obtain the marginalized likelihood value:
\begin{equation}
p(y | g) = \int p(w) p(y | g, w) \, dw
\end{equation}
This approach effectively considers all possible values of \( w \) and can more accurately reflect the relationship between features and labels, avoiding overfitting issues. Here, \( p(w) \) and \( p(y | g, w) \) are determined by hyperparameters, and they do not require grid search and can be directly solved by maximizing the evidence. 

{\bf Linear MSE} 
\cite{nguyen2023simpletransferabilityestimationregression} 
uses the linear regression model between features extracted from the source model and true labels of the target training set. 
The transferability estimation is given by a transferability estimator $Trf(S\rightarrow T)$, where the Linear MSE estimator is formulated as:
\begin{equation}
Trf(S\rightarrow T) = -\min_{A, b} \{ \frac{1}{n_t} \sum^{n_T}_{i = 1} ||y_{T,i} - Aw^*(x_{T,i})- b ||^2  \}.
\end{equation}

{\bf LEEP} (Log Expected Empirical Prediction) \cite{pmlr-v119-nguyen20b} 
is the average log-likelihood of the expected empirical predictor, a simple classifier that makes prediction based on the expected empirical conditional distribution between source and target labels. 
Consider the likelihood based on the outputs of a model trained on the source dataset:
\begin{equation}
{\hat{P}}(y|z)= \frac{{\hat{P}}(y,z)}{{\hat{P}}(z)} 
= \frac{\frac{1}{n}\sum _{i:y_i=y}f (x_{T,i})_z}{\frac{1}{n}\sum ^n_{i}f (x_{T,i})_z},
\end{equation} 
\noindent where $z$ are the label outputs of the pre-trained model, $f (x_{t,i})_z$ is the estimated probability that sample $x_{t,i}$ has label $z$, and $\sum _{i:y_i=y}$ indicates all samples whose true label is $y$. LEEP is equal to the average log-likelihood of the Bayesian classifier on the target domain: 
\begin{equation}
T(\theta ) = \frac{1}{n}\sum ^n_i\log \left( \sum _z {\hat{P}}(y_i|z)f (x_{t,i})_z\right) . 
\end{equation}

\begin{tcolorbox}[breakable,enhanced,title={\em Remarks},left=2mm,right=2mm,colback=white,colbacktitle=white!80!gray,coltitle=black,fonttitle=\small,fontupper=\small, fontlower=\small]
\begin{itemize}[leftmargin=*]
\item 
\noindent{\bf Connection between   uncertainty approximation and linear loss approximation}: In OTCE, uncertainty of the predicted target label given source label $H(y_t|y_s)$ to quantify task transferability. This uncertainty term is shown to be part of   the  lower bound  of the target loss, assuming the target network is a linear softmax.

\item \noindent{\bf Connection between representation analysis and linear loss approximation}: The fisher score is a representation analysis method for feature selection. It has the same form as H-score, which can be interpreted as  a  linear loss approximation approach for transferability.  

\end{itemize}
\end{tcolorbox}


\vspace{1pt}
\subsubsection{Gradient-based Methods}~{\bf PGE} (Principle Gradient Expectation) \cite{qi2023transferabilityestimationbasedprincipal} 
considers the optimization procedure to be a distance approximation between the initial and the optimal points in the parameter space, calculate PGE of the source and target dataset and compare their similarity to approximate the transferability. The transferability gap is calculated using PGE metric, which is formulated as: 
\begin{equation}
\mathcal{G}[(D_S, T_S; (D_T, T_T))] = \frac{||PGE(D_T, T_T) - PGE(D_S, T_S)||_2}{||PGE(D_T, T_T)||_2||PGE(D_S, T_S)||_2},
\end{equation} 
\noindent where PGE is defined as:
\begin{equation}
PGE(D, T) = \mathbb{E}_{\theta_0}[\nabla L_{(D, T)}(\theta_0)].
\end{equation}


\subsubsection{Fine-tuning Evaluation}

\begin{figure}[htbp]
     \centering
     \includegraphics[width=0.25\linewidth]{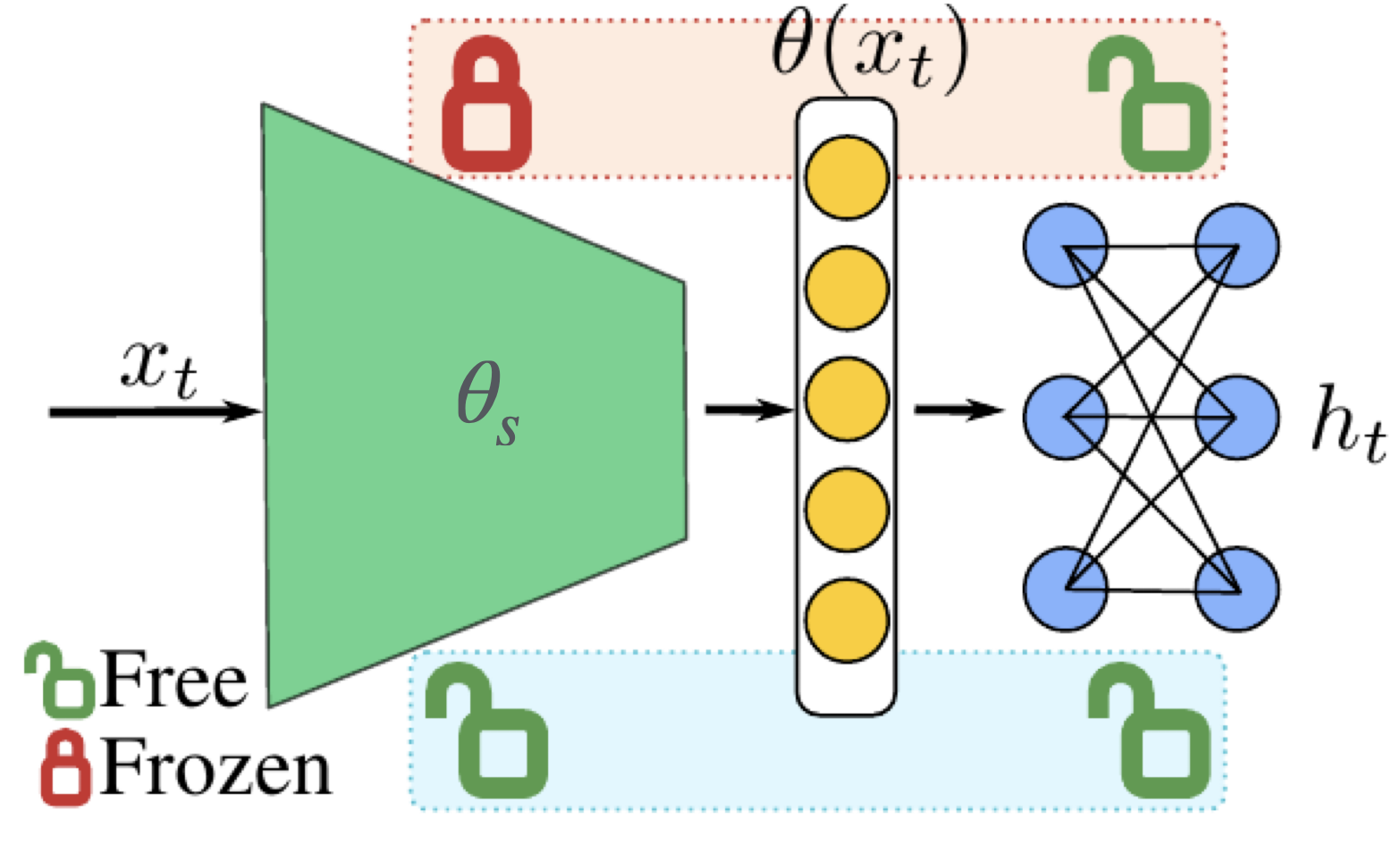}
     \caption{Simple architecture of NN-based transfer learning, where transferability is the expected likelihood target with optimal decoder (retrain head) when source features are frozen, otherwise with target feature extractor $\theta_t$ initialized by the source $\theta_s$. 
}
     \label{fig:nn-finetune}
 \end{figure}

 Evaluate transferability based the finetuning test accuracy on target data directly as shown in Fig.~\ref{fig:nn-finetune}.  

{\bf Taskonomy} \cite{DBLP:journals/corr/abs-1804-08328} 
obtains a task similarity score by comparing the transfer learning performance from each of the task-specific models and computing an affinity matrix using a function of transfer learning performance.

{\bf Attribution Maps} \cite{DeepAttributionMaps} 
measure the task relatedness via the similarity of attribution maps, which is generated by forward-and-backward propagation. For each model $m_{i}$ in $\mathcal{M}$ we can produce an attribution map $A_{j}^{i}=\left[a_{j 1}^{i}, a_{j 2}^{i}, \ldots\right] \in \mathbb{R}^{W H C}$ for each test data $X_{j}$ in the probe data $\mathcal{X}$. Each attribution map $A_{j}^{i, k}$ can be computed by adopting three off-the-shelf attribution methods to produce the attribution maps: saliency map, gradient input, and $\epsilon$-LRP. The overall attribution map $A_{j}^{i}$ can be computed through one single forward-and-backward propagation. 

After obtaining $N_{p}$ attribution maps $\mathcal{A}^{i}=\left\{A_{1}^{i}, A_{2}^{i}, \ldots, A_{N_{p}}^{i}\right\}$ for each model $m_{i}$, where $A_{j}^{i}$ denotes the attribution map of $j$-th test data $X_{j}$ in $\mathcal{X}$. The distance of two models are taken to be
\begin{equation}
d\left(m_{i}, m_{j}\right)=\frac{N_{p}}{\sum_{k=1}^{N_{p}} \cos \_\operatorname{sim}\left(A_{k}^{i}, A_{k}^{j}\right)}, 
\end{equation}
where $\cos \_\operatorname{sim}\left(A_{k}^{i}, A_{k}^{j}\right)=\frac{A_{k}^{i} \cdot A_{k}^{j}}{\left\|A_{k}^{i}\right\| \cdot\left\|A_{k}^{j}\right\|}$. The model transferability map, which measures the pairwise transferability relationships, can then be derived based on these distances.


\begin{tcolorbox}[breakable,enhanced,title = {\em Summary},left=2mm,right=2mm,fonttitle=\small,fontupper=\small, fontlower=\small] 
Model transferability metrics offer diverse approaches to quantify the ability of models or features to transfer knowledge from a source task to a target task. These metrics, ranging from representation analysis to gradient-based methods, measure different aspects of transferability such as the similarity of feature representations, task complexity, and uncertainty. 
\begin{itemize}[leftmargin=*]
    \item {\bf Representation Analysis} methods 
    focus on task and model similarity, using embeddings and similarity metrics for transferability estimation.
    \item {\bf Uncertainty Approximation} methods 
    approximate target task loss using surrogate models, balancing efficiency and accuracy.
    \item {\bf Linear Loss Approximation} methods 
    are computationally efficient and work well for estimating transferability in simpler scenarios, but may lack precision compared to more complex models.
    \item {\bf Gradient-based} methods 
    calculate transferability by comparing the gradient-based optimization gaps between source and target tasks, providing a nuanced, task-specific estimation.
    \item {\bf Fine-tuning Evaluation} methods measures the transferability through 
    fine-tuning on the target task, offering a more empirical but computationally demanding metric.
\end{itemize}
\end{tcolorbox}

\section{Applying Transferability in Diverse Learning Paradigms}
 \label{sec:sec5}

Transferability plays a crucial role in a variety of learning paradigms that extend beyond model selection, which will be discussed in this section. 
Each of these learning paradigms presents unique challenges and opportunities for optimizing knowledge transferability, making it a central consideration in modern machine learning research.


\subsection{Single-Task Model Transfer}  



Single-task transfer learning represents the most straightforward model adaptation paradigm, where knowledge from a pre-trained model is transferred to a single target task, typically through finetuning. The goal is to enhance performance on the target task, thus reducing the computational cost of training from scratch.

One of the primary challenges in single-task transfer is selecting the most appropriate pre-trained model. To address this, pre-hoc model transferability metrics introduced in Section \ref{sec:4.2} can be used to estimate a model’s suitability for a given task before finetuning. Methods such as LEEP \citep{pmlr-v119-nguyen20b} and NCE \citep{tran2019transferabilityhardnesssupervisedclassification} quantify the alignment between source and target task distributions, enabling more informed model selection. 

Beyond model selection, transferability also informs strategies for improving transfer performance. Methods such as OTCE finetune enhances the transferability of pretrained features in cross-task cross-domain scenarios by maximizing the OTCE score \cite{tan2024transferability}. 


\subsection{Domain Adaptation}

Domain adaptation focuses on transferring knowledge from a labeled source domain to an unlabeled or sparsely labeled target domain, where the two domains differ in data distribution. The main difference with model transfer is that domain adaptation tries to learn a domain-invariant model (which can extract transferable knowledge across domains) using both source and target data.

One main mainstream approach is obtaining a transferable model by aligning the features of the source and target domains. To enhance the model transferability, several statistical measures are introduced as a guide to reduce the distributional discrepancy between the two domains.
Some of the most classic used metrics for comparing and reducing distribution shift include maximum mean discrepancy (MMD) \cite{long2015learning,long2016unsupervised, long2017deep}, Kullback–Leibler (KL) divergence \cite{zhuang2015supervised, priyatikanto2023improving}, correlation distances \cite{sun2017correlation}, $\mathcal{A}$-distance \cite{huang2022balancing}, and Wasserstein distance\cite{shen2018wasserstein}.
These statistical methods aim to produce feature representations that are invariant across both domains, fostering transferability.

\begin{wrapfigure}{r}{0.30\textwidth}
    \centering
    \includegraphics[width=1.05\linewidth]{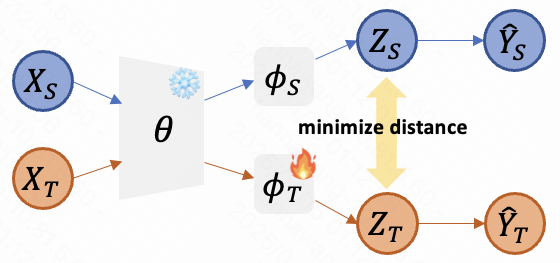}
    \caption{Domain adaptation problem.}
    \label{fig:DA}
\end{wrapfigure}

Another line of works offers a different perspective on how transferability can be achieved between domains. Incorporating with a domain discriminator, adversarial-based methods encourage domain confusion by training a domain-invariant feature extractor to deceive the discriminator, ensuring that the source and target features become indistinguishable.
To be more specific, this is achieved by minimizing an approximate domain discrepancy between the source and target distributions via an adversarial objective that makes samples from two domains sufficiently indistinguishable to confuse the domain discriminator \cite{ganin2016domain, tzeng2017adversarial, long2018conditional, li2022enhancing}.
The model is thus encouraged to obtain more generalizable features that are not tied to specific domains, promoting better adaptation.

\subsection{Multi-source Transfer Learning}

\begin{wrapfigure}{l}{0.29\textwidth}
    \centering
    \includegraphics[width=1.0\linewidth]{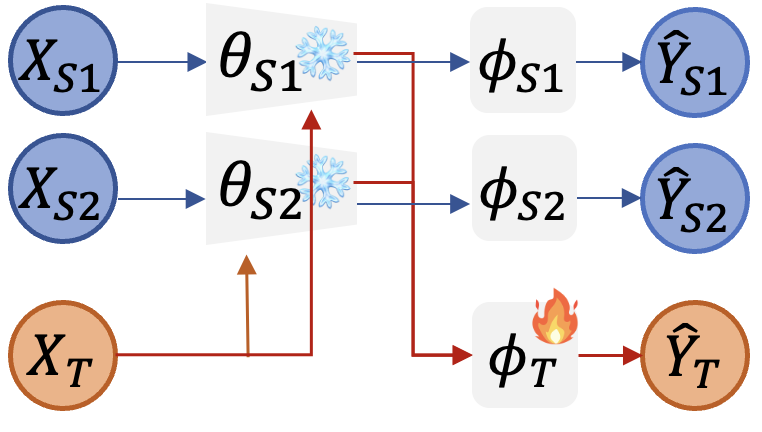}
    \caption{Multi-source transfer problem.}
    \label{fig:multisource-real}
\end{wrapfigure}

Multi-source transfer learning (as illustrated in Fig.~\ref{fig:multisource-real}) aims to leverage knowledge from multiple sources to boost the learning of target task \citep{sun2015survey}. Transferring using multiple sources is a long-proposed idea. Elementary works on this topic include naive multi-source transfer by concatenating all source feature extractors together for finetuning \citep{christodoulidis2016multisource} and theoretical works as in \citep{crammer2008learning, mansour2008domain, ben2010theory}.

With the development of this area, later works have been adopting different strategies to distinctively deal with different sources, incorporating transferability as an essential perspective when measuring the source importance. Source selection methods adopt multi-source transferability metrics to dynamically elect the source model emsemble that optimizes transferability and finetune it for transfer \citep{agostinelli2022transferability}. Re-weighting approaches linearly combine the sources, assigning source specific weights proportional to their importance in transfer. Specifically, MCW \citep{lee2019learning} interprets the features in a neural network as maximal correlation functions, viewing the maximized correlations of their feature functions on target data as an indicator for transferability and weighting the sources accordingly. Tong et al. \citep{tong2021mathematical} developed a mathematical framework to analyse and quantify transferability with high interpretability, leading to an iterative algorithm to optimize the source combination for multi-source transfer learning. DATE \citep{han2023discriminability} tackles multi-source-free domain adaptation by balancing discriminability and transferability through weight adjustments in source model ensembles, measuring transferability via a source-similarity perception module. CAiDA \citep{dong2021confident} boosts transfer by comprehensively utilizes pseudo label generation, class-relationship consistency and entropy based source-specific transferable perception. H-ensemble \citep{wu2024h} formulates multi-source transfer and source re-weighting into a maximal correlation regression framework and a transferability optimization problem respectively, where H-score transferability is adapted to suit the multi-source setting. Li et al. \citep{li2024agile} used attention mechanism to measure feature-output and feature-feature similarity, learning both intra-domain weights for resolving transferability mismatch and inter-domain ensemble weights for integrating source outputs into the final prediction.

\subsection{Meta Learning}
\begin{wrapfigure}{r}{0.25\textwidth}
    \centering
    \includegraphics[width=1.0\linewidth]{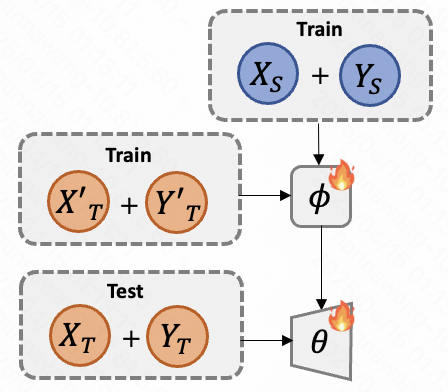}
    \caption{Meta learning problem.}
    \label{fig:multisource}
\end{wrapfigure}

Meta learning, often known as "learning to learn", aims to optimize the learning process itself across multiple tasks, enabling rapid adaptation to new, unseen tasks \cite{sun2020meta}. 

The incorporation of transferability metrics into meta-learning processes has emerged as a promising approach, particularly for model or task selection. This integration has shown remarkable potential in enhancing overall learning effectiveness. 
The current landscape of transferability measures in meta-learning can be broadly categorized into two main approaches: model-based and feature-based. 
Model-based transferability approaches focus on utilizing transferability scores to guide the selection of models or tasks during the meta-learning process. This strategy has proven particularly effective in improving learning outcomes across diverse domains. 

Complementing these model-based approaches, feature-based methods have also made substantial contributions by focusing on analyzing task relationships through feature representations. A notable example is TASK2VEC, introduced by \cite{achille2019task2vec}, which presents a novel task embedding approach for meta-learning. By constructing task embeddings using the Fisher Information Matrix (FIM) derived from pre-trained network feature maps, TASK2VEC offers a robust mechanism for analyzing task relationships and guiding knowledge transfer. This approach has opened new avenues for understanding the underlying structure of tasks in meta-learning contexts.

Advancing the field further, Tan et al. \cite{tan2024transferability} proposed two innovative transferability metrics: F-OTCE (Fast Optimal Transport based Conditional Entropy) and JC-OTCE (Joint Correspondence OTCE). These metrics leverage the principles of optimal transport theory to measure the alignment between source and target data distributions. By doing so, they provide a comprehensive framework for evaluating transferability across both domains and tasks, addressing the complex challenges of cross-domain and cross-task learning scenarios.

\subsection{Domain Generalization}
\begin{wrapfigure}{l}{0.3\textwidth}
    \centering
    \includegraphics[width=1.0\linewidth]{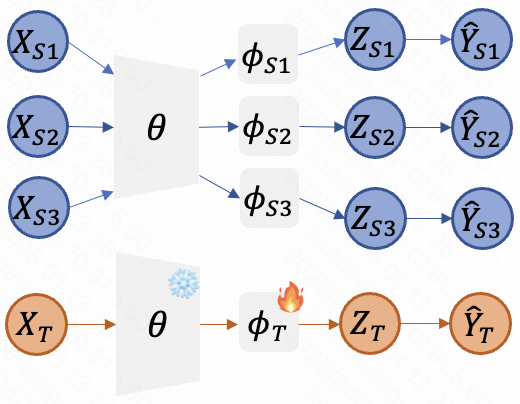}
    \caption{Domain generalization problem.}
    \label{fig:DG}
\end{wrapfigure}

Domain generalization seeks to train a model on multiple source domains, enabling it to perform well on a new target domain with a different data distribution, without requiring any target domain data during training. The primary objective in this setting is to learn a feature representation that is invariant to domain shifts while remaining discriminative for the task.
Zhang et al. \cite{zhang2021quantifying} provides a theoretical understanding of quantifying the transferability for domain generalization. By picking the worst-case pairwise domains among all pairs of domains (including the target domain) that have the largest error gap, the transferability reflects the domain gap among all domains.
Based on this, an algorithm for learning transferable features is proposed, focusing on learning better feature representations between non-transferable pairwise domains among all pairs, thus improving the transferability among all domains.
Similarly, some other works also apply minimax optimization learning paradigm to learn generalizable representations by improving the worst-case performance among all source domains \cite{sinha2018certifiable,sagawadistributionally,wang2024generalizing}.

To obtain the domain-invariant feature representation, while mitigating the negative effect of the distant source domains to keep more universal source knowledge, Shi et al. \cite{shi2022domain} leverages the concept of domain transferability to adjust the contributions of different source domains in the sample-level. In the context of information theory, entropy serves as a measure of uncertainty, with a higher entropy value indicating higher transferability. 
Thus, the entropy of the domain-conditional probability distribution effectively quantifies the transferability scores of samples from various source domains. The average transferability score of each batch of samples characterizes the transferability of their respective source domain. Utilizing this sample-level, information theory-based transferability metric, the model's generalization ability is enhanced by simultaneously learning a robust feature extractor by assigning greater weight to source domains with more universal domain knowledge in an online manner.

\subsection{Knowledge Distillation}

\begin{wrapfigure}{r}{0.27\textwidth}
    \centering
    \includegraphics[width=1.0\linewidth]{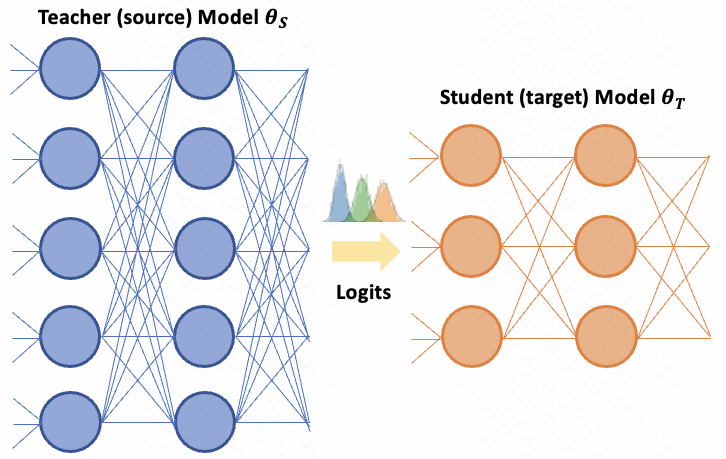}
    \caption{Knowledge distillation.}
    \label{fig:enter-label}
\end{wrapfigure}

Knowledge distillation is a form of model-to-model transfer learning, where a smaller, more efficient student model learns from a larger, more complex teacher model. A critical aspect of this process is the transferability between the teacher and student models. Transferability refers to the extent to which knowledge from the teacher model can be effectively transferred to and utilized by the student model. However, knowledge distillation often struggles with transferability due to mismatches between teacher and student models \cite{wang2021knowledge}. To address this, Lin et al. \cite{lin2022knowledge} introduces a “one-to-all” spatial matching approach using a target-aware transformer, which enhances transferability by employing parametric correlations to measure semantic distance, ensuring effective knowledge transfer despite architectural differences between the models. 

Transferability estimation methods can address this challenge by evaluating the transferability of different teacher-student model pairs, allowing researchers to make informed decisions on model selection and hyperparameter tuning, thus reducing the need for extensive training resources. By applying transferability estimation methods into knowledge distillation, researchers can effectively assess compatibility between teacher and student models, leading to efficient and accurate model compression techniques and enabling the deployment of small and fast models without sacrificing performance.

\subsection{Continual Learning}
\begin{wrapfigure}{l}{0.32\textwidth}
    \centering
    \includegraphics[width=1.0\linewidth]{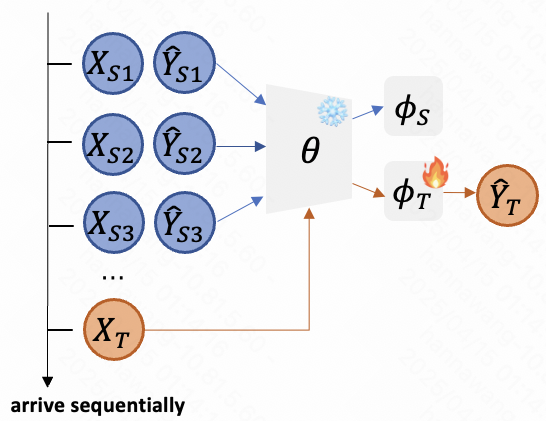}
    \caption{Continual learning problem.}
    \label{fig:enter-label}
\end{wrapfigure}

 Catastrophic forgetting is a critical issue in fine-tuning neural networks, especially with limited labeled data. This problem leads to representational collapse, where generalizable features degrade during fine-tuning \cite{aghajanyan2021intrinsic}. To mitigate this, strategies like using a small learning rate or freezing certain layers can help, but they risk trapping the model in local minima, especially when there is a large gap between pre-training and downstream task parameters \cite{kirkpatrick2017overcoming}. Research shows that the transferability of different layers varies, with early layers learning general features and later layers learning task-specific ones \cite{yosinski2014transferable}. To retain pre-training knowledge while adapting to the target task, early layers should preserve pre-trained knowledge, while later layers should adapt to the new task. Techniques like ULMFiT and DAN adjust learning rates for different layers to optimize this process \cite{howard2018universal,long2015learning}.

Another approach is Domain Adaptive Tuning, which addresses domain shifts between pre-training and the target task. Methods such as ULMFiT \cite{howard2018universal} and DAPT \cite{gururangan2020don} first fine-tune on data related to the target domain, then fine-tune again on the target task. Pre-training on unsupervised tasks provides valuable information about the target domain distribution, reducing labeling costs. Methods like SiATL also add auxiliary losses to alleviate catastrophic forgetting by learning task-specific features \cite{chronopoulou2019embarrassingly}. Additionally, Regularization Tuning prevents models from straying far from pre-trained parameters. Elastic Weight Consolidation (EWC) limits weight changes between pre-trained and fine-tuned models to combat forgetting \cite{kirkpatrick2017overcoming}. Techniques like Learning Without Forgetting (LWF) constrain the output layer’s response to ensure consistency during fine-tuning \cite{li2017learning}.

\subsection{Task Curriculum Learning}

\begin{wrapfigure}{r}{0.31\textwidth}
    \centering
    \includegraphics[width=1.0\linewidth]{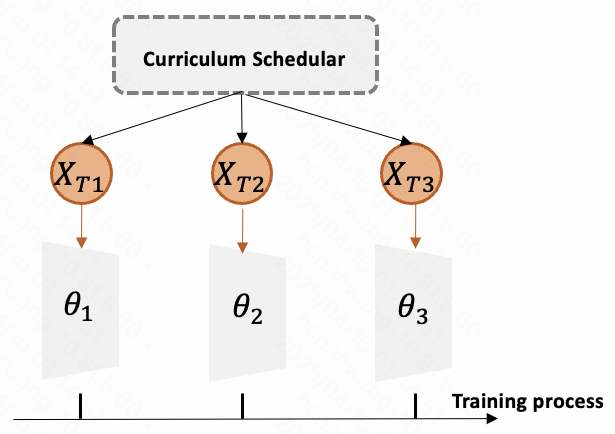}
    \caption{Task curriculum learning.}
    \label{fig:CL}
\end{wrapfigure}

Curriculum learning is the training strategy that trains a machine learning model in a meaningful order.
This learning paradigm aims to determine the best order of tasks that should be learned to maximize performance on the target task \cite{soviany2022curriculum}.
By measuring how effectively knowledge from related tasks can be transferred to the target task, transferability serves as a guiding principle in sequencing the learning steps to enhance model performance.
Gradient-based curriculum learning methods, for example, utilize gradient similarity \cite {du2018adapting,shi2020auxiliary} or gradient magnitude similarity \cite{yu2020gradient} between the target task and related tasks. These methods adaptively assign weights to different tasks based on how similar their gradients are, ensuring that tasks that share more relevant information with the target task are prioritized during training.
On the other hand, loss-based methods assume that the loss function indicates the significance of each sample or task for the target task.
Chen et al.\cite{chen2022auxiliary} propose using the correctness of a label as a criterion to evaluate whether a sample is beneficial to the target task. 
Mindermann et al. \cite{mindermann2022prioritized} propose the reducible holdout loss to identify data points that most enhance the target task, thereby improving the learning process.
Curriculum learning, combined with transferability, strategically orders tasks and adapts learning steps based on their relevance to the target task, resulting in more efficient and effective training.
Similarly, active learning seeks to select the most valuable instances for training, focusing on optimizing learning efficiency by querying the most informative data points. In a multi-domain setting, akin to curriculum learning's task-ordering approach, AL identifies instances across domains that contribute the most to the target task, helping transfer knowledge effectively between domains. 
For instance, Jiang et al. \cite{jiang2007instance} proposes a method which actively removes misleading instances where the target and source distributions differ, ensuring only useful knowledge is transferred. Dat et al. \cite{dai2007boosting} assigned weights to source data to balance their contribution, transferring relevant data to optimize the target domain's performance.
In both curriculum and active learning, the goal is to prioritize learning steps or instances that maximize the target task's performance, they ensure that only the most relevant knowledge is transferred to the target task, resulting in more effective and efficient learning.

\begin{tcolorbox}[breakable,enhanced,title={\bf \em Summary},left=2mm,right=2mm,fonttitle=\small,fontupper=\small, fontlower=\small]
The application of transferability metrics varies across different transfer learning paradigms:
\begin{itemize}[leftmargin=*]
\item {\bf Single-task Model Transfer} primarily utilizes {\em model transferability} metrics. These methods help predict how well a pretrained model will perform on a new target task without requiring full retraining. 
\item {\bf Domain Adaptation}, {\bf Multi-source Transfer} and {\bf Domain Generalization} commonly employs {\em dataset transferablity} metrics, which assists in weighting and selecting the most useful source domains or tasks. 
These metrics quantify domain shifts and identify which source domains or samples align best with the target distribution, and help in learning domain-invariant features robust to distributional shifts. 
\item {\bf Meta Learning} and {\bf Knowledge Distillation} relies on {\em model transferability} metrics to guide the selection of tasks for meta-training or evaluate compatibility between teacher and student models in the distillation process. 
\item {\bf Continual Learning} amd {\bf Task Curriculum Learning} benefits from {\em model transferability} metrics to preserve reusable knowledge while preventing catastrophic forgetting. These metrics guide which parts of the model to retain or adapt and prioritizes tasks that transfer useful knowledge. 

\end{itemize}
\end{tcolorbox}


\section{Challenges, Opportunities and Connections to Broader Topics}
 \label{sec:sec6}

This section explores the emerging challenges and opportunities associated with transferability, particularly in sequential transfer settings and large-scale foundation models. It also examines how transferability estimation intersects with broader topics such as model editing, prompt-based adaptation, and adversarial robustness. 
In particular, we highlight the role of transferability in real-world domains such as medical imaging and remote sensing.
Furthermore, transferability research shares deep connections with adjacent areas such as out-of-distribution (OOD) detection, uncertainty estimation, and causal learning. 
We provide understandings to the interdependencies between transferability and relevant topic.

\subsection{Challenges and Opportunities}
\subsubsection{Transferability for Sequential Transfer} 

Transferability plays a crucial role in both sequential transfer and active domain adaptation, particularly when there exists a large domain gap between the source and target tasks. In these scenarios, transferability metrics guide the learning process by informing what to transfer, how to transfer, and in what order. Both paradigms involve selecting data or domains strategically, often under resource constraints, and thus require careful planning based on domain similarity and potential knowledge reuse, which can be guided by transferability.

Sequential trensferability is the transferability of sequential transfer, which is also known as incremental transfer or continuous domain adaptation. Recent works include \cite{WEI2020106420} which proposes a novel multi-domain adaptation method for object detection based on incremental learning, and \cite{liu2024enhancingcontinuousdomainadaptation} which proposes a method W-MPOT enhancing continuous domain adaptation with multi-path transfer curriculum framework based on Wasserstein distance. These work mostly utilizes a source model to be transfered to a target model or task via one to more intermediate tasks.
In active domain adaptation, transferability metrics assess the relevance of source domain knowledge to target domains, helping select the most informative data samples, domains, or models for adaptation. By quantifying task similarity, researchers can prioritize adaptation strategies that maximize performance gains in the target domain. This approach is particularly useful in settings with limited labeled data in the target domain, where efficient transfer from the source domain is critical.
Despite the current work on the transferability studied within the field of sequential transfer and active domain adaptation, there exists several open challenges to be considered.

\noindent{\bf Introducing intermediate samples, datasets or domains.} 
The first challenge 
    is how to decide when the intermediate sample, dataset or domain can help the training adapt better to the target task, in other words, how to calculate the final transferability of using intermediate knowlegde and not using it, e.g. $Trf(A \to C)$ and $Trf(A\to B\to C)$ where $A$ is the source task, $B$ is the intermediate task and $C$ is the target task. Studies have found that this sequential transferability is related to task sample size, task complexity, and distribution distance. 

\noindent{\bf Optimal number of intermediate samples, datasets or domains.} 
The second challenge is to decide how many intermediate samples, datasets or domains is optimal for learning a given task. Previous work have tried to find a best number of intermediate samples, datasets or domains, but remains to be a theoretical study which lacks universality and is hard to realize.

\subsubsection{Transferability for Large-scale Foundation Models}
With the rise of large-scale models like BERT \cite{devlin2019bert} and GPT \cite{ouyang2022training} in large language models (LLM), Vision Transformer (ViT) \cite{yuan2021tokens} for vision usage, and CLIP \cite{radford2021learning} for vision-Language Models applications, transferability measurement has emerged as a crucial tool.


\noindent{\bf Transferability guides the update direction of model editing.}  
    For model editing, transferability measurement is crucial in identifying which parameters or layers of a model contain specific knowledge that can be edited or transferred without disrupting other functionalities. Methods like ROME, MEMIT, and MEND \cite{yao2023editinglargelanguagemodels} leverage such measurements to perform targeted updates in large-scale models like T5-XL and GPT-J. These techniques use transferability insights to refine editing strategies, enabling simultaneous edits across multiple knowledge points while preserving the model’s overall integrity and performance.

\noindent{\bf Prompt transferability improves efficiency of LLM adaptation.} 
    In the context of LLMs, transferability measurement helps evaluate the effectiveness of transferring learned prompts between tasks. The PANDA framework \cite{zhong2024pandaprompttransfermeets} measures prompt transferability by constructing a semantic space of tasks and using cosine similarity between task embeddings to predict the suitability of source-task prompts for target tasks. This approach significantly enhances the efficiency of parameter-efficient training methods like prompt-tuning, allowing researchers to reuse and adapt prompts with reduced computational overhead. Transferability metrics can optimize the initialization of prompts and improve knowledge transfer, addressing challenges like sensitivity to initialization and performance drops with smaller model scales.
    Su et al. \cite{su-etal-2022-transferability} finds that soft prompts can effectively transfer to similar tasks and other models via prompt projectors, which accelerates PT and improves performance. They also highlight that prompt transferability is strongly linked to how prompts stimulate the PLM’s internal neurons, suggesting that future research should focus on the interaction between prompts and PLM activations for better transferability.

\noindent{\bf Adversarial transferability across large models reveals necessity of strong defenses.} 
Adversarial transferability refers to the ability of adversarial perturbations crafted for a source model to remain effective in fooling a different target model, typically across tasks or domains, by maximizing the target model's prediction loss \cite{gu2023survey}.
Han et al. \cite{han2023ot} focus on adversarial transferability in Vision-Language Pretraining (VLP) models. They propose the Set-level Guidance Attack (SGA), which boosts transferability by leveraging cross-modal interactions between image and text modalities. By optimizing adversarial examples using set-level alignment, SGA achieves up to 30\% higher attack success rates than prior methods in image-text retrieval tasks. The cross-modal guidance further ensures the effectiveness of adversarial examples in both white-box and black-box settings. Understanding adversarial transferability helps researchers grasp the complexities of attacks on ML models and develop stronger defenses against them.

\subsection{Transferability Estimation and Optimization in Trustworthy AI Applications}
Transferability plays a critical role in Trustworthy AI, as it enhances the generalization ability of models across various domains, making AI systems more robust and reliable. This is especially important in fields like medical imaging, remote sensing
, where data heterogeneity and domain shifts are prevalent.

\noindent{\bf Medical analysis.}
In medical AI, models need to adapt across diverse imaging modalities and patient populations. Traditional transferability methods often struggle due to the subtle local texture variations present in medical images, unlike the more distinct features in natural images. 
DataMap \cite{DataMap} provides a novel framework for estimating transferability between datasets in medical image classification, while a study on transferability estimation for medical image segmentation \cite{Yang2023} introduces a source-free framework evaluating class consistency and feature variety. In the field of clinical text classification, Chen et al. \cite{chen2022leveraging} introduced the Normalized Negative Conditional Entropy (NNCE) as a measure of task transferability. By ingeniously combining the Reptile meta-learning algorithm with NNCE, this method enables the selection of beneficial source tasks for low-resource targets. The result is enhanced meta-learning performance in scenarios where data is limited, a common challenge in many real-world applications. For medical image segmentation, Yang et al. \cite{yang2023transferability} introduced the Label-Free Transferability Metric (LFTM) for multi-source model adaptation. LFTM innovatively employs instance-level and domain-level transferability matrices for pseudo-label correction and model selection, achieving effective target domain adaptation without the need to access source data. 
Duan et al. \cite{duan2025transfer} introduced a pixel-level transfer risk
map, which quantifies transferability using LEEP for each pixel to calculate the potential risks of negative transfer in the task of medical segmantation.

\noindent{\bf Remote sensing.}
Remote sensing has been a central research serving critical applications such as urban planning, environmental monitoring, land mapping, and GIS data updates. However, effective deployment of segmentation models in real-world scenarios remains challenging due to disparities in data sources, complex scenes, radiometric and spectral uncertainties, atmospheric effects, and evolving sensor technologies~\cite{qin2023transferabilitylearningmodelssemantic}. These factors often lead to significant domain shifts, causing models trained on one dataset to perform poorly on others. Thus, measuring and improving transferability is essential for robust and scalable remote sensing systems.
Filippelli et al. \cite{filippelli_testing_2024} presented a framework for evaluating the temporal transferability of remote sensing models used for large-area monitoring, which combines temporal cross-validation with a multiscale assessment to identify biases in model predictions and small area estimates when applying models to new time periods. Qin et al. \cite{qin_transferability_nodate} developed a straightforward method to quantify the transferability of a model using the spectral indices as a medium without using target labels. Li et al. \cite{li_retrieval_2022} combined Maximum Mean Discrepancy (MMD) with Transfer Component Analysis (TCA) to mitigate domain shifts between MODIS and VIIRS surface reflectance datasets. \cite{wan_combining_2022} enhanced this paradigm by integrating TCA with support vector regression, enabling cross-species transfer of leaf nitrogen concentration assessment models.
Moreover, transferability optimization strategies focus on methods improving the model generalization ability, including crafting domain-invariant feature extractors, applying adversarial training strategies, or using other domain adaptation methods. These methods have shown promise in learning domain-invariant features, as demonstrated in crop yield prediction and land cover classification across regions~\cite{ma_bayesian_2022, martini_domain-adversarial_2021, wang_exploring_2022}.
\subsection{Connections to Relevant Topics}

\noindent{\bf OOD detection.}
Out-of-Distribution (OOD) detection refers to a model's ability to recognize data sample that deviates significantly from the training data distribution. In contrast, instance-level transferability refers to the ability of a dataset or model to transfer or adapt knowledge from one domain or task to another at the level of individual samples. The goal of OOD detection is to detect a single analmolous or novel sample, while the goal of instance-level transferability is to use sample-wise methods of metrics to calculate a dataset or model's ability to transfer. Though they are slightly different, instance-level transferability methods can be widely applied in OOD sample detection tasks since the instance-level transferability can represent the deviation of data from the training data.


\noindent{\bf Uncertainty estimation.}
Uncertainty refers to the lack of confidence for each output of a machine learning algorithm. Uncertainty estimation is particularly important for neural networks, which have an inclination towards overconfident predictions. Incorrect overconfident predictions can be harmful in critical use cases such as healthcare or autonomous vehicles. Model uncertainty estimates source model on target task uncertainty. Recent methods uses Bayesian view to measure model uncertainty, which a popular method Bayesian Information Criterion (BIC) uses a linear combination of the model's log-likelihood and a logrithmic sample size to select a model.
The main difference from the model uncertainty and model transferability is that uncertainty estimation mostly require source data information in order to learn a posterior distribution on the model space, while model transferability focuses more on the model generalzation ability and can be source-data free.



\noindent{\bf Causal learning.}
Causal learning and transferability measurement are two interconnected but distinct areas in machine learning that deal with how models generalize across different environments or domains. Causal learning focuses on identifying causal relationships that are invariant to domain changes or environmental variations. The goal is to uncover underlying cause-effect structures in data that remain stable across different contexts. 
In contrast, transferability measurement evaluates how well these invariant causal relationships can be applied to specific tasks in new domains. It is more concerned with the practical use of causal information, which assessing how effectively learned causal structures can solve problems in different settings. While causal learning is primarily focused on identifying the cause-effect relationships that remain consistent, transferability measurement aims to determine how well these relationships can generalize to specific tasks, often using metrics such as task accuracy or robustness across domains. 

\section{Conclusions} 
 \label{sec:sec7}

 

In this survey, we provided a comprehensive overview of transferability estimation methods and their application in transfer learning, focusing on the importance of understanding the theoretical foundations behind each metric to ensure the success and trustworthiness of AI systems. We presented a unified taxonomy of transferability metrics, categorized by knowledge types and measurement granularity, and analyzed their applicability across various learning paradigms. By offering insights into how these metrics perform under different conditions, we aim to guide researchers and practitioners in selecting the most suitable metrics for specific tasks, ultimately contributing to more efficient, reliable, and trustworthy AI systems. Finally, we discussed open challenges and future directions to further advance the application of transferability metrics in trustworthy transfer learning.

 


\section{Acknowledgement}

This work is supported in part by the Natural Science Foundation of China (Grant 62371270). 



\bibliographystyle{ACM-Reference-Format}
\bibliography{sample-base}


\begin{thebibliography}{110}


\ifx \showCODEN    \undefined \def \showCODEN     #1{\unskip}     \fi
\ifx \showISBNx    \undefined \def \showISBNx     #1{\unskip}     \fi
\ifx \showISBNxiii \undefined \def \showISBNxiii  #1{\unskip}     \fi
\ifx \showISSN     \undefined \def \showISSN      #1{\unskip}     \fi
\ifx \showLCCN     \undefined \def \showLCCN      #1{\unskip}     \fi
\ifx \shownote     \undefined \def \shownote      #1{#1}          \fi
\ifx \showarticletitle \undefined \def \showarticletitle #1{#1}   \fi
\ifx \showURL      \undefined \def \showURL       {\relax}        \fi
\providecommand\bibfield[2]{#2}
\providecommand\bibinfo[2]{#2}
\providecommand\natexlab[1]{#1}
\providecommand\showeprint[2][]{arXiv:#2}

\bibitem[Achiam et~al\mbox{.}(2023)]%
        {achiam2023gpt}
\bibfield{author}{\bibinfo{person}{Josh Achiam}, \bibinfo{person}{Steven Adler}, \bibinfo{person}{Sandhini Agarwal}, \bibinfo{person}{Lama Ahmad}, \bibinfo{person}{Ilge Akkaya}, \bibinfo{person}{Florencia~Leoni Aleman}, \bibinfo{person}{Diogo Almeida}, \bibinfo{person}{Janko Altenschmidt}, \bibinfo{person}{Sam Altman}, \bibinfo{person}{Shyamal Anadkat}, {et~al\mbox{.}}} \bibinfo{year}{2023}\natexlab{}.
\newblock \showarticletitle{Gpt-4 technical report}.
\newblock \bibinfo{journal}{\emph{arXiv preprint arXiv:2303.08774}} (\bibinfo{year}{2023}).
\newblock


\bibitem[Achille et~al\mbox{.}(2019)]%
        {achille2019task2vec}
\bibfield{author}{\bibinfo{person}{Alessandro Achille}, \bibinfo{person}{Michael Lam}, \bibinfo{person}{Rahul Tewari}, \bibinfo{person}{Avinash Ravichandran}, \bibinfo{person}{Subhransu Maji}, \bibinfo{person}{Charless~C Fowlkes}, \bibinfo{person}{Stefano Soatto}, {and} \bibinfo{person}{Pietro Perona}.} \bibinfo{year}{2019}\natexlab{}.
\newblock \showarticletitle{Task2vec: Task embedding for meta-learning}. In \bibinfo{booktitle}{\emph{Proceedings of the IEEE/CVF international conference on computer vision}}. \bibinfo{pages}{6430--6439}.
\newblock


\bibitem[Aghajanyan et~al\mbox{.}(2021)]%
        {aghajanyan2021intrinsic}
\bibfield{author}{\bibinfo{person}{Armen Aghajanyan}, \bibinfo{person}{Sonal Gupta}, {and} \bibinfo{person}{Luke Zettlemoyer}.} \bibinfo{year}{2021}\natexlab{}.
\newblock \showarticletitle{Intrinsic Dimensionality Explains the Effectiveness of Language Model Fine-Tuning}. In \bibinfo{booktitle}{\emph{Proceedings of the 59th Annual Meeting of the Association for Computational Linguistics and the 11th International Joint Conference on Natural Language Processing (Volume 1: Long Papers)}}. \bibinfo{pages}{7319--7328}.
\newblock


\bibitem[Agostinelli et~al\mbox{.}(2022a)]%
        {agostinelli2022stable}
\bibfield{author}{\bibinfo{person}{Andrea Agostinelli}, \bibinfo{person}{Michal P{\'a}ndy}, \bibinfo{person}{Jasper Uijlings}, \bibinfo{person}{Thomas Mensink}, {and} \bibinfo{person}{Vittorio Ferrari}.} \bibinfo{year}{2022}\natexlab{a}.
\newblock \showarticletitle{How stable are transferability metrics evaluations?}. In \bibinfo{booktitle}{\emph{European Conference on Computer Vision}}. Springer, \bibinfo{pages}{303--321}.
\newblock


\bibitem[Agostinelli et~al\mbox{.}(2022b)]%
        {agostinelli2022transferability}
\bibfield{author}{\bibinfo{person}{Andrea Agostinelli}, \bibinfo{person}{Jasper Uijlings}, \bibinfo{person}{Thomas Mensink}, {and} \bibinfo{person}{Vittorio Ferrari}.} \bibinfo{year}{2022}\natexlab{b}.
\newblock \showarticletitle{Transferability metrics for selecting source model ensembles}. In \bibinfo{booktitle}{\emph{Proceedings of the IEEE/CVF Conference on Computer Vision and Pattern Recognition}}. \bibinfo{pages}{7936--7946}.
\newblock


\bibitem[Alvarez-Melis and Fusi(2020)]%
        {DBLP:journals/corr/abs-2002-02923}
\bibfield{author}{\bibinfo{person}{David Alvarez-Melis} {and} \bibinfo{person}{Nicolo Fusi}.} \bibinfo{year}{2020}\natexlab{}.
\newblock \showarticletitle{Geometric dataset distances via optimal transport}.
\newblock \bibinfo{journal}{\emph{Advances in Neural Information Processing Systems}}  \bibinfo{volume}{33} (\bibinfo{year}{2020}), \bibinfo{pages}{21428--21439}.
\newblock


\bibitem[Bao et~al\mbox{.}(2019)]%
        {bao2022informationtheoreticapproachtransferabilitytask}
\bibfield{author}{\bibinfo{person}{Yajie Bao}, \bibinfo{person}{Yang Li}, \bibinfo{person}{Shao-Lun Huang}, \bibinfo{person}{Lin Zhang}, \bibinfo{person}{Lizhong Zheng}, \bibinfo{person}{Amir Zamir}, {and} \bibinfo{person}{Leonidas Guibas}.} \bibinfo{year}{2019}\natexlab{}.
\newblock \showarticletitle{An information-theoretic approach to transferability in task transfer learning}. In \bibinfo{booktitle}{\emph{2019 IEEE international conference on image processing (ICIP)}}. IEEE, \bibinfo{pages}{2309--2313}.
\newblock


\bibitem[Ben-David et~al\mbox{.}(2010)]%
        {ben2010theory}
\bibfield{author}{\bibinfo{person}{Shai Ben-David}, \bibinfo{person}{John Blitzer}, \bibinfo{person}{Koby Crammer}, \bibinfo{person}{Alex Kulesza}, \bibinfo{person}{Fernando Pereira}, {and} \bibinfo{person}{Jennifer~Wortman Vaughan}.} \bibinfo{year}{2010}\natexlab{}.
\newblock \showarticletitle{A theory of learning from different domains}.
\newblock \bibinfo{journal}{\emph{Machine learning}}  \bibinfo{volume}{79} (\bibinfo{year}{2010}), \bibinfo{pages}{151--175}.
\newblock


\bibitem[Ben-David et~al\mbox{.}(2006)]%
        {NIPS2006_b1b0432c}
\bibfield{author}{\bibinfo{person}{Shai Ben-David}, \bibinfo{person}{John Blitzer}, \bibinfo{person}{Koby Crammer}, {and} \bibinfo{person}{Fernando Pereira}.} \bibinfo{year}{2006}\natexlab{}.
\newblock \showarticletitle{Analysis of Representations for Domain Adaptation}. In \bibinfo{booktitle}{\emph{Advances in Neural Information Processing Systems}}, \bibfield{editor}{\bibinfo{person}{B.~Sch\"{o}lkopf}, \bibinfo{person}{J.~Platt}, {and} \bibinfo{person}{T.~Hoffman}} (Eds.), Vol.~\bibinfo{volume}{19}. \bibinfo{publisher}{MIT Press}.
\newblock


\bibitem[Chen et~al\mbox{.}(2022b)]%
        {chen2022auxiliary}
\bibfield{author}{\bibinfo{person}{Hong Chen}, \bibinfo{person}{Xin Wang}, \bibinfo{person}{Chaoyu Guan}, \bibinfo{person}{Yue Liu}, {and} \bibinfo{person}{Wenwu Zhu}.} \bibinfo{year}{2022}\natexlab{b}.
\newblock \showarticletitle{Auxiliary learning with joint task and data scheduling}. In \bibinfo{booktitle}{\emph{International Conference on Machine Learning}}. PMLR, \bibinfo{pages}{3634--3647}.
\newblock


\bibitem[Chen et~al\mbox{.}(2022a)]%
        {chen2022leveraging}
\bibfield{author}{\bibinfo{person}{Zhuohao Chen}, \bibinfo{person}{Jangwon Kim}, \bibinfo{person}{Ram Bhakta}, {and} \bibinfo{person}{Mustafa Sir}.} \bibinfo{year}{2022}\natexlab{a}.
\newblock \showarticletitle{Leveraging task transferability to meta-learning for clinical section classification with limited data}. In \bibinfo{booktitle}{\emph{Proceedings of the 60th Annual Meeting of the Association for Computational Linguistics (Volume 1: Long Papers)}}. \bibinfo{pages}{6690--6702}.
\newblock


\bibitem[Christodoulidis et~al\mbox{.}(2016)]%
        {christodoulidis2016multisource}
\bibfield{author}{\bibinfo{person}{Stergios Christodoulidis}, \bibinfo{person}{Marios Anthimopoulos}, \bibinfo{person}{Lukas Ebner}, \bibinfo{person}{Andreas Christe}, {and} \bibinfo{person}{Stavroula Mougiakakou}.} \bibinfo{year}{2016}\natexlab{}.
\newblock \showarticletitle{Multisource transfer learning with convolutional neural networks for lung pattern analysis}.
\newblock \bibinfo{journal}{\emph{IEEE journal of biomedical and health informatics}} \bibinfo{volume}{21}, \bibinfo{number}{1} (\bibinfo{year}{2016}), \bibinfo{pages}{76--84}.
\newblock


\bibitem[Chronopoulou et~al\mbox{.}(2019)]%
        {chronopoulou2019embarrassingly}
\bibfield{author}{\bibinfo{person}{Alexandra Chronopoulou}, \bibinfo{person}{Christos Baziotis}, {and} \bibinfo{person}{Alexandros Potamianos}.} \bibinfo{year}{2019}\natexlab{}.
\newblock \showarticletitle{An embarrassingly simple approach for transfer learning from pretrained language models}.
\newblock \bibinfo{journal}{\emph{arXiv preprint arXiv:1902.10547}} (\bibinfo{year}{2019}).
\newblock


\bibitem[Crammer et~al\mbox{.}(2008)]%
        {crammer2008learning}
\bibfield{author}{\bibinfo{person}{Koby Crammer}, \bibinfo{person}{Michael Kearns}, {and} \bibinfo{person}{Jennifer Wortman}.} \bibinfo{year}{2008}\natexlab{}.
\newblock \showarticletitle{Learning from Multiple Sources.}
\newblock \bibinfo{journal}{\emph{Journal of Machine Learning Research}} \bibinfo{volume}{9}, \bibinfo{number}{8} (\bibinfo{year}{2008}).
\newblock


\bibitem[Dai et~al\mbox{.}(2007)]%
        {dai2007boosting}
\bibfield{author}{\bibinfo{person}{Wenyuan Dai}, \bibinfo{person}{Qiang Yang}, \bibinfo{person}{Gui-Rong Xue}, {and} \bibinfo{person}{Yong Yu}.} \bibinfo{year}{2007}\natexlab{}.
\newblock \showarticletitle{Boosting for transfer learning}. In \bibinfo{booktitle}{\emph{Proceedings of the 24th international conference on Machine learning}}. \bibinfo{pages}{193--200}.
\newblock


\bibitem[Devlin et~al\mbox{.}(2019a)]%
        {2019BERT}
\bibfield{author}{\bibinfo{person}{Jacob Devlin}, \bibinfo{person}{Ming~Wei Chang}, \bibinfo{person}{Kenton Lee}, {and} \bibinfo{person}{Kristina Toutanova}.} \bibinfo{year}{2019}\natexlab{a}.
\newblock \showarticletitle{BERT: Pre-training of Deep Bidirectional Transformers for Language Understanding}. In \bibinfo{booktitle}{\emph{Conference of the North American Chapter of the Association for Computational Linguistics: Human Language Technologies, vol. 6. long and short papers: Conference of the North American Chapter of the Association for Computational Linguistics: Human Language Technologies (NAACL HLT 2019), 2-7 June 2019, Minneapolis, Minnesota, USA}}.
\newblock


\bibitem[Devlin et~al\mbox{.}(2019b)]%
        {devlin2019bert}
\bibfield{author}{\bibinfo{person}{Jacob Devlin}, \bibinfo{person}{Ming-Wei Chang}, \bibinfo{person}{Kenton Lee}, {and} \bibinfo{person}{Kristina Toutanova}.} \bibinfo{year}{2019}\natexlab{b}.
\newblock \showarticletitle{Bert: Pre-training of deep bidirectional transformers for language understanding}. In \bibinfo{booktitle}{\emph{Proceedings of the 2019 conference of the North American chapter of the association for computational linguistics: human language technologies, volume 1 (long and short papers)}}. \bibinfo{pages}{4171--4186}.
\newblock


\bibitem[Ding et~al\mbox{.}(2024)]%
        {ding2024modeltransfersurveytransferability}
\bibfield{author}{\bibinfo{person}{Yuhe Ding}, \bibinfo{person}{Bo Jiang}, \bibinfo{person}{Aijing Yu}, \bibinfo{person}{Aihua Zheng}, {and} \bibinfo{person}{Jian Liang}.} \bibinfo{year}{2024}\natexlab{}.
\newblock \showarticletitle{Which model to transfer? a survey on transferability estimation}.
\newblock \bibinfo{journal}{\emph{arXiv preprint arXiv:2402.15231}} (\bibinfo{year}{2024}).
\newblock


\bibitem[Dong et~al\mbox{.}(2021)]%
        {dong2021confident}
\bibfield{author}{\bibinfo{person}{Jiahua Dong}, \bibinfo{person}{Zhen Fang}, \bibinfo{person}{Anjin Liu}, \bibinfo{person}{Gan Sun}, {and} \bibinfo{person}{Tongliang Liu}.} \bibinfo{year}{2021}\natexlab{}.
\newblock \showarticletitle{Confident anchor-induced multi-source free domain adaptation}.
\newblock \bibinfo{journal}{\emph{Advances in Neural Information Processing Systems}}  \bibinfo{volume}{34} (\bibinfo{year}{2021}), \bibinfo{pages}{2848--2860}.
\newblock


\bibitem[Du et~al\mbox{.}(2024)]%
        {DataMap}
\bibfield{author}{\bibinfo{person}{Xiangtong Du}, \bibinfo{person}{Zhidong Liu}, \bibinfo{person}{Zunlei Feng}, {and} \bibinfo{person}{Hai Deng}.} \bibinfo{year}{2024}\natexlab{}.
\newblock \showarticletitle{DataMap: Dataset transferability map for medical image classification}.
\newblock \bibinfo{journal}{\emph{Pattern Recognition}}  \bibinfo{volume}{146} (\bibinfo{year}{2024}), \bibinfo{pages}{110044}.
\newblock


\bibitem[Du et~al\mbox{.}(2018)]%
        {du2018adapting}
\bibfield{author}{\bibinfo{person}{Yunshu Du}, \bibinfo{person}{Wojciech~M Czarnecki}, \bibinfo{person}{Siddhant~M Jayakumar}, \bibinfo{person}{Mehrdad Farajtabar}, \bibinfo{person}{Razvan Pascanu}, {and} \bibinfo{person}{Balaji Lakshminarayanan}.} \bibinfo{year}{2018}\natexlab{}.
\newblock \showarticletitle{Adapting auxiliary losses using gradient similarity}.
\newblock \bibinfo{journal}{\emph{arXiv preprint arXiv:1812.02224}} (\bibinfo{year}{2018}).
\newblock


\bibitem[Duan et~al\mbox{.}(2025)]%
        {duan2025transfer}
\bibfield{author}{\bibinfo{person}{Shutong Duan}, \bibinfo{person}{Jingyun Yang}, \bibinfo{person}{Yang Tan}, \bibinfo{person}{Guoqing Zhang}, \bibinfo{person}{Yang Li}, {and} \bibinfo{person}{Xiao-Ping Zhang}.} \bibinfo{year}{2025}\natexlab{}.
\newblock \showarticletitle{Transfer Risk Map: Mitigating Pixel-level Negative Transfer in Medical Segmentation}.
\newblock \bibinfo{journal}{\emph{arXiv preprint arXiv:2502.02340}} (\bibinfo{year}{2025}).
\newblock


\bibitem[Dwivedi et~al\mbox{.}(2020)]%
        {dwivedi2020dualitydiagramsimilaritygeneric}
\bibfield{author}{\bibinfo{person}{Kshitij Dwivedi}, \bibinfo{person}{Jiahui Huang}, \bibinfo{person}{Radoslaw~Martin Cichy}, {and} \bibinfo{person}{Gemma Roig}.} \bibinfo{year}{2020}\natexlab{}.
\newblock \showarticletitle{Duality diagram similarity: a generic framework for initialization selection in task transfer learning}. In \bibinfo{booktitle}{\emph{European Conference on Computer Vision}}. Springer, \bibinfo{pages}{497--513}.
\newblock


\bibitem[Dwivedi and Roig(2019)]%
        {dwivedi2019representationsimilarityanalysisefficient}
\bibfield{author}{\bibinfo{person}{Kshitij Dwivedi} {and} \bibinfo{person}{Gemma Roig}.} \bibinfo{year}{2019}\natexlab{}.
\newblock \showarticletitle{Representation similarity analysis for efficient task taxonomy \& transfer learning}. In \bibinfo{booktitle}{\emph{Proceedings of the IEEE/CVF Conference on Computer Vision and Pattern Recognition}}. \bibinfo{pages}{12387--12396}.
\newblock


\bibitem[Filippelli et~al\mbox{.}(2024)]%
        {filippelli_testing_2024}
\bibfield{author}{\bibinfo{person}{Steven~K. Filippelli}, \bibinfo{person}{Karen Schleeweis}, \bibinfo{person}{Mark~D. Nelson}, \bibinfo{person}{Patrick~A. Fekety}, {and} \bibinfo{person}{Jody~C. Vogeler}.} \bibinfo{year}{2024}\natexlab{}.
\newblock \showarticletitle{Testing temporal transferability of remote sensing models for large area monitoring}.
\newblock \bibinfo{journal}{\emph{Science of Remote Sensing}}  \bibinfo{volume}{9} (\bibinfo{date}{June} \bibinfo{year}{2024}), \bibinfo{pages}{100119}.
\newblock
\showISSN{26660172}


\bibitem[Fukumizu et~al\mbox{.}(2009)]%
        {fukumizu2009kernel}
\bibfield{author}{\bibinfo{person}{Kenji Fukumizu}, \bibinfo{person}{Arthur Gretton}, \bibinfo{person}{Gert Lanckriet}, \bibinfo{person}{Bernhard Sch{\"o}lkopf}, {and} \bibinfo{person}{Bharath~K Sriperumbudur}.} \bibinfo{year}{2009}\natexlab{}.
\newblock \showarticletitle{Kernel choice and classifiability for RKHS embeddings of probability distributions}.
\newblock \bibinfo{journal}{\emph{Advances in neural information processing systems}}  \bibinfo{volume}{22} (\bibinfo{year}{2009}).
\newblock


\bibitem[Ganin et~al\mbox{.}(2016)]%
        {ganin2016domain}
\bibfield{author}{\bibinfo{person}{Yaroslav Ganin}, \bibinfo{person}{Evgeniya Ustinova}, \bibinfo{person}{Hana Ajakan}, \bibinfo{person}{Pascal Germain}, \bibinfo{person}{Hugo Larochelle}, \bibinfo{person}{Fran{\c{c}}ois Laviolette}, \bibinfo{person}{Mario March}, {and} \bibinfo{person}{Victor Lempitsky}.} \bibinfo{year}{2016}\natexlab{}.
\newblock \showarticletitle{Domain-adversarial training of neural networks}.
\newblock \bibinfo{journal}{\emph{Journal of machine learning research}} \bibinfo{volume}{17}, \bibinfo{number}{59} (\bibinfo{year}{2016}), \bibinfo{pages}{1--35}.
\newblock


\bibitem[Gu et~al\mbox{.}(2023)]%
        {gu2023survey}
\bibfield{author}{\bibinfo{person}{Jindong Gu}, \bibinfo{person}{Xiaojun Jia}, \bibinfo{person}{Pau de Jorge}, \bibinfo{person}{Wenqain Yu}, \bibinfo{person}{Xinwei Liu}, \bibinfo{person}{Avery Ma}, \bibinfo{person}{Yuan Xun}, \bibinfo{person}{Anjun Hu}, \bibinfo{person}{Ashkan Khakzar}, \bibinfo{person}{Zhijiang Li}, {et~al\mbox{.}}} \bibinfo{year}{2023}\natexlab{}.
\newblock \showarticletitle{A survey on transferability of adversarial examples across deep neural networks}.
\newblock \bibinfo{journal}{\emph{arXiv preprint arXiv:2310.17626}} (\bibinfo{year}{2023}).
\newblock


\bibitem[Gururangan et~al\mbox{.}(2020)]%
        {gururangan2020don}
\bibfield{author}{\bibinfo{person}{Suchin Gururangan}, \bibinfo{person}{Ana Marasovi{\'c}}, \bibinfo{person}{Swabha Swayamdipta}, \bibinfo{person}{Kyle Lo}, \bibinfo{person}{Iz Beltagy}, \bibinfo{person}{Doug Downey}, {and} \bibinfo{person}{Noah~A Smith}.} \bibinfo{year}{2020}\natexlab{}.
\newblock \showarticletitle{Don't stop pretraining: Adapt language models to domains and tasks}.
\newblock \bibinfo{journal}{\emph{arXiv preprint arXiv:2004.10964}} (\bibinfo{year}{2020}).
\newblock


\bibitem[Han et~al\mbox{.}(2023a)]%
        {han2023ot}
\bibfield{author}{\bibinfo{person}{Dongchen Han}, \bibinfo{person}{Xiaojun Jia}, \bibinfo{person}{Yang Bai}, \bibinfo{person}{Jindong Gu}, \bibinfo{person}{Yang Liu}, {and} \bibinfo{person}{Xiaochun Cao}.} \bibinfo{year}{2023}\natexlab{a}.
\newblock \showarticletitle{Ot-attack: Enhancing adversarial transferability of vision-language models via optimal transport optimization}.
\newblock \bibinfo{journal}{\emph{arXiv preprint arXiv:2312.04403}} (\bibinfo{year}{2023}).
\newblock


\bibitem[Han et~al\mbox{.}(2023b)]%
        {han2023discriminability}
\bibfield{author}{\bibinfo{person}{Zhongyi Han}, \bibinfo{person}{Zhiyan Zhang}, \bibinfo{person}{Fan Wang}, \bibinfo{person}{Rundong He}, \bibinfo{person}{Wan Su}, \bibinfo{person}{Xiaoming Xi}, {and} \bibinfo{person}{Yilong Yin}.} \bibinfo{year}{2023}\natexlab{b}.
\newblock \showarticletitle{Discriminability and Transferability Estimation: A Bayesian Source Importance Estimation Approach for Multi-Source-Free Domain Adaptation}. In \bibinfo{booktitle}{\emph{Proceedings of the AAAI Conference on Artificial Intelligence}}, Vol.~\bibinfo{volume}{37}. \bibinfo{pages}{7811--7820}.
\newblock


\bibitem[Hendrycks and Gimpel(2017)]%
        {hendrycks2022baseline}
\bibfield{author}{\bibinfo{person}{Dan Hendrycks} {and} \bibinfo{person}{Kevin Gimpel}.} \bibinfo{year}{2017}\natexlab{}.
\newblock \showarticletitle{A Baseline for Detecting Misclassified and Out-of-Distribution Examples in Neural Networks}. In \bibinfo{booktitle}{\emph{International Conference on Learning Representations}}.
\newblock


\bibitem[Howard and Ruder(2018)]%
        {howard2018universal}
\bibfield{author}{\bibinfo{person}{Jeremy Howard} {and} \bibinfo{person}{Sebastian Ruder}.} \bibinfo{year}{2018}\natexlab{}.
\newblock \showarticletitle{Universal language model fine-tuning for text classification}.
\newblock \bibinfo{journal}{\emph{arXiv preprint arXiv:1801.06146}} (\bibinfo{year}{2018}).
\newblock


\bibitem[Huang et~al\mbox{.}(2022)]%
        {huang2022balancing}
\bibfield{author}{\bibinfo{person}{Jingke Huang}, \bibinfo{person}{Ni Xiao}, {and} \bibinfo{person}{Lei Zhang}.} \bibinfo{year}{2022}\natexlab{}.
\newblock \showarticletitle{Balancing transferability and discriminability for unsupervised domain adaptation}.
\newblock \bibinfo{journal}{\emph{IEEE Transactions on Neural Networks and Learning Systems}} \bibinfo{volume}{35}, \bibinfo{number}{4} (\bibinfo{year}{2022}), \bibinfo{pages}{5807--5814}.
\newblock


\bibitem[Jiang et~al\mbox{.}(2022)]%
        {jiang2022transferabilitydeeplearningsurvey}
\bibfield{author}{\bibinfo{person}{Junguang Jiang}, \bibinfo{person}{Yang Shu}, \bibinfo{person}{Jianmin Wang}, {and} \bibinfo{person}{Mingsheng Long}.} \bibinfo{year}{2022}\natexlab{}.
\newblock \showarticletitle{Transferability in deep learning: A survey}.
\newblock \bibinfo{journal}{\emph{arXiv preprint arXiv:2201.05867}} (\bibinfo{year}{2022}).
\newblock


\bibitem[Jiang and Zhai(2007)]%
        {jiang2007instance}
\bibfield{author}{\bibinfo{person}{Jing Jiang} {and} \bibinfo{person}{ChengXiang Zhai}.} \bibinfo{year}{2007}\natexlab{}.
\newblock \showarticletitle{Instance Weighting for Domain Adaptation in NLP}. In \bibinfo{booktitle}{\emph{Proceedings of the 45th Annual Meeting of the Association of Computational Linguistics}}. \bibinfo{pages}{264--271}.
\newblock


\bibitem[Kalhor et~al\mbox{.}(2020)]%
        {kalhor2020rankingrejectingpretraineddeep}
\bibfield{author}{\bibinfo{person}{Mostafa Kalhor}, \bibinfo{person}{Ahmad Kalhor}, {and} \bibinfo{person}{Mehdi Rahmani}.} \bibinfo{year}{2020}\natexlab{}.
\newblock \showarticletitle{Ranking and rejecting of pre-trained deep neural networks in transfer learning based on separation index}.
\newblock \bibinfo{journal}{\emph{arXiv preprint arXiv:2012.13717}} (\bibinfo{year}{2020}).
\newblock


\bibitem[Kantorovich(2006)]%
        {kantorovich2006translocation}
\bibfield{author}{\bibinfo{person}{Leonid~V Kantorovich}.} \bibinfo{year}{2006}\natexlab{}.
\newblock \showarticletitle{On the Translocation of Masses.}
\newblock \bibinfo{journal}{\emph{Journal of mathematical sciences}} \bibinfo{volume}{133}, \bibinfo{number}{4} (\bibinfo{year}{2006}).
\newblock


\bibitem[Kirkpatrick et~al\mbox{.}(2017)]%
        {kirkpatrick2017overcoming}
\bibfield{author}{\bibinfo{person}{James Kirkpatrick}, \bibinfo{person}{Razvan Pascanu}, \bibinfo{person}{Neil Rabinowitz}, \bibinfo{person}{Joel Veness}, \bibinfo{person}{Guillaume Desjardins}, \bibinfo{person}{Andrei~A Rusu}, \bibinfo{person}{Kieran Milan}, \bibinfo{person}{John Quan}, \bibinfo{person}{Tiago Ramalho}, \bibinfo{person}{Agnieszka Grabska-Barwinska}, {et~al\mbox{.}}} \bibinfo{year}{2017}\natexlab{}.
\newblock \showarticletitle{Overcoming catastrophic forgetting in neural networks}.
\newblock \bibinfo{journal}{\emph{Proceedings of the national academy of sciences}} \bibinfo{volume}{114}, \bibinfo{number}{13} (\bibinfo{year}{2017}), \bibinfo{pages}{3521--3526}.
\newblock


\bibitem[Lee et~al\mbox{.}(2019)]%
        {lee2019learning}
\bibfield{author}{\bibinfo{person}{Joshua Lee}, \bibinfo{person}{Prasanna Sattigeri}, {and} \bibinfo{person}{Gregory Wornell}.} \bibinfo{year}{2019}\natexlab{}.
\newblock \showarticletitle{Learning new tricks from old dogs: Multi-source transfer learning from pre-trained networks}.
\newblock \bibinfo{journal}{\emph{Advances in neural information processing systems}}  \bibinfo{volume}{32} (\bibinfo{year}{2019}).
\newblock


\bibitem[Lee et~al\mbox{.}(2018)]%
        {NEURIPS2018_abdeb6f5}
\bibfield{author}{\bibinfo{person}{Kimin Lee}, \bibinfo{person}{Kibok Lee}, \bibinfo{person}{Honglak Lee}, {and} \bibinfo{person}{Jinwoo Shin}.} \bibinfo{year}{2018}\natexlab{}.
\newblock \showarticletitle{A Simple Unified Framework for Detecting Out-of-Distribution Samples and Adversarial Attacks}. In \bibinfo{booktitle}{\emph{Advances in Neural Information Processing Systems}}, \bibfield{editor}{\bibinfo{person}{S.~Bengio}, \bibinfo{person}{H.~Wallach}, \bibinfo{person}{H.~Larochelle}, \bibinfo{person}{K.~Grauman}, \bibinfo{person}{N.~Cesa-Bianchi}, {and} \bibinfo{person}{R.~Garnett}} (Eds.), Vol.~\bibinfo{volume}{31}. \bibinfo{publisher}{Curran Associates, Inc.}
\newblock


\bibitem[Li et~al\mbox{.}(2022a)]%
        {li2022enhancing}
\bibfield{author}{\bibinfo{person}{Jingyao Li}, \bibinfo{person}{Shuai L{\"u}}, \bibinfo{person}{Wenbo Zhu}, {and} \bibinfo{person}{Zhanshan Li}.} \bibinfo{year}{2022}\natexlab{a}.
\newblock \showarticletitle{Enhancing transferability and discriminability simultaneously for unsupervised domain adaptation}.
\newblock \bibinfo{journal}{\emph{Knowledge-Based Systems}}  \bibinfo{volume}{247} (\bibinfo{year}{2022}), \bibinfo{pages}{108705}.
\newblock


\bibitem[Li et~al\mbox{.}(2022b)]%
        {li_retrieval_2022}
\bibfield{author}{\bibinfo{person}{Juan Li}, \bibinfo{person}{Zhiqiang Xiao}, \bibinfo{person}{Rui Sun}, {and} \bibinfo{person}{Jinling Song}.} \bibinfo{year}{2022}\natexlab{b}.
\newblock \showarticletitle{Retrieval of the {Leaf} {Area} {Index} from {Visible} {Infrared} {Imaging} {Radiometer} {Suite} ({VIIRS}) {Surface} {Reflectance} {Based} on {Unsupervised} {Domain} {Adaptation}}.
\newblock \bibinfo{journal}{\emph{Remote Sensing}} \bibinfo{volume}{14}, \bibinfo{number}{8} (\bibinfo{date}{April} \bibinfo{year}{2022}), \bibinfo{pages}{1826}.
\newblock
\showISSN{2072-4292}


\bibitem[Li et~al\mbox{.}(2024)]%
        {li2024agile}
\bibfield{author}{\bibinfo{person}{Xinyao Li}, \bibinfo{person}{Jingjing Li}, \bibinfo{person}{Fengling Li}, \bibinfo{person}{Lei Zhu}, {and} \bibinfo{person}{Ke Lu}.} \bibinfo{year}{2024}\natexlab{}.
\newblock \showarticletitle{Agile Multi-Source-Free Domain Adaptation}. In \bibinfo{booktitle}{\emph{Proceedings of the AAAI Conference on Artificial Intelligence}}, Vol.~\bibinfo{volume}{38}. \bibinfo{pages}{13673--13681}.
\newblock


\bibitem[Li and Hoiem(2017)]%
        {li2017learning}
\bibfield{author}{\bibinfo{person}{Zhizhong Li} {and} \bibinfo{person}{Derek Hoiem}.} \bibinfo{year}{2017}\natexlab{}.
\newblock \showarticletitle{Learning without forgetting}.
\newblock \bibinfo{journal}{\emph{IEEE transactions on pattern analysis and machine intelligence}} \bibinfo{volume}{40}, \bibinfo{number}{12} (\bibinfo{year}{2017}), \bibinfo{pages}{2935--2947}.
\newblock


\bibitem[Liang et~al\mbox{.}(2017)]%
        {liang2017enhancing}
\bibfield{author}{\bibinfo{person}{Shiyu Liang}, \bibinfo{person}{Yixuan Li}, {and} \bibinfo{person}{Rayadurgam Srikant}.} \bibinfo{year}{2017}\natexlab{}.
\newblock \showarticletitle{Enhancing the reliability of out-of-distribution image detection in neural networks}.
\newblock \bibinfo{journal}{\emph{arXiv preprint arXiv:1706.02690}} (\bibinfo{year}{2017}).
\newblock


\bibitem[Lin et~al\mbox{.}(2022)]%
        {lin2022knowledge}
\bibfield{author}{\bibinfo{person}{Sihao Lin}, \bibinfo{person}{Hongwei Xie}, \bibinfo{person}{Bing Wang}, \bibinfo{person}{Kaicheng Yu}, \bibinfo{person}{Xiaojun Chang}, \bibinfo{person}{Xiaodan Liang}, {and} \bibinfo{person}{Gang Wang}.} \bibinfo{year}{2022}\natexlab{}.
\newblock \showarticletitle{Knowledge distillation via the target-aware transformer}. In \bibinfo{booktitle}{\emph{Proceedings of the IEEE/CVF Conference on Computer Vision and Pattern Recognition}}. \bibinfo{pages}{10915--10924}.
\newblock


\bibitem[Lin et~al\mbox{.}(2021)]%
        {lin2021mood}
\bibfield{author}{\bibinfo{person}{Ziqian Lin}, \bibinfo{person}{Sreya~Dutta Roy}, {and} \bibinfo{person}{Yixuan Li}.} \bibinfo{year}{2021}\natexlab{}.
\newblock \showarticletitle{Mood: Multi-level out-of-distribution detection}. In \bibinfo{booktitle}{\emph{Proceedings of the IEEE/CVF conference on Computer Vision and Pattern Recognition}}. \bibinfo{pages}{15313--15323}.
\newblock


\bibitem[Liu et~al\mbox{.}(2024)]%
        {liu2024enhancingcontinuousdomainadaptation}
\bibfield{author}{\bibinfo{person}{Hanbing Liu}, \bibinfo{person}{Jingge Wang}, \bibinfo{person}{Xuan Zhang}, \bibinfo{person}{Ye Guo}, {and} \bibinfo{person}{Yang Li}.} \bibinfo{year}{2024}\natexlab{}.
\newblock \showarticletitle{Enhancing Continuous Domain Adaptation with Multi-path Transfer Curriculum}. In \bibinfo{booktitle}{\emph{Pacific-Asia Conference on Knowledge Discovery and Data Mining}}. Springer, \bibinfo{pages}{286--298}.
\newblock


\bibitem[Liu et~al\mbox{.}(2020)]%
        {liu2020energy}
\bibfield{author}{\bibinfo{person}{Weitang Liu}, \bibinfo{person}{Xiaoyun Wang}, \bibinfo{person}{John Owens}, {and} \bibinfo{person}{Yixuan Li}.} \bibinfo{year}{2020}\natexlab{}.
\newblock \showarticletitle{Energy-based out-of-distribution detection}.
\newblock \bibinfo{journal}{\emph{Advances in neural information processing systems}}  \bibinfo{volume}{33} (\bibinfo{year}{2020}), \bibinfo{pages}{21464--21475}.
\newblock


\bibitem[Long et~al\mbox{.}(2015)]%
        {long2015learning}
\bibfield{author}{\bibinfo{person}{Mingsheng Long}, \bibinfo{person}{Yue Cao}, \bibinfo{person}{Jianmin Wang}, {and} \bibinfo{person}{Michael Jordan}.} \bibinfo{year}{2015}\natexlab{}.
\newblock \showarticletitle{Learning transferable features with deep adaptation networks}. In \bibinfo{booktitle}{\emph{International conference on machine learning}}. PMLR, \bibinfo{pages}{97--105}.
\newblock


\bibitem[Long et~al\mbox{.}(2018)]%
        {long2018conditional}
\bibfield{author}{\bibinfo{person}{Mingsheng Long}, \bibinfo{person}{Zhangjie Cao}, \bibinfo{person}{Jianmin Wang}, {and} \bibinfo{person}{Michael~I Jordan}.} \bibinfo{year}{2018}\natexlab{}.
\newblock \showarticletitle{Conditional adversarial domain adaptation}.
\newblock \bibinfo{journal}{\emph{Advances in neural information processing systems}}  \bibinfo{volume}{31} (\bibinfo{year}{2018}).
\newblock


\bibitem[Long et~al\mbox{.}(2016)]%
        {long2016unsupervised}
\bibfield{author}{\bibinfo{person}{Mingsheng Long}, \bibinfo{person}{Han Zhu}, \bibinfo{person}{Jianmin Wang}, {and} \bibinfo{person}{Michael~I Jordan}.} \bibinfo{year}{2016}\natexlab{}.
\newblock \showarticletitle{Unsupervised domain adaptation with residual transfer networks}.
\newblock \bibinfo{journal}{\emph{Advances in neural information processing systems}}  \bibinfo{volume}{29} (\bibinfo{year}{2016}).
\newblock


\bibitem[Long et~al\mbox{.}(2017)]%
        {long2017deep}
\bibfield{author}{\bibinfo{person}{Mingsheng Long}, \bibinfo{person}{Han Zhu}, \bibinfo{person}{Jianmin Wang}, {and} \bibinfo{person}{Michael~I Jordan}.} \bibinfo{year}{2017}\natexlab{}.
\newblock \showarticletitle{Deep transfer learning with joint adaptation networks}. In \bibinfo{booktitle}{\emph{International conference on machine learning}}. PMLR, \bibinfo{pages}{2208--2217}.
\newblock


\bibitem[Ma and Zhang(2022)]%
        {ma_bayesian_2022}
\bibfield{author}{\bibinfo{person}{Yuchi Ma} {and} \bibinfo{person}{Zhou Zhang}.} \bibinfo{year}{2022}\natexlab{}.
\newblock \showarticletitle{A {Bayesian} {Domain} {Adversarial} {Neural} {Network} for {Corn} {Yield} {Prediction}}.
\newblock \bibinfo{journal}{\emph{IEEE Geoscience and Remote Sensing Letters}}  \bibinfo{volume}{19} (\bibinfo{year}{2022}), \bibinfo{pages}{1--5}.
\newblock
\showISSN{1545-598X, 1558-0571}


\bibitem[Mansour et~al\mbox{.}(2008)]%
        {mansour2008domain}
\bibfield{author}{\bibinfo{person}{Yishay Mansour}, \bibinfo{person}{Mehryar Mohri}, {and} \bibinfo{person}{Afshin Rostamizadeh}.} \bibinfo{year}{2008}\natexlab{}.
\newblock \showarticletitle{Domain adaptation with multiple sources}.
\newblock \bibinfo{journal}{\emph{Advances in neural information processing systems}}  \bibinfo{volume}{21} (\bibinfo{year}{2008}).
\newblock


\bibitem[Martini et~al\mbox{.}(2021)]%
        {martini_domain-adversarial_2021}
\bibfield{author}{\bibinfo{person}{Mauro Martini}, \bibinfo{person}{Vittorio Mazzia}, \bibinfo{person}{Aleem Khaliq}, {and} \bibinfo{person}{Marcello Chiaberge}.} \bibinfo{year}{2021}\natexlab{}.
\newblock \showarticletitle{Domain-{Adversarial} {Training} of {Self}-{Attention}-{Based} {Networks} for {Land} {Cover} {Classification} {Using} {Multi}-{Temporal} {Sentinel}-2 {Satellite} {Imagery}}.
\newblock \bibinfo{journal}{\emph{Remote Sensing}} \bibinfo{volume}{13}, \bibinfo{number}{13} (\bibinfo{date}{June} \bibinfo{year}{2021}), \bibinfo{pages}{2564}.
\newblock
\showISSN{2072-4292}


\bibitem[Mindermann et~al\mbox{.}(2022)]%
        {mindermann2022prioritized}
\bibfield{author}{\bibinfo{person}{S{\"o}ren Mindermann}, \bibinfo{person}{Jan~M Brauner}, \bibinfo{person}{Muhammed~T Razzak}, \bibinfo{person}{Mrinank Sharma}, \bibinfo{person}{Andreas Kirsch}, \bibinfo{person}{Winnie Xu}, \bibinfo{person}{Benedikt H{\"o}ltgen}, \bibinfo{person}{Aidan~N Gomez}, \bibinfo{person}{Adrien Morisot}, \bibinfo{person}{Sebastian Farquhar}, {et~al\mbox{.}}} \bibinfo{year}{2022}\natexlab{}.
\newblock \showarticletitle{Prioritized training on points that are learnable, worth learning, and not yet learnt}. In \bibinfo{booktitle}{\emph{International Conference on Machine Learning}}. PMLR, \bibinfo{pages}{15630--15649}.
\newblock


\bibitem[Morteza and Li(2022)]%
        {morteza2022provable}
\bibfield{author}{\bibinfo{person}{Peyman Morteza} {and} \bibinfo{person}{Yixuan Li}.} \bibinfo{year}{2022}\natexlab{}.
\newblock \showarticletitle{Provable guarantees for understanding out-of-distribution detection}. In \bibinfo{booktitle}{\emph{Proceedings of the AAAI Conference on Artificial Intelligence}}, Vol.~\bibinfo{volume}{36}. \bibinfo{pages}{7831--7840}.
\newblock


\bibitem[Muandet et~al\mbox{.}(2017)]%
        {Muandet_2017}
\bibfield{author}{\bibinfo{person}{Krikamol Muandet}, \bibinfo{person}{Kenji Fukumizu}, \bibinfo{person}{Bharath Sriperumbudur}, {and} \bibinfo{person}{Bernhard Schölkopf}.} \bibinfo{year}{2017}\natexlab{}.
\newblock \showarticletitle{Kernel Mean Embedding of Distributions: A Review and Beyond}.
\newblock \bibinfo{journal}{\emph{Foundations and Trends® in Machine Learning}} \bibinfo{volume}{10}, \bibinfo{number}{1–2} (\bibinfo{year}{2017}), \bibinfo{pages}{1–141}.
\newblock
\showISSN{1935-8245}


\bibitem[Nguyen et~al\mbox{.}(2020)]%
        {pmlr-v119-nguyen20b}
\bibfield{author}{\bibinfo{person}{Cuong Nguyen}, \bibinfo{person}{Tal Hassner}, \bibinfo{person}{Matthias Seeger}, {and} \bibinfo{person}{Cedric Archambeau}.} \bibinfo{year}{2020}\natexlab{}.
\newblock \showarticletitle{{LEEP}: A New Measure to Evaluate Transferability of Learned Representations}. In \bibinfo{booktitle}{\emph{Proceedings of the 37th International Conference on Machine Learning}} \emph{(\bibinfo{series}{Proceedings of Machine Learning Research}, Vol.~\bibinfo{volume}{119})}, \bibfield{editor}{\bibinfo{person}{Hal~Daumé III} {and} \bibinfo{person}{Aarti Singh}} (Eds.). \bibinfo{publisher}{PMLR}, \bibinfo{pages}{7294--7305}.
\newblock


\bibitem[Nguyen et~al\mbox{.}(2023)]%
        {nguyen2023simpletransferabilityestimationregression}
\bibfield{author}{\bibinfo{person}{Cuong~N Nguyen}, \bibinfo{person}{Phong Tran}, \bibinfo{person}{Lam Si~Tung Ho}, \bibinfo{person}{Vu Dinh}, \bibinfo{person}{Anh~T Tran}, \bibinfo{person}{Tal Hassner}, {and} \bibinfo{person}{Cuong~V Nguyen}.} \bibinfo{year}{2023}\natexlab{}.
\newblock \showarticletitle{Simple transferability estimation for regression tasks}. In \bibinfo{booktitle}{\emph{Uncertainty in Artificial Intelligence}}. PMLR, \bibinfo{pages}{1510--1521}.
\newblock


\bibitem[Ouyang et~al\mbox{.}(2022)]%
        {ouyang2022training}
\bibfield{author}{\bibinfo{person}{Long Ouyang}, \bibinfo{person}{Jeffrey Wu}, \bibinfo{person}{Xu Jiang}, \bibinfo{person}{Diogo Almeida}, \bibinfo{person}{Carroll Wainwright}, \bibinfo{person}{Pamela Mishkin}, \bibinfo{person}{Chong Zhang}, \bibinfo{person}{Sandhini Agarwal}, \bibinfo{person}{Katarina Slama}, \bibinfo{person}{Alex Ray}, {et~al\mbox{.}}} \bibinfo{year}{2022}\natexlab{}.
\newblock \showarticletitle{Training language models to follow instructions with human feedback}.
\newblock \bibinfo{journal}{\emph{Advances in neural information processing systems}}  \bibinfo{volume}{35} (\bibinfo{year}{2022}), \bibinfo{pages}{27730--27744}.
\newblock


\bibitem[Pan and Yang(2010)]%
        {Pan2010Survey}
\bibfield{author}{\bibinfo{person}{Sinno~Jialin Pan} {and} \bibinfo{person}{Qiang Yang}.} \bibinfo{year}{2010}\natexlab{}.
\newblock \showarticletitle{A Survey on Transfer Learning}.
\newblock \bibinfo{journal}{\emph{IEEE Transactions on Knowledge and Data Engineering}} \bibinfo{volume}{22}, \bibinfo{number}{10} (\bibinfo{year}{2010}), \bibinfo{pages}{1345--1359}.
\newblock


\bibitem[Prabhu et~al\mbox{.}(2021)]%
        {DBLP:journals/corr/abs-2010-08666}
\bibfield{author}{\bibinfo{person}{Viraj Prabhu}, \bibinfo{person}{Arjun Chandrasekaran}, \bibinfo{person}{Kate Saenko}, {and} \bibinfo{person}{Judy Hoffman}.} \bibinfo{year}{2021}\natexlab{}.
\newblock \showarticletitle{Active domain adaptation via clustering uncertainty-weighted embeddings}. In \bibinfo{booktitle}{\emph{Proceedings of the IEEE/CVF international conference on computer vision}}. \bibinfo{pages}{8505--8514}.
\newblock


\bibitem[Priyatikanto et~al\mbox{.}(2023)]%
        {priyatikanto2023improving}
\bibfield{author}{\bibinfo{person}{Rhorom Priyatikanto}, \bibinfo{person}{Yang Lu}, \bibinfo{person}{Jadu Dash}, {and} \bibinfo{person}{Justin Sheffield}.} \bibinfo{year}{2023}\natexlab{}.
\newblock \showarticletitle{Improving generalisability and transferability of machine-learning-based maize yield prediction model through domain adaptation}.
\newblock \bibinfo{journal}{\emph{Agricultural and Forest Meteorology}}  \bibinfo{volume}{341} (\bibinfo{year}{2023}), \bibinfo{pages}{109652}.
\newblock


\bibitem[Qi et~al\mbox{.}(2022)]%
        {qi2023transferabilityestimationbasedprincipal}
\bibfield{author}{\bibinfo{person}{Huiyan Qi}, \bibinfo{person}{Lechao Cheng}, \bibinfo{person}{Jingjing Chen}, \bibinfo{person}{Yue Yu}, \bibinfo{person}{Xue Song}, \bibinfo{person}{Zunlei Feng}, {and} \bibinfo{person}{Yu-Gang Jiang}.} \bibinfo{year}{2022}\natexlab{}.
\newblock \showarticletitle{Transferability estimation based on principal gradient expectation}.
\newblock \bibinfo{journal}{\emph{arXiv preprint arXiv:2211.16299}} (\bibinfo{year}{2022}).
\newblock


\bibitem[Qin et~al\mbox{.}(2023a)]%
        {qin2023transferabilitylearningmodelssemantic}
\bibfield{author}{\bibinfo{person}{Rongjun Qin}, \bibinfo{person}{Guixiang Zhang}, {and} \bibinfo{person}{Yang Tang}.} \bibinfo{year}{2023}\natexlab{a}.
\newblock \showarticletitle{On the transferability of learning models for semantic segmentation for remote sensing data}.
\newblock \bibinfo{journal}{\emph{arXiv preprint arXiv:2310.10490}} (\bibinfo{year}{2023}).
\newblock


\bibitem[Qin et~al\mbox{.}(2023b)]%
        {qin_transferability_nodate}
\bibfield{author}{\bibinfo{person}{Rongjun Qin}, \bibinfo{person}{Guixiang Zhang}, {and} \bibinfo{person}{Yang Tang}.} \bibinfo{year}{2023}\natexlab{b}.
\newblock \showarticletitle{On the transferability of learning models for semantic segmentation for remote sensing data}.
\newblock \bibinfo{journal}{\emph{arXiv preprint arXiv:2310.10490}} (\bibinfo{year}{2023}).
\newblock


\bibitem[Radford et~al\mbox{.}(2021)]%
        {radford2021learning}
\bibfield{author}{\bibinfo{person}{Alec Radford}, \bibinfo{person}{Jong~Wook Kim}, \bibinfo{person}{Chris Hallacy}, \bibinfo{person}{Aditya Ramesh}, \bibinfo{person}{Gabriel Goh}, \bibinfo{person}{Sandhini Agarwal}, \bibinfo{person}{Girish Sastry}, \bibinfo{person}{Amanda Askell}, \bibinfo{person}{Pamela Mishkin}, \bibinfo{person}{Jack Clark}, {et~al\mbox{.}}} \bibinfo{year}{2021}\natexlab{}.
\newblock \showarticletitle{Learning transferable visual models from natural language supervision}. In \bibinfo{booktitle}{\emph{International conference on machine learning}}. PmLR, \bibinfo{pages}{8748--8763}.
\newblock


\bibitem[Rosenblat(1956)]%
        {rosenblat1956remarks}
\bibfield{author}{\bibinfo{person}{M Rosenblat}.} \bibinfo{year}{1956}\natexlab{}.
\newblock \showarticletitle{Remarks on some nonparametric estimates of a density function}.
\newblock \bibinfo{journal}{\emph{Ann. Math. Stat}}  \bibinfo{volume}{27} (\bibinfo{year}{1956}), \bibinfo{pages}{832--837}.
\newblock


\bibitem[Sagawa et~al\mbox{.}({[n.\,d.]})]%
        {sagawadistributionally}
\bibfield{author}{\bibinfo{person}{Shiori Sagawa}, \bibinfo{person}{Pang~Wei Koh}, \bibinfo{person}{Tatsunori~B Hashimoto}, {and} \bibinfo{person}{Percy Liang}.} \bibinfo{year}{[n.\,d.]}\natexlab{}.
\newblock \showarticletitle{Distributionally Robust Neural Networks}. In \bibinfo{booktitle}{\emph{International Conference on Learning Representations}}.
\newblock


\bibitem[Shen et~al\mbox{.}(2018)]%
        {shen2018wasserstein}
\bibfield{author}{\bibinfo{person}{Jian Shen}, \bibinfo{person}{Yanru Qu}, \bibinfo{person}{Weinan Zhang}, {and} \bibinfo{person}{Yong Yu}.} \bibinfo{year}{2018}\natexlab{}.
\newblock \showarticletitle{Wasserstein distance guided representation learning for domain adaptation}. In \bibinfo{booktitle}{\emph{Proceedings of the AAAI conference on artificial intelligence}}, Vol.~\bibinfo{volume}{32}.
\newblock


\bibitem[Shi et~al\mbox{.}(2020)]%
        {shi2020auxiliary}
\bibfield{author}{\bibinfo{person}{Baifeng Shi}, \bibinfo{person}{Judy Hoffman}, \bibinfo{person}{Kate Saenko}, \bibinfo{person}{Trevor Darrell}, {and} \bibinfo{person}{Huijuan Xu}.} \bibinfo{year}{2020}\natexlab{}.
\newblock \showarticletitle{Auxiliary task reweighting for minimum-data learning}.
\newblock \bibinfo{journal}{\emph{Advances in Neural Information Processing Systems}}  \bibinfo{volume}{33} (\bibinfo{year}{2020}), \bibinfo{pages}{7148--7160}.
\newblock


\bibitem[Shi et~al\mbox{.}(2022)]%
        {shi2022domain}
\bibfield{author}{\bibinfo{person}{Yaowei Shi}, \bibinfo{person}{Aidong Deng}, \bibinfo{person}{Minqiang Deng}, \bibinfo{person}{Jing Li}, \bibinfo{person}{Meng Xu}, \bibinfo{person}{Shun Zhang}, \bibinfo{person}{Xue Ding}, {and} \bibinfo{person}{Shuo Xu}.} \bibinfo{year}{2022}\natexlab{}.
\newblock \showarticletitle{Domain transferability-based deep domain generalization method towards actual fault diagnosis scenarios}.
\newblock \bibinfo{journal}{\emph{IEEE Transactions on Industrial Informatics}} \bibinfo{volume}{19}, \bibinfo{number}{6} (\bibinfo{year}{2022}), \bibinfo{pages}{7355--7366}.
\newblock


\bibitem[Sinha et~al\mbox{.}(2017)]%
        {sinha2018certifiable}
\bibfield{author}{\bibinfo{person}{Aman Sinha}, \bibinfo{person}{Hongseok Namkoong}, {and} \bibinfo{person}{John~C Duchi}.} \bibinfo{year}{2017}\natexlab{}.
\newblock \showarticletitle{Certifiable Distributional Robustness with Principled Adversarial Training. CoRR, abs/1710.10571}.
\newblock \bibinfo{journal}{\emph{arXiv preprint arXiv:1710.10571}} (\bibinfo{year}{2017}).
\newblock


\bibitem[Song et~al\mbox{.}(2019)]%
        {DeepAttributionMaps}
\bibfield{author}{\bibinfo{person}{Jie Song}, \bibinfo{person}{Yixin Chen}, \bibinfo{person}{Xinchao Wang}, \bibinfo{person}{Chengchao Shen}, {and} \bibinfo{person}{Mingli Song}.} \bibinfo{year}{2019}\natexlab{}.
\newblock \showarticletitle{Deep model transferability from attribution maps}.
\newblock \bibinfo{journal}{\emph{Advances in Neural Information Processing Systems}}  \bibinfo{volume}{32} (\bibinfo{year}{2019}).
\newblock


\bibitem[Soviany et~al\mbox{.}(2022)]%
        {soviany2022curriculum}
\bibfield{author}{\bibinfo{person}{Petru Soviany}, \bibinfo{person}{Radu~Tudor Ionescu}, \bibinfo{person}{Paolo Rota}, {and} \bibinfo{person}{Nicu Sebe}.} \bibinfo{year}{2022}\natexlab{}.
\newblock \showarticletitle{Curriculum learning: A survey}.
\newblock \bibinfo{journal}{\emph{International Journal of Computer Vision}} \bibinfo{volume}{130}, \bibinfo{number}{6} (\bibinfo{year}{2022}), \bibinfo{pages}{1526--1565}.
\newblock


\bibitem[Su et~al\mbox{.}(2022)]%
        {su-etal-2022-transferability}
\bibfield{author}{\bibinfo{person}{Yusheng Su}, \bibinfo{person}{Xiaozhi Wang}, \bibinfo{person}{Yujia Qin}, \bibinfo{person}{Chi-Min Chan}, \bibinfo{person}{Yankai Lin}, \bibinfo{person}{Huadong Wang}, \bibinfo{person}{Kaiyue Wen}, \bibinfo{person}{Zhiyuan Liu}, \bibinfo{person}{Peng Li}, \bibinfo{person}{Juanzi Li}, \bibinfo{person}{Lei Hou}, \bibinfo{person}{Maosong Sun}, {and} \bibinfo{person}{Jie Zhou}.} \bibinfo{year}{2022}\natexlab{}.
\newblock \showarticletitle{On Transferability of Prompt Tuning for Natural Language Processing}. In \bibinfo{booktitle}{\emph{Proceedings of the 2022 Conference of the North American Chapter of the Association for Computational Linguistics: Human Language Technologies}}, \bibfield{editor}{\bibinfo{person}{Marine Carpuat}, \bibinfo{person}{Marie-Catherine de~Marneffe}, {and} \bibinfo{person}{Ivan~Vladimir Meza~Ruiz}} (Eds.). \bibinfo{publisher}{Association for Computational Linguistics}, \bibinfo{address}{Seattle, United States}, \bibinfo{pages}{3949--3969}.
\newblock


\bibitem[Sun et~al\mbox{.}(2017)]%
        {sun2017correlation}
\bibfield{author}{\bibinfo{person}{Baochen Sun}, \bibinfo{person}{Jiashi Feng}, {and} \bibinfo{person}{Kate Saenko}.} \bibinfo{year}{2017}\natexlab{}.
\newblock \showarticletitle{Correlation alignment for unsupervised domain adaptation}.
\newblock \bibinfo{journal}{\emph{Domain adaptation in computer vision applications}} (\bibinfo{year}{2017}), \bibinfo{pages}{153--171}.
\newblock


\bibitem[Sun et~al\mbox{.}(2020)]%
        {sun2020meta}
\bibfield{author}{\bibinfo{person}{Qianru Sun}, \bibinfo{person}{Yaoyao Liu}, \bibinfo{person}{Zhaozheng Chen}, \bibinfo{person}{Tat-Seng Chua}, {and} \bibinfo{person}{Bernt Schiele}.} \bibinfo{year}{2020}\natexlab{}.
\newblock \showarticletitle{Meta-transfer learning through hard tasks}.
\newblock \bibinfo{journal}{\emph{IEEE Transactions on Pattern Analysis and Machine Intelligence}} \bibinfo{volume}{44}, \bibinfo{number}{3} (\bibinfo{year}{2020}), \bibinfo{pages}{1443--1456}.
\newblock


\bibitem[Sun et~al\mbox{.}(2015)]%
        {sun2015survey}
\bibfield{author}{\bibinfo{person}{Shiliang Sun}, \bibinfo{person}{Honglei Shi}, {and} \bibinfo{person}{Yuanbin Wu}.} \bibinfo{year}{2015}\natexlab{}.
\newblock \showarticletitle{A survey of multi-source domain adaptation}.
\newblock \bibinfo{journal}{\emph{Information Fusion}}  \bibinfo{volume}{24} (\bibinfo{year}{2015}), \bibinfo{pages}{84--92}.
\newblock


\bibitem[Tan et~al\mbox{.}(2024)]%
        {tan2024transferability}
\bibfield{author}{\bibinfo{person}{Yang Tan}, \bibinfo{person}{Enming Zhang}, \bibinfo{person}{Yang Li}, \bibinfo{person}{Shao-Lun Huang}, {and} \bibinfo{person}{Xiao-Ping Zhang}.} \bibinfo{year}{2024}\natexlab{}.
\newblock \showarticletitle{Transferability-guided cross-domain cross-task transfer learning}.
\newblock \bibinfo{journal}{\emph{IEEE Transactions on Neural Networks and Learning Systems}} (\bibinfo{year}{2024}).
\newblock


\bibitem[Tong et~al\mbox{.}(2021)]%
        {tong2021mathematical}
\bibfield{author}{\bibinfo{person}{Xinyi Tong}, \bibinfo{person}{Xiangxiang Xu}, \bibinfo{person}{Shao-Lun Huang}, {and} \bibinfo{person}{Lizhong Zheng}.} \bibinfo{year}{2021}\natexlab{}.
\newblock \showarticletitle{A mathematical framework for quantifying transferability in multi-source transfer learning}.
\newblock \bibinfo{journal}{\emph{Advances in Neural Information Processing Systems}}  \bibinfo{volume}{34} (\bibinfo{year}{2021}), \bibinfo{pages}{26103--26116}.
\newblock


\bibitem[Tran et~al\mbox{.}(2019)]%
        {tran2019transferabilityhardnesssupervisedclassification}
\bibfield{author}{\bibinfo{person}{Anh~T Tran}, \bibinfo{person}{Cuong~V Nguyen}, {and} \bibinfo{person}{Tal Hassner}.} \bibinfo{year}{2019}\natexlab{}.
\newblock \showarticletitle{Transferability and hardness of supervised classification tasks}. In \bibinfo{booktitle}{\emph{Proceedings of the IEEE/CVF international conference on computer vision}}. \bibinfo{pages}{1395--1405}.
\newblock


\bibitem[Tzeng et~al\mbox{.}(2017)]%
        {tzeng2017adversarial}
\bibfield{author}{\bibinfo{person}{Eric Tzeng}, \bibinfo{person}{Judy Hoffman}, \bibinfo{person}{Kate Saenko}, {and} \bibinfo{person}{Trevor Darrell}.} \bibinfo{year}{2017}\natexlab{}.
\newblock \showarticletitle{Adversarial discriminative domain adaptation}. In \bibinfo{booktitle}{\emph{Proceedings of the IEEE conference on computer vision and pattern recognition}}. \bibinfo{pages}{7167--7176}.
\newblock


\bibitem[Wan et~al\mbox{.}(2022)]%
        {wan_combining_2022}
\bibfield{author}{\bibinfo{person}{Liang Wan}, \bibinfo{person}{Weijun Zhou}, \bibinfo{person}{Yong He}, \bibinfo{person}{Thomas~Cherico Wanger}, {and} \bibinfo{person}{Haiyan Cen}.} \bibinfo{year}{2022}\natexlab{}.
\newblock \showarticletitle{Combining transfer learning and hyperspectral reflectance analysis to assess leaf nitrogen concentration across different plant species datasets}.
\newblock \bibinfo{journal}{\emph{Remote Sensing of Environment}}  \bibinfo{volume}{269} (\bibinfo{date}{Feb.} \bibinfo{year}{2022}), \bibinfo{pages}{112826}.
\newblock
\showISSN{00344257}


\bibitem[Wang et~al\mbox{.}(2024)]%
        {wang2024generalizing}
\bibfield{author}{\bibinfo{person}{Jingge Wang}, \bibinfo{person}{Liyan Xie}, \bibinfo{person}{Yao Xie}, \bibinfo{person}{Shao-Lun Huang}, {and} \bibinfo{person}{Yang Li}.} \bibinfo{year}{2024}\natexlab{}.
\newblock \showarticletitle{Generalizing to unseen domains with Wasserstein distributional robustness under limited source knowledge}.
\newblock \bibinfo{journal}{\emph{IEEE Journal of Selected Topics in Signal Processing}} (\bibinfo{year}{2024}).
\newblock


\bibitem[Wang and Yoon(2021)]%
        {wang2021knowledge}
\bibfield{author}{\bibinfo{person}{Lin Wang} {and} \bibinfo{person}{Kuk-Jin Yoon}.} \bibinfo{year}{2021}\natexlab{}.
\newblock \showarticletitle{Knowledge distillation and student-teacher learning for visual intelligence: A review and new outlooks}.
\newblock \bibinfo{journal}{\emph{IEEE transactions on pattern analysis and machine intelligence}} \bibinfo{volume}{44}, \bibinfo{number}{6} (\bibinfo{year}{2021}), \bibinfo{pages}{3048--3068}.
\newblock


\bibitem[Wang et~al\mbox{.}(2022)]%
        {wang_exploring_2022}
\bibfield{author}{\bibinfo{person}{Yumiao Wang}, \bibinfo{person}{Luwei Feng}, \bibinfo{person}{Weiwei Sun}, \bibinfo{person}{Zhou Zhang}, \bibinfo{person}{Hanyu Zhang}, \bibinfo{person}{Gang Yang}, {and} \bibinfo{person}{Xiangchao Meng}.} \bibinfo{year}{2022}\natexlab{}.
\newblock \showarticletitle{Exploring the potential of multi-source unsupervised domain adaptation in crop mapping using {Sentinel}-2 images}.
\newblock \bibinfo{journal}{\emph{GIScience \& Remote Sensing}} \bibinfo{volume}{59}, \bibinfo{number}{1} (\bibinfo{date}{Dec.} \bibinfo{year}{2022}), \bibinfo{pages}{2247--2265}.
\newblock
\showISSN{1548-1603, 1943-7226}


\bibitem[Wei et~al\mbox{.}(2020)]%
        {WEI2020106420}
\bibfield{author}{\bibinfo{person}{Xing Wei}, \bibinfo{person}{Shaofan Liu}, \bibinfo{person}{Yaoci Xiang}, \bibinfo{person}{Zhangling Duan}, \bibinfo{person}{Chong Zhao}, {and} \bibinfo{person}{Yang Lu}.} \bibinfo{year}{2020}\natexlab{}.
\newblock \showarticletitle{Incremental learning based multi-domain adaptation for object detection}.
\newblock \bibinfo{journal}{\emph{Knowledge-Based Systems}}  \bibinfo{volume}{210} (\bibinfo{year}{2020}), \bibinfo{pages}{106420}.
\newblock
\showISSN{0950-7051}


\bibitem[Wu and He(2024)]%
        {wu2024trustworthy}
\bibfield{author}{\bibinfo{person}{Jun Wu} {and} \bibinfo{person}{Jingrui He}.} \bibinfo{year}{2024}\natexlab{}.
\newblock \showarticletitle{Trustworthy Transfer Learning: A Survey}.
\newblock \bibinfo{journal}{\emph{arXiv preprint arXiv:2412.14116}} (\bibinfo{year}{2024}).
\newblock


\bibitem[Wu et~al\mbox{.}(2024a)]%
        {wu2024generalization}
\bibfield{author}{\bibinfo{person}{Xuetong Wu}, \bibinfo{person}{Jonathan~H Manton}, \bibinfo{person}{Uwe Aickelin}, {and} \bibinfo{person}{Jingge Zhu}.} \bibinfo{year}{2024}\natexlab{a}.
\newblock \showarticletitle{On the generalization for transfer learning: An information-theoretic analysis}.
\newblock \bibinfo{journal}{\emph{IEEE Transactions on Information Theory}} (\bibinfo{year}{2024}).
\newblock


\bibitem[Wu et~al\mbox{.}(2024b)]%
        {wu2024h}
\bibfield{author}{\bibinfo{person}{Yanru Wu}, \bibinfo{person}{Jianning Wang}, \bibinfo{person}{Weida Wang}, {and} \bibinfo{person}{Yang Li}.} \bibinfo{year}{2024}\natexlab{b}.
\newblock \showarticletitle{H-ensemble: An Information Theoretic Approach to Reliable Few-Shot Multi-Source-Free Transfer}. In \bibinfo{booktitle}{\emph{Proceedings of the AAAI Conference on Artificial Intelligence}}, Vol.~\bibinfo{volume}{38}. \bibinfo{pages}{15970--15978}.
\newblock


\bibitem[Xu and Kang(2023)]%
        {xu2023fastaccuratetransferabilitymeasurement}
\bibfield{author}{\bibinfo{person}{Huiwen Xu} {and} \bibinfo{person}{U Kang}.} \bibinfo{year}{2023}\natexlab{}.
\newblock \showarticletitle{Fast and accurate transferability measurement by evaluating intra-class feature variance}. In \bibinfo{booktitle}{\emph{Proceedings of the IEEE/CVF International Conference on Computer Vision}}. \bibinfo{pages}{11474--11482}.
\newblock


\bibitem[Xue et~al\mbox{.}(2024)]%
        {10639517}
\bibfield{author}{\bibinfo{person}{Yihao Xue}, \bibinfo{person}{Rui Yang}, \bibinfo{person}{Xiaohan Chen}, \bibinfo{person}{Weibo Liu}, \bibinfo{person}{Zidong Wang}, {and} \bibinfo{person}{Xiaohui Liu}.} \bibinfo{year}{2024}\natexlab{}.
\newblock \showarticletitle{A Review on Transferability Estimation in Deep Transfer Learning}.
\newblock \bibinfo{journal}{\emph{IEEE Transactions on Artificial Intelligence}} \bibinfo{volume}{5}, \bibinfo{number}{12} (\bibinfo{year}{2024}), \bibinfo{pages}{5894--5914}.
\newblock


\bibitem[Yang et~al\mbox{.}(2023b)]%
        {yang2023transferability}
\bibfield{author}{\bibinfo{person}{Chen Yang}, \bibinfo{person}{Yifan Liu}, {and} \bibinfo{person}{Yixuan Yuan}.} \bibinfo{year}{2023}\natexlab{b}.
\newblock \showarticletitle{Transferability-Guided Multi-source Model Adaptation for Medical Image Segmentation}. In \bibinfo{booktitle}{\emph{International Conference on Medical Image Computing and Computer-Assisted Intervention}}. Springer, \bibinfo{pages}{703--712}.
\newblock


\bibitem[Yang et~al\mbox{.}(2024)]%
        {yang2024generalized}
\bibfield{author}{\bibinfo{person}{Jingkang Yang}, \bibinfo{person}{Kaiyang Zhou}, \bibinfo{person}{Yixuan Li}, {and} \bibinfo{person}{Ziwei Liu}.} \bibinfo{year}{2024}\natexlab{}.
\newblock \showarticletitle{Generalized out-of-distribution detection: A survey}.
\newblock \bibinfo{journal}{\emph{International Journal of Computer Vision}} (\bibinfo{year}{2024}), \bibinfo{pages}{1--28}.
\newblock


\bibitem[Yang et~al\mbox{.}(2023a)]%
        {Yang2023}
\bibfield{author}{\bibinfo{person}{Yuncheng Yang} {et~al\mbox{.}}} \bibinfo{year}{2023}\natexlab{a}.
\newblock \showarticletitle{Pick the Best Pre-trained Model: Towards Transferability Estimation for Medical Image Segmentation}.
\newblock \bibinfo{journal}{\emph{arXiv preprint arXiv:2307.11958}} (\bibinfo{year}{2023}).
\newblock


\bibitem[Yang et~al\mbox{.}(2023c)]%
        {yang2023pickbestpretrainedmodel}
\bibfield{author}{\bibinfo{person}{Yuncheng Yang}, \bibinfo{person}{Meng Wei}, \bibinfo{person}{Junjun He}, \bibinfo{person}{Jie Yang}, \bibinfo{person}{Jin Ye}, {and} \bibinfo{person}{Yun Gu}.} \bibinfo{year}{2023}\natexlab{c}.
\newblock \showarticletitle{Pick the best pre-trained model: Towards transferability estimation for medical image segmentation}. In \bibinfo{booktitle}{\emph{International Conference on Medical Image Computing and Computer-Assisted Intervention}}. Springer, \bibinfo{pages}{674--683}.
\newblock


\bibitem[Yao et~al\mbox{.}(2023)]%
        {yao2023editinglargelanguagemodels}
\bibfield{author}{\bibinfo{person}{Yunzhi Yao}, \bibinfo{person}{Peng Wang}, \bibinfo{person}{Bozhong Tian}, \bibinfo{person}{Siyuan Cheng}, \bibinfo{person}{Zhoubo Li}, \bibinfo{person}{Shumin Deng}, \bibinfo{person}{Huajun Chen}, {and} \bibinfo{person}{Ningyu Zhang}.} \bibinfo{year}{2023}\natexlab{}.
\newblock \showarticletitle{Editing large language models: Problems, methods, and opportunities}.
\newblock \bibinfo{journal}{\emph{arXiv preprint arXiv:2305.13172}} (\bibinfo{year}{2023}).
\newblock


\bibitem[Yosinski et~al\mbox{.}(2014a)]%
        {NIPS2014_375c7134}
\bibfield{author}{\bibinfo{person}{Jason Yosinski}, \bibinfo{person}{Jeff Clune}, \bibinfo{person}{Yoshua Bengio}, {and} \bibinfo{person}{Hod Lipson}.} \bibinfo{year}{2014}\natexlab{a}.
\newblock \showarticletitle{How transferable are features in deep neural networks?}. In \bibinfo{booktitle}{\emph{Advances in Neural Information Processing Systems}}, \bibfield{editor}{\bibinfo{person}{Z.~Ghahramani}, \bibinfo{person}{M.~Welling}, \bibinfo{person}{C.~Cortes}, \bibinfo{person}{N.~Lawrence}, {and} \bibinfo{person}{K.Q. Weinberger}} (Eds.), Vol.~\bibinfo{volume}{27}. \bibinfo{publisher}{Curran Associates, Inc.}
\newblock


\bibitem[Yosinski et~al\mbox{.}(2014b)]%
        {yosinski2014transferable}
\bibfield{author}{\bibinfo{person}{Jason Yosinski}, \bibinfo{person}{Jeff Clune}, \bibinfo{person}{Yoshua Bengio}, {and} \bibinfo{person}{Hod Lipson}.} \bibinfo{year}{2014}\natexlab{b}.
\newblock \showarticletitle{How transferable are features in deep neural networks?}
\newblock \bibinfo{journal}{\emph{Advances in neural information processing systems}}  \bibinfo{volume}{27} (\bibinfo{year}{2014}).
\newblock


\bibitem[You et~al\mbox{.}(2021)]%
        {you2021logmepracticalassessmentpretrained}
\bibfield{author}{\bibinfo{person}{Kaichao You}, \bibinfo{person}{Yong Liu}, \bibinfo{person}{Jianmin Wang}, {and} \bibinfo{person}{Mingsheng Long}.} \bibinfo{year}{2021}\natexlab{}.
\newblock \showarticletitle{Logme: Practical assessment of pre-trained models for transfer learning}. In \bibinfo{booktitle}{\emph{International Conference on Machine Learning}}. PMLR, \bibinfo{pages}{12133--12143}.
\newblock


\bibitem[Yu et~al\mbox{.}(2020)]%
        {yu2020gradient}
\bibfield{author}{\bibinfo{person}{Tianhe Yu}, \bibinfo{person}{Saurabh Kumar}, \bibinfo{person}{Abhishek Gupta}, \bibinfo{person}{Sergey Levine}, \bibinfo{person}{Karol Hausman}, {and} \bibinfo{person}{Chelsea Finn}.} \bibinfo{year}{2020}\natexlab{}.
\newblock \showarticletitle{Gradient surgery for multi-task learning}.
\newblock \bibinfo{journal}{\emph{Advances in Neural Information Processing Systems}}  \bibinfo{volume}{33} (\bibinfo{year}{2020}), \bibinfo{pages}{5824--5836}.
\newblock


\bibitem[Yuan et~al\mbox{.}(2021)]%
        {yuan2021tokens}
\bibfield{author}{\bibinfo{person}{Li Yuan}, \bibinfo{person}{Yunpeng Chen}, \bibinfo{person}{Tao Wang}, \bibinfo{person}{Weihao Yu}, \bibinfo{person}{Yujun Shi}, \bibinfo{person}{Zi-Hang Jiang}, \bibinfo{person}{Francis~EH Tay}, \bibinfo{person}{Jiashi Feng}, {and} \bibinfo{person}{Shuicheng Yan}.} \bibinfo{year}{2021}\natexlab{}.
\newblock \showarticletitle{Tokens-to-token vit: Training vision transformers from scratch on imagenet}. In \bibinfo{booktitle}{\emph{Proceedings of the IEEE/CVF international conference on computer vision}}. \bibinfo{pages}{558--567}.
\newblock


\bibitem[Zamir et~al\mbox{.}(2018)]%
        {DBLP:journals/corr/abs-1804-08328}
\bibfield{author}{\bibinfo{person}{Amir~R Zamir}, \bibinfo{person}{Alexander Sax}, \bibinfo{person}{William Shen}, \bibinfo{person}{Leonidas~J Guibas}, \bibinfo{person}{Jitendra Malik}, {and} \bibinfo{person}{Silvio Savarese}.} \bibinfo{year}{2018}\natexlab{}.
\newblock \showarticletitle{Taskonomy: Disentangling task transfer learning}. In \bibinfo{booktitle}{\emph{Proceedings of the IEEE conference on computer vision and pattern recognition}}. \bibinfo{pages}{3712--3722}.
\newblock


\bibitem[Zhang et~al\mbox{.}(2021)]%
        {zhang2021quantifying}
\bibfield{author}{\bibinfo{person}{Guojun Zhang}, \bibinfo{person}{Han Zhao}, \bibinfo{person}{Yaoliang Yu}, {and} \bibinfo{person}{Pascal Poupart}.} \bibinfo{year}{2021}\natexlab{}.
\newblock \showarticletitle{Quantifying and improving transferability in domain generalization}.
\newblock \bibinfo{journal}{\emph{Advances in Neural Information Processing Systems}}  \bibinfo{volume}{34} (\bibinfo{year}{2021}), \bibinfo{pages}{10957--10970}.
\newblock


\bibitem[Zhong et~al\mbox{.}(2024)]%
        {zhong2024pandaprompttransfermeets}
\bibfield{author}{\bibinfo{person}{Qihuang Zhong}, \bibinfo{person}{Liang Ding}, \bibinfo{person}{Juhua Liu}, \bibinfo{person}{Bo Du}, {and} \bibinfo{person}{Dacheng Tao}.} \bibinfo{year}{2024}\natexlab{}.
\newblock \showarticletitle{Panda: Prompt transfer meets knowledge distillation for efficient model adaptation}.
\newblock \bibinfo{journal}{\emph{IEEE Transactions on Knowledge and Data Engineering}} (\bibinfo{year}{2024}).
\newblock


\bibitem[Zhuang et~al\mbox{.}(2015)]%
        {zhuang2015supervised}
\bibfield{author}{\bibinfo{person}{Fuzhen Zhuang}, \bibinfo{person}{Xiaohu Cheng}, \bibinfo{person}{Ping Luo}, \bibinfo{person}{Sinno~Jialin Pan}, {and} \bibinfo{person}{Qing He}.} \bibinfo{year}{2015}\natexlab{}.
\newblock \showarticletitle{Supervised representation learning: Transfer learning with deep autoencoders}. In \bibinfo{booktitle}{\emph{Twenty-fourth international joint conference on artificial intelligence}}.
\newblock


\end{thebibliography}

\appendix

\end{document}